\documentclass[journal,twocolumn,final]{IEEEtran}
\usepackage{latexsym}
\usepackage{graphicx}
\usepackage{amsfonts,amssymb,amsmath}
\usepackage{xspace,xcolor}

\usepackage[inline]{enumitem}

\usepackage{jabbrv}
\usepackage{cite}

\usepackage[hyphens]{url}

\def\BibTeX{{\rm B\kern-.05em{\sc i\kern-.025em b}\kern-.08em T\kern-.1667em\lower.7ex\hbox{E}\kern-.125emX}}
\newcommand{\Cms}{C_{\mathrm{S}}}
\newcommand{\Cmt}{C_{\mathrm{T}}}
\newcommand{\gms}{\gamma_{\mathrm{S}}}
\newcommand{\gmt}{\gamma_{\mathrm{T}}}
\newcommand{\hgms}{\hat{\gamma}_{\mathrm{S}}}
\newcommand{\hgmt}{\hat{\gamma}_{\mathrm{T}}}
\newcommand{\C}{\mathbf{C}} 
\newcommand{\tC}{\widetilde{C}}
\newcommand{\ga}{\gamma}

\newcommand{\Co}{\mathbb{C}}
\newcommand{\Rd}{\mathbb{R}^d}
\newcommand{\bfs}{{\mathbf s}}
\newcommand{\bfk}{{\bf k}}
\newcommand{\bfo}{{\mathbf{0}}}
\newcommand{\kk}{\|{\bfk}\|}
\newcommand{\bfr}{{\bf r}}

\newcommand{\rr}{\|{\bfr}\|}

\newcommand{\E}{\mathrm{e}} 
\newcommand{\EE}{\mathbb{E}}

\newcommand{\Do}{\mathcal{D}}

\newcommand{\om}{\zeta}
\newcommand{\atan}{{\text{tan}^{-1}}}

\newcommand{\Rei}{g_{\mathrm{re}}}
\newcommand{\Imi}{g_{\mathrm{im}}}

\newcommand{\rint}{Q_{\textrm{int}}}

\renewcommand{\d}{\text{d}}

\renewcommand{\Re}{{\rm Re}}
\renewcommand{\Im}{{\rm Im}}

\newcommand{\beq}{\begin{equation}}
	\newcommand{\eeq}{\end{equation}}
\newcommand{\bea}{\begin{eqnarray}}
	\newcommand{\eea}{\end{eqnarray}}

\newcommand{\bfc}{\mathbf{c}}   %
\newcommand{\bfx}{\mathbf{x}}

\newcommand{\gp}{\cal{GP}\xspace}
\newcommand{\gpm}{\mathcal{GP}} 

\newcommand{\bfA}{\mathbf{A}}
\newcommand{\R}{\mathbb{R}}
\newcommand{\Or}{\mathcal{O}}

\newcommand{\Na}{\mathbb{N}}

\newcommand{\bmthe}{\boldsymbol{\theta}}
\newcommand{\ft}{{\mathrm{FT}}}
\newcommand{\ift}{{\mathrm{IFT}}}
\newcommand{\D}{{\mathrm{d}}}
\newcommand{\I}{\jmath}

\DeclareMathOperator{\var}{\mathbb{V}\textrm{ar}}

\newtheorem{definition}{Definition} 
\newtheorem{propo}{Proposition}
\newtheorem{rem}{Remark}
\newtheorem{coro}{Corollary}
\newtheorem{theorem}{Theorem}
\newtheorem{lemma}{Lemma}
\begin{document}
	
\title{Non-separable Covariance Kernels  for Spatiotemporal  Gaussian Processes based on a Hybrid Spectral Method and the Harmonic Oscillator}
	
\author{Dionissios T. Hristopulos
\IEEEmembership{Senior Member, IEEE}
\thanks{D. T. Hristopulos is with the School of Electrical and Computer Engineering, Technical University of Crete, Chania 73100, Greece (e-mail: dchristopoulos@tuc.gr).}
}
	
\markboth{IEEE Transactions on Information Theory}{Hristopulos: Non-separable Covariance Kernels for Spatiotemporal  Gaussian Processes}

\IEEEpubid{\makebox[\columnwidth]{978-1-5386-5541-2/18/\$31.00~\copyright2018 IEEE \hfill}
\hspace{\columnsep}\makebox[\columnwidth]{ }}

\maketitle

\IEEEpubidadjcol

\IEEEpeerreviewmaketitle

\begin{abstract}
Gaussian processes provide a flexible, non-parametric framework for the approximation of functions in high-dimensional spaces. The  covariance kernel is the main engine of Gaussian processes,  incorporating  correlations that underpin the predictive distribution.  For applications with spatiotemporal  datasets, suitable kernels should model joint spatial and temporal dependence. Separable space-time covariance kernels offer simplicity and computational efficiency. However,
non-separable kernels include space-time interactions that  better capture observed correlations.
Most non-separable kernels that admit explicit  expressions are based on mathematical  considerations (admissibility conditions) rather than first-principles derivations.
We present a hybrid spectral approach for generating covariance kernels which is  based on physical arguments.  We use this approach to derive a new class of physically motivated, non-separable covariance kernels which have their roots in the stochastic,  linear, damped, harmonic oscillator (LDHO).  The new kernels incorporate functions with both monotonic and oscillatory decay of space-time correlations.  The LDHO covariance kernels involve space-time interactions which are introduced by  dispersion relations that modulate the oscillator coefficients.  We derive explicit relations for the spatiotemporal covariance kernels in the three oscillator regimes (underdamping, critical damping, overdamping) and investigate their properties.  We further illustrate the hybrid spectral method by deriving covariance kernels that are based on the Ornstein-Uhlenbeck model.
\end{abstract}
		
	
	
\section{INTRODUCTION}
	
\IEEEPARstart{G}{aussian} processes are a data-driven, non-parametric  machine learning method used for  nonlinear regression and classification tasks~\cite{Rasmussen06} as well as adaptive control and reinforcement learning~\cite{Liu18}. Gaussian processes define a prior over a class of functions or models.  Hence, they provide a powerful framework for the analysis of time series as well spatial and spatiotemporal data~\cite{Roberts13,dth20}.
The core of Gaussian processes is the covariance kernel, which incorporates correlations that are learned from the data.  Only non-negative definite functions can be used as covariance kernels.  Various generic admissible models  are available in the literature~\cite{Genton02,Rasmussen06,Wilson13}.  However, there is still
great interest in deriving kernels for spatiotemporal datasets~\cite{Porcu21}. Such functions, which are necessary for modeling
dynamic phenomena, need to include  physically meaningful space-time interactions. Applications of Gaussian processes with spatiotemporal kernels involve object tracking~\cite{Aftab19}, control of dynamic systems~\cite{Liu18}, systems identification~\cite{Alvarez13}, mobile sensor networks~\cite{Gu21}, and environmental process mapping~\cite{dth22}.
In spatial statistics, a predictive framework similar  to  Gaussian processes  has been developed independently based on the theory of random fields and has found many applications in the natural sciences~\cite{Adler81,Chiles12,Christakos17,dth20}. The covariance kernel is also instrumental in determining the properties of random fields.  The main differences between the Gaussian process and the random field predictive frameworks are as follows: (i) in the case of random fields the input vector is restricted to the spatial (or space-time) coordinates; (ii) Gaussian processes are embedded in a Bayesian framework;   (iii) the two approaches use different nomenclature~\cite{Genton02}.
The results of this paper are applicable to both frameworks.

\IEEEpubidadjcol
Covariance kernels for multidimensional input spaces often involve \emph{separable models} which are formed as products or as linear superposition of lower-dimensional kernels~\cite{Deiaco01,Duvenaud11,Wilson13}.
Separability is also invoked to construct simplified space-time covariance kernels by means of products or linear superposition of spatial and temporal components.  In separable models, spatial and temporal correlations are decoupled; this behavior is not physically justifiable~\cite{dth17} and can lead to numerical instabilities in calculations of conditional means and variances~\cite{Cressie99}.
Non-separable, flexible and physically motivated covariance kernels are thus in great demand~\cite{Cressie99,Gneiting02,Kolovos04}.  Cressie and Huang~\cite{Cressie99} constructed kernels that involve space and time interactions by inverting  admissible mathematical expressions for the spatial Fourier modes of the kernels.  Gneiting~\cite{Gneiting02} introduced a method for constructing admissible kernels directly in the space-time domain, thus avoiding the calculation of inverse Fourier transforms.  His method takes advantage of the powerful properties of completely monotone functions and generates a broad class of functions. Kolovos \emph{et al.}~\cite{Kolovos04} review various methods for generating non-separable space-time kernels including the use of Radon transforms and stochastic partial differential equations.

Properties, existing models, and open research questions for spatiotemporal covariance kernels are discussed in two recent reviews~\cite{Chen21,Porcu21}.  Ideally, space-time covariance kernels should be  solutions of  partial differential equations (PDEs) that characterize the particular system under study~\cite{dth20}.  However, such PDEs are not  amenable to explicit solutions except in certain special cases~\cite{Heine55,Jones97,Christakos98,dth17,Bakka20}. Hybrid approaches that use Gaussian processes and differential equations to combine data-driven modeling with a physical model have  been proposed to address this issue~\cite{Alvarez13}.

Motivated by the scarcity of solvable physics-based models for covariance kernels, we  derive   a new family of non-separable covariance kernels which are based on  the stochastic, linear, damped harmonic oscillator (LDHO).  The LDHO model is herein suitably generalized for spatially extended processes by means of the Cressie-Huang approach~\cite{Cressie99}.  The spatial LDHO Fourier modes are generated by injecting intuitive dispersion relations  in the oscillator's coefficients. The dispersion relations  then translate into physically meaningful covariance kernel hyperparameters. To our knowledge, this is the first non-separable space-time covariance kernel family in the literature which allows for oscillatory temporal correlations. Hence, the LDHO kernels are particularly interesting for geo-referenced data that exhibit periodicity (e.g., diurnal, weekly, monthly or yearly) in their correlation functions~\cite{dth19}.  The LDHO covariance kernels are also applicable to temporal Gaussian processes with different input spaces, so long as  the Euclidean distance measure is meaningful for the input space (excluding the time dimension).

The remainder of this paper has the following structure: Section~\ref{sec:methods} presents necessary notation and definitions.  Section~\ref{sec:ldho} focuses on the linear damped harmonic oscillator driven by white noise and the respective covariance kernels.  Section~\ref{sec:hybrid-spectral} introduces the hybrid spectral method for the construction of spatiotemporal covariance kernels.  In Section~\ref{sec:hybrid-spectral-ldho}, the hybrid spectral method is applied to the LDHO and spatiotemporal covariance kernels are derived for the three different oscillator regimes based on dispersion functions with $\Or(f(k^2))$ dependence.   Several  properties of the LDHO kernels are discussed in Section~\ref{sec:properties}. The hybrid spectral method is further illustrated in Section~\ref{sec:other-kernels}, where LDHO kernels are obtained for dispersion functions with $\Or(f(k))$ dependence, as well as kernels derived from the Ornstein-Uhlenbeck equation.
Finally, conclusions and directions for future research are given in~\ref{sec:conclusions}.  Long proofs are relayed to the Appendices, and additional material is presented in the online Supplement.
	
\section{Methods and Procedures}
\label{sec:methods}
	
\subsection{Notation and Definitions}
We use lowercase boldface symbols, e.g., $\mathbf{a}, \mathbf{b}$, to denote vector variables and uppercase boldfaced letters to represent matrices. 	The transpose of a matrix $\bfA$ is denoted by $\bfA^\top$, its inverse by
$\bfA^{-1}$, and the matrix determinant by $\det\bfA$. The dot defines the inner product of two vectors, i.e., $\mathbf{a} \cdot \mathbf{b} = \sum_{i=1}^{n} a_{i} b_{i}$, where $n$ is the dimension of vectors $\mathbf{a}, \mathbf{b}$.  The Euclidean norm of the vector $\mathbf{a}$ will be denoted by $\lVert \mathbf{a}\rVert$.

$\Na$ is the set of natural numbers,
$\R$ denotes the set of real numbers, $\R_{+}$ denotes the set of positive real numbers, and $\R_{+,0}$ the set of non-negative real numbers.  The zero vector in $\Rd$, where $d \in \Na$, is denoted by $\bfo$, i.e.,
$\bfo_{i} = 0$ for $i=1, \ldots, d$.
$\Co$ is the set of complex numbers. If $a \in \Co$ then $a=\Re(a) + \I \Im(a)$, where $\Re(a), \Im(a)$ are respectively the real and imaginary parts of $a$ and $\I=\sqrt{-1}$. The complex conjugate of $a \in \Co$ is denoted by $a^\dagger \in \Co$ and $\lvert a \rvert =(a\,a^\dagger)^{1/2}$ is the magnitude of $a$.  Finally, the symbol $\triangleq$ will be used for definitions.

\subsection{Gaussian processes and random fields}
A Gaussian process defines a prior distribution over functions, which can then be used for Bayesian regression~\cite{Neal98}. 	We will denote the \gp by $z(\bfx) \sim \gpm\left( m(\bfx), C(\bfx, \bfx')  \right)$, where $m(\bfx): \R^{D} \to \R$ is the mean function (expectation) and 	$C(\bfx, \bfx'): \R^{D} \times \R^{D} \to \R$ is the covariance kernel, which is a non-negative definite function~\cite{Rasmussen06}.  The matrix $\C$ with elements  $[\C]_{i,j} = C(\bfx_{i}, \bfx_{j})$, for all $i,j=1, \ldots, n$ 	(where $n \in \Na$) is the \emph{kernel covariance} (Gram) matrix.

For geo-referenced data $\bfx=(\bfs,t)$ where $\bfs \in \Do \subset \Rd$ is the spatial coordinate inside the spatial domain $\Do$ and $t \in \R_{+,0}$ is the time instant. In this case, the dimension of the input vector $\bfx$ is $D=d+1$. 	To be more precise, given a probability space $(\Omega, \mathcal{F}, \textsl{P})$, where ${\Omega}$ is the sample space, $\mathcal{F}$ is a $\sigma-$field of subspaces of ${\Omega}$, and $\textsl{P}$ is a probability measure, the collection of real-valued, scalar random variables  $ \{ z(\bfs,t;\om): \bfs \in \Do, \, t\in   \R_{+,0}, \, \om \in \Omega \}$ is a scalar, real-valued \emph{spatiotemporal process} $Z: \Do \times R_{+,0} \times \Omega \mapsto \R$.  The expectation operator over  the probability space is denoted
 by $\EE[\cdot]$. The functions (realizations) of this space are denoted by $z(\bfs,t)$.
Herein we focus on \emph{weakly (second-order) stationary} spatiotemporal processes which  have (i) constant mean and (ii) covariance that depends purely on the space-time lag, i.e.,
$C(\bfx_{1},\bfx_{2}) = C(\bfx_{1}-\bfx_{2})= C(\bfs_{1}-\bfs_{2}, t_{1}-t_{2})$.
We use $\bfr = \bfs_{i} - \bfs_{j}\in \Rd$ to denote the spatial lag and $\tau = t_{i}- t_{j}\in \R$ for the temporal lag  between  two space-time points $\bfx_{i}$ and $\bfx_{j}$.   The  indices $i,j$ are not needed for the space-time lags in the stationary case.

 \subsection{Covariance kernels}
	
\begin{definition}[Non-negative definite functions]
The function $C(\cdot, \cdot)$ is \emph{non-negative definite} (positive definite) if and only if for all sets $\{ \bfx_{i} \}_{i=1}^{n}$ and all real-valued vectors $\mathbf{v} \in \R^{n}$ it holds that $\mathbf{v}^\top \C \mathbf{v} \ge 0$
(resp.,  $\mathbf{v}^\top \C \mathbf{v} > 0$) for all $n \in \Na$  and $\mathbf{v} \neq \bfo$, where
$[\C]_{i,j} = C(\bfx_{i},\bfx_{j})$.
\end{definition}
\begin{rem}[Units]
Covariance kernels depend on a vector of hyperparameters $\bmthe$.  The box notation, $[\theta]$, denotes the units of a scalar hypeparameter $\theta$; e.g.,  $[\theta]=[L]/[T]$ implies that $\theta$ units of length over time.

\end{rem}

\smallskip
 \begin{definition}[Fourier transforms]
Let $C(\bfr,\tau): \Rd \times \R \to \R$ represent a space-time function which is absolutely integrable over $\Rd \times \R$. Then, the   \emph{Fourier transform} $\tC(\bfk,\omega) = \ft[ C(\bfr,\tau)]$ and its inverse $C(\bfr,\tau)  =\ift[ {\tC}(\bfk,\omega)]$ exist. The FT is given by means of the multidimensional improper integral
\begin{equation}
\label{eq:covft}
\tC(\bfk,\omega) = \int_{\Rd}  \int_{\R} \E^{-\I (\bfk\cdot \bfr +\omega \tau)} C(\bfr,\tau) \D\bfr \D\tau,
\end{equation}
where  $\omega \in \R$ is the \emph{cyclic frequency}, and $\bfk \in \Rd$ is the \emph{spatial frequency vector (wavevector)} in reciprocal (Fourier) space.  The Euclidean norm $\kk$  is known as the \emph{wavenumber}.

The inverse FT is given by means of the following integral
\begin{equation}
\label{eq:invcovft}
C(\bfr,\tau)  = \frac{1}{(2\,\pi)^{d+1}}\,\int_{\Rd}  \int_{\R}  \E^{\I (\bfk\cdot \bfr +\omega \tau)}  {\tC}(\bfk,\omega) \D\bfk \D\omega.
\end{equation}

\end{definition}

\begin{theorem}[Bochner's theorem]
\label{theo:bochner}
A function $C(\bfr,\tau)$ is an \emph{admissible covariance kernel} for a stationary random field if and only if the Fourier transform $\tC(\bfk,\omega)$ of $C(\bfr,\tau)$ exists, is non-negative,  and its integral over $\Rd \times \R$ is finite~\cite{Bochner59}.
\end{theorem}

\smallskip

Bochner's theorem specifies conditions for $C(\bfr,\tau)$ to be an admissible covariance kernel for some random field without requiring that the latter be normally distributed.
\smallskip

\begin{definition}[Radial functions]
\label{defi:radial}
A function $C(\bfr,\tau)$ is called a \emph{radial function} if $C(\bfr,\tau)=C_{r}(r,\tau)$ where $r=\rr$ is the Euclidean norm of $\bfr$. For simplicity of notation we drop the index $r$ in $C_{r}(r,\tau)$. A covariance kernel defined by a radial function is called \emph{isotropic}.
\end{definition}

The Fourier transform of a radial function $C(r,\tau)$, if it exists, is also a radial function $\tC(k,\omega)$, where $k=\kk$; the converse is also true.

\smallskip
\begin{definition}[Marginal covariance kernels]
\label{defi:marginal}
The functions $\Cms(r)\triangleq C(r,\tau=0)$ and $\Cmt(\tau)\triangleq C(r=0,\tau)$ represent  the spatial and temporal \emph{marginal covariance kernels} at zero space and time lags  respectively.
\end{definition}

\section{Covariance Kernel of Harmonic Oscillator Driven  by White Noise}
\label{sec:ldho}

We denote by $z(t;\om)$ the displacement from equilibrium of a classical, linear, damped harmonic oscillator (LDHO)  as a function of time  $t \in \R_{+,0}$. It is  assumed that
$m>0$ is the inertial mass of the oscillator, $\gamma>0$  is the friction coefficient, and $\kappa>0$ is Hooke's constant.

\subsection{Equation of motion}
The equation of motion (EOM) due to random forcing (e.g., if the oscillator is placed in a heat bath)  is given by the following stochastic ordinary differential equation (SODE)
\beq  \label{eq:ldho-sode}
	\frac{\d^{2}{z}(t;\om)}{\d t^{2}} + \frac{\gamma}{m} \frac{\d{z}(t;\om)}{\d t}  + \frac{\kappa}{m} z(t;\om) = \sigma_{\eta} \eta(t;\om).
\eeq

The function $\sigma_{\eta} \eta(t)$, where $\sigma_{\eta}>0$, models the random force acting on the oscillator. The
noise  $\eta(t;\om)$ represents a realization of a standard Gaussian white noise stochastic process,
i.e., $d\eta(t;\om)=dW(t;\om)$ where $dW(t;\om)$ is the differential of the \emph{Wiener process}; hence
\beq  \label{eq:whitenoise}
	\EE[\eta(t;\om)] = 0; \;
	\EE[ \eta(t;\om) \eta(t';\om)] =\delta(t-t'), \text{  for all } t, t', \enspace
\eeq
where $\delta(\cdot)$ is the Dirac delta function. The LDHO hyperparameters
$m, \kappa, \ga, \sigma_{\eta}$ can be replaced by the more intuitive quantities, $\omega_{0}$,  $\tau_{c}$, and $\sigma$ where
\[
\omega_0 \triangleq \sqrt{\kappa/m}
\]
is the \emph{natural frequency of the undamped oscillator}, and
\[
\tau_{c} \triangleq m/\gamma
\]
is the \emph{characteristic damping time} and $\sigma \triangleq \sigma_{\eta}/m$.
The  \emph{natural frequency of the damped oscillator} is given by
\beq
\label{eq:cyclical-frequency}
\omega_{d}=\sqrt{\kappa/m-\gamma^2/(4m^2)}= \sqrt{\omega_0^2 - 1/(4\tau_{c}^2)}.
\eeq
The damped frequency~\eqref{eq:cyclical-frequency} is real-valued if $\gamma^2 < 4m\kappa$; the value
$\gamma_{\mathrm{crit}} \triangleq 2(m\kappa)^{1/2}$ represents the critical damping.

\begin{rem}[RLC Oscillator]
The EOM~\eqref{eq:ldho-sode} for the LDHO is parametrized  for a mechanical oscillator. However, using the substitutions $m \to L$, $\gamma \to R$, and $\kappa \to 1/C$, the EOM describes current oscillations in an electrical RLC circuit in the presence of thermal noise.
\end{rem}

\subsection{Covariance equation of motion}
Since the oscillator displacement $z(t;\om)$ is governed by a second-order linear SODE, the displacement covariance $C(\tau) \triangleq \EE[z(t+\tau;\om)\,z(t;\om)]-\EE[z(t+\tau;\om)]\,\EE[z(t;\om)]$ is the fundamental solution (Green's function) of a fourth-order, linear  ordinary differential equation (ODE).

\smallskip

\begin{coro}[Green's function equivalence of LDHO Covariance]
\label{corol:green-ldho}
If the stochastic process $z(t;\om)$ is governed by the second-order SODE~\eqref{eq:ldho-sode}, its covariance kernel is the fundamental solution (Green's function); the latter satisfies the following fourth-order (biharmonic) generative ODE, where $\sigma = \sigma_{\eta}/m$:
\beq
\label{eq:ldho-cov-ode}
\frac{d^{4}C(\tau)}{d\tau^4}  + \left( 2 \,\omega_{0}^{2} - \frac{1}{\tau_{c}^{2}} \right) \frac{d^{2}C(\tau)}{d\tau^2} +\omega_{0}^{4}C(\tau) =\sigma^{2}_{\eta} \delta(\tau).
\eeq
\end{coro}

\begin{IEEEproof}
The proof is given in Appendix~\ref{app:ldho-ode}.
\end{IEEEproof}

\smallskip

The connection between covariance kernels of stochastic processes satisfying linear SODEs and Green's functions is well-known~\cite{Dolph52,Rue11,dth20}.
\medskip

\begin{coro}[Spectral density from generative ODE]
\label{corol:spd-ode}
The covariance kernel which satisfies the ODE~\eqref{eq:ldho-cov-ode} corresponds to a \emph{spectral density} $\tC(\omega)$. If $\tau_c >0$ the latter is given by the following rational function of the cyclic frequency $\omega$:
\beq
\label{eq:spd-ldho}
	\tC(\omega) = \frac{\sigma^{2}_{\eta}\,\tau_{c}^2 }{ \tau_{c}^2 (\omega^2 - \omega_0^2)^2 + \omega^2 }.
\eeq
\end{coro}

\begin{IEEEproof}
The \emph{spectral density} of  $z(t;\om)$ is obtained according to Bochner's theorem~\cite{Bochner59} from the Fourier transform of   $C(\tau)$.  We multiply  both sides of~\eqref{eq:ldho-cov-ode} in Corollary~\ref{corol:green-ldho} with $\tau_{c}^{2}$ and apply the Fourier transform.  Since  the image of the time derivative operator in the Fourier domain is  $\ft[\D/\D\tau] = \I\omega$~\cite{Trefethen05}, it follows that
\beq
\label{eq:ft-time-deriv}
\ft\left[\frac{\D^n C(\tau)}{\D\tau^n}\right] = (\I \omega)^{n} \, \tC(\omega), \; n \in \Na.
\eeq
The spectral density~\eqref{eq:spd-ldho} then follows by recalling that $\ft[\delta(\tau)]=1$.
\end{IEEEproof}

\smallskip

\begin{rem}[Admissibility of spectral density] The function
$\tC(\omega)$ defined in~\eqref{eq:spd-ldho} is demonstrably non-negative for all $\omega \in \R$ and integrable over $\omega \in \R$. Therefore, it satisfies Bochner's admissibility conditions.
\end{rem}


\subsection{Covariance kernel}
The covariance kernel for the LDHO is given by calculating the
inverse Fourier transform of the spectral density. This can be explicitly evaluated
as shown in~\cite{dthsel07} (with slightly different parametrization).  The results,
which correspond to three different LDHO damping regimes, are
reviewed below.

\smallskip

\subsubsection{Underdamping}
This regime is obtained for $\omega_d >0$, i.e., for $\omega_{0} \tau_{c} > 1/2$. In this case,
\begin{subequations}
\label{eq:cov-ldho}
\beq
\label{eq:cov-ldho-u}
C(\tau)  = \frac{\sigma^2}{2\omega_0^2 \,\tau_{c}}  \, \E^{ -\frac{\lvert \tau \rvert}{2\tau_{c}} } \, \left( \cos \omega_{d} \tau + \frac{\sin\omega_{d} \lvert \tau \rvert}{2\omega_{d}\tau_{c}}  \right)
\eeq
These covariance kernels oscillate  with amplitudes that decrease exponentially with characteristic time $2\tau_{c}$.

\smallskip

\subsubsection{Overdamping}
This regime is obtained for $\omega_d$ imaginary, i.e., for $\omega_{0} \tau_{c} < 1/2$.
\beq
\label{eq:cov-ldho-o}
C(\tau)  = \frac{\sigma^2}{4 \omega_0^2 \, \tau_{c}} \left( \frac{1 }{ \lvert\omega_{d}\rvert \tau_f } \E^{-\lvert \tau \rvert/\tau_s} -  \frac{1 }{ \lvert\omega_{d}\rvert \tau_s } \E^{-\lvert \tau \rvert/\tau_f}  \right)\,.
\eeq

Hence, the covariance kernel decays as a superposition of two exponential functions with two characteristic times, a slow time, $\tau_s$, and a fast time, $\tau_f$:
\beq
\label{eq:fast-slow}
\tau_s = \frac{2\tau_{c}}{1- 2\tau_{c} \lvert\omega_{d}\rvert}, \;
\tau_f = \frac{2\tau_{c}}{1 + 2\tau_{c} \lvert\omega_{d}\rvert}, \; \tau_s > \tau_f\,.
\eeq

\begin{rem}[Admissible difference of exponential kernels] Since $\omega_d$ is imaginary in this regime, it follows from~\eqref{eq:cyclical-frequency} that
$2 \omega_{0} \tau_{c}<1$ and  $2 \lvert\omega_{d}\rvert \tau_{c}=\sqrt{1 - (2\omega_{0}\tau_c)^2}$.  Hence, it holds that $0 < 2 \lvert\omega_{d}\rvert \tau_{c} <1$. Therefore, the slow time
$\tau_s$ is a positive number.  An interesting fact about the covariance~\eqref{eq:cov-ldho-o} is that it involves the difference of two admissible
(exponential) kernels, and it is admissible nonetheless.  While this may seem trivial, one needs to recall that there are no simple, general conditions that render a linear superposition of kernels admissible  unless the  coefficients of the superposition are non-negative~\cite{Ma05}.
\end{rem}

\smallskip

\subsubsection{Critical damping}
This regime is obtained for  $\omega_{d}=0$, i.e., for $\omega_{0} \tau_{c} = 1/2$.
\beq
\label{eq:cov-ldho-c}
C(\tau)  = \frac{\sigma^2}{ 2\omega_{0}^{2} \,\tau_{c}}\, \E^{ -\frac{\lvert \tau \rvert}{2\tau_{c}} } \, \left(  1 + \frac{\lvert \tau \rvert}{2 \tau_{c}}  \right).
\eeq
\end{subequations}
The function~\eqref{eq:cov-ldho-c} is also known as the modified exponential kernel~\cite{Spanos07,dth16}.

\smallskip


\section{Hybrid Spectral Approach for Spatiotemporal Kernel Construction}
\label{sec:hybrid-spectral}

In this section we present the hybrid spectral approach.
We assume that  $\tC(\bfk,\omega)$ is the  spectral density of a spatiotemporal kernel.
The space-time inverse Fourier transform of $\tC(\bfk,\omega)$ satisfies the
following decomposition property
\begin{align}
\label{eq:ift-decomposition}
C(\bfr, \tau ) \triangleq & \, \ift[\tC(\bfk,\omega)] =  \ift_{\bfk} \left[  \ift_{\omega} [\tC(\bfk,\omega)]  \right]
\nonumber \\
= & \ift_{\bfk} [  \tC_{-\omega}(\bfk,\tau)].\end{align}
In~\eqref{eq:ift-decomposition}, $\ift_{\omega}$ ($\ift_{\bfk}$) is the inverse Fourier transform with respect to the temporal (spatial) dimension,  and the function  $\tC_{-\omega}(\bfk,\tau) \triangleq  \ift_{\omega} [\tC(\bfk,\omega)]$ represents the \emph{temporal Fourier modes} of the covariance kernel.  The temporal modes are thus defined by means of the partial (with respect to $\omega$) inverse Fourier transform of $\tC(\bfk,\omega)$.

The hybrid spectral approach involves the following steps:
\begin{enumerate}[wide, labelwidth=!, labelindent=0pt]\itemsep0.5em
    \item \emph{Generative ODE:} A purely temporal covariance kernel, $C(\tau;\bmthe_{0})$, is derived as the fundamental solution (Green's function) of a generative linear ODE with constant coefficients given by the vector $\bfc(\bmthe_{0}) \triangleq\left(c_{1}(\bmthe_{0}), \ldots, c_{P}(\bmthe_{0})\right)^\top$, where  $\bmthe_{0} \in \R^{m}$ is a  hyperparameter vector. Thus, $C(\tau;\bmthe_{0})$ satisfies the following equation (in terms of the linear differential operator $ {\mathcal L}_{\tau}$)
    \beq
    \label{eq:cov-ode}
    {\mathcal L}_{\tau}C(\tau;\bmthe_{0})= \delta(\tau), \; {\mathcal L}_{\tau}=\sum_{p=0}^{P} c_{p}(\bmthe_{0}) \frac{\D^{2p}}{\D \tau^{2p}}.
    \eeq

    \item \emph{Spectral density:} The spectral density corresponding to $C(\tau;\bmthe_{0})$ is given by
    \begin{subequations}
    \label{eq:psd-ode}
     \beq
    \tC(\omega;\bmthe_{0})=  \frac{1}{\Pi(\omega;\bmthe_0)},
    \eeq

\noindent where   $\Pi(\omega;\bmthe_0)$ is the characteristic polynomial of the differential operator $ {\mathcal L}_{\tau}$ given by
    \beq
    \Pi(\omega;\bmthe_0) =  \sum_{p=0}^{P} (-1)^{p}\,c_{p}(\bmthe_{0}) \omega^{2p},  \; P \in \Na\,.
    \eeq
    \end{subequations}

The equations~\eqref{eq:psd-ode} are obtained by applying the Fourier transform on both sides of~\eqref{eq:cov-ode},  using~\eqref{eq:ft-time-deriv} to calculate the FT of derivatives.
Bochner's theorem requires that $c_{0}(\bmthe_0)>0$ and $\Pi(\omega;\bmthe_0)>0$ for all $\omega \in \R$.

\item  The generative ODE coefficients are replaced by the vector $\tilde{\bfc}(\bfk;\bmthe)$. The latter incorporates \emph{dispersion relations} which  implement the space-time interactions.
Respectively, the characteristic polynomial becomes
\beq
\label{eq:char-poly-dispersion}
\tilde{\Pi}(\omega,\bfk;\bmthe) =  \sum_{p=0}^{P} (-1)^{p}\,\tilde{c}_{p}(\bfk;\bmthe) \,\omega^{2p},  \; P \in \Na\,.
\eeq
The dispersion relations must be compatible with the admissibility conditions of Bochner's theorem~\ref{theo:bochner} which specify that: (i) $\tilde{\Pi}(\omega,\bfk;\bmthe) \ge 0$ for $\bfk \in \Rd$ and (ii) the integral of $\tC(\bfk,\omega)$  over $\Rd \times \R$ is finite.

\item The spectral density of the temporal process,  $\tC(\omega;\bmthe_{0})$, generates the  spectral density $\tC(\bfk,\omega;\bmthe)$ of the spatiotemporal kernel; the latter is obtained from $\tC(\omega;\bmthe_{0})$ by replacing $c_{p}(\bmthe_0)$ with $\tilde{c}_{p}(\bfk;\bmthe)$ for all $p=1, \ldots, P$.

\item \emph{Temporal Fourier modes} of the associated spatiotemporal kernel, $\tC_{-\omega}(\bfk,\tau;\bmthe)$ are obtained from $C(\tau;\bmthe_{0})$ in~\eqref{eq:psd-ode} by replacing  $\bfc(\bmthe_{0})$ with  $\bfk$-dependent coefficients     $\tilde{\bfc}(\bfk;\bmthe)$, where   $\bmthe=(\bmthe_{0}^\top, {\bmthe'}^\top)^\top$ is the augmented hyperparameter vector and $\bmthe' \in \R^{\ell}$ is the hyperparameter vector used to define the $\bfk$ dependence.   The temporal modes $\tC_{-\omega}(\bfk,\tau;\bmthe)$ are fundamental solutions of the \emph{$\bfk$-dependent generative ODEs:}
    \beq
    \label{eq:cov-ode-bfk}
    \tilde{\mathcal L}_{\tau}\tC_{-\omega}(\bfk,\tau;\bmthe)= \delta(\tau), \; \tilde{\mathcal L}_{\tau}=\sum_{p=1}^{P} \tilde{c}_{p}(\bfk;\bmthe) \frac{\D^{2p}}{\D \tau^{2p}} \,.
    \eeq

The fundamental solution for each $\bfk$ corresponds to a different coefficient vector $\tilde{\bfc}(\bfk;\bmthe)$.  The assumption underlying~\eqref{eq:cov-ode-bfk} is that the  mode for a given $\bfk$ evolves in time independently of the modes for $\bfk' \neq \bfk$.

\hspace{1em}Based on the decomposition property~\eqref{eq:ift-decomposition}, proving the integrability of $\tC(\bfk,\omega)$   over $\Rd \times \R$ is equivalent to proving that the modes $\tC_{-\omega}(\bfk,\tau)$  are integrable over $\Rd$.  This requires  showing that $\ift_{\bfk}[\tC_{-\omega}(\bfk,\tau)]$ exists and is not singular at $\bfr=\mathbf{0}$ (the lack of   singularity implies that $\tC_{-\omega}(\bfk,\tau)$ is integrable over $\bfk \in \Rd$).

\item If the temporal Fourier modes are  explicitly known by solving the generative ODE~\eqref{eq:cov-ode-bfk},  the space-time covariance kernel $C(\bfr,\tau)$ can be obtained, according to the decomposition property~\eqref{eq:ift-decomposition}, by calculating the IFT of $\tC(\bfk,\omega;\bmthe)$ with respect to the wavevector $\bfk$.  The latter is given by a multi-dimensional integral, which in certain cases can be exactly evaluated.

\end{enumerate}

The spatiotemporal  kernel $C(\bfr,\tau)$ is physically motivated if the generative ODE  that governs $\tC_{-\omega}(\bfk,\tau)$ is associated with a SODE that, at least approximately, describes the equation of motion (EOM) of the studied process~\cite{Dolph52,dth20}.

\section{Hybrid Spectral Approach Applied to the Harmonic Oscillator}
\label{sec:hybrid-spectral-ldho}

In the following, we suppress the kernel dependence on $\bmthe_{0}$ and $\bmthe$ for brevity.
We apply the hybrid spectral approach using the LDHO generative ODE given by~\eqref{eq:ldho-cov-ode} (Step~1 in Section~\ref{sec:hybrid-spectral}).  The associated LDHO spectral density is given by~\eqref{eq:spd-ldho} in Corollary~\ref{corol:spd-ode} (Step~2).
However, the spectral density~\eqref{eq:spd-ldho} involves the noise variance $\sigma^{2}_{\eta}$ instead of the coefficient $\sigma^2$ used in~\eqref{eq:cov-ldho}.
\begin{rem}[LDHO variance]
The variance $\sigma_{z}^2 \triangleq C(0)$ of the LDHO covariance in all three regimes is equal to $\sigma_{z}^2  = \sigma^{2}/2\omega_{0}^{2} \,\tau_{c}$. This is straightforward for the kernels~\eqref{eq:cov-ldho-u} and~\eqref{eq:cov-ldho-c}, while for the kernel~\eqref{eq:cov-ldho-o} it can be shown with simple algebraic manipulations. The variance $\sigma_{z}^2$ can also be evaluated by integrating the spectral density $\tC({\omega})$ over all $\omega$, i.e.,
$\sigma_{z}^2=\frac{1}{\pi}\int_{0}^{\infty} \d\omega \,\tC({\omega})$,  leading to
$\sigma_{z}^2=\frac{1}{2}\sigma_{\eta}^2\, \tau_{c}/\omega^{2}_{0}$. Equating the two expressions for the variance we obtain
$\sigma^{2}_{\eta}=\sigma^{2}/\tau_{c}^{2}$. Then, the LDHO spectral density becomes
\end{rem}
\smallskip
\beq
\label{eq:spd-ldho-2}
\tC(\omega) = \frac{\sigma^{2} }{ \tau_{c}^2 (\omega^2 - \omega_0^2)^2 + \omega^2 }\,,
\eeq
and the LDHO hyperparameter vector is  $\bmthe_{0} \triangleq \left(\sigma^{2}, \omega_{0},  \tau_{c}\right)^\top$.

\subsection{Dispersion relations}
\label{ssec:dispersion}
Dispersion relations  link  the LDHO hyperparameters with the spatial frequency $\bfk$.  Let us assume the following general form for the dispersion functions:
\begin{align}
\label{eq:dispersion}
& \sigma^{2} \to \sigma^{2}(\bfk) = \sigma_{0}^{2} A(\bfk), \, \tau_{c} \to \tau_{c}(\bfk)= \frac{\tilde{\tau}_{c}}{B(\bfk)}, \nonumber \\
& \omega_{0} \to \omega_{0}(\bfk)=\tilde{\omega}_{0}\, B(\bfk),
\end{align}
where $\tilde{\omega}_{0}, \tilde{\tau}_{c} \in \R_{+}$,  $A(\bfk), B(\bfk) >0$ for all $\bfk$. $A(\bfk)$ and $B(\bfk)$ are dimensionless functions that allow considerable flexibility.  We use physical considerations to constrain the form of these functions.  We postulate the following principles:

\begin{enumerate}[wide, labelwidth=!,labelindent=0pt]\itemsep=0.2em

\item[(i)] $A(\bfk)$ is a bounded and decreasing function of $\kk$, to ensure that the mode variance is finite and declines with increasing  $\kk$.  If $A(\bfk)$   increased with $\kk$, the temporal modes would not be integrable.  Non-exponential decline of $A(\bfk)$ is possible, but it is not sufficient to ensure integrability for all $d$ [cf. the dispersion functions~\eqref{eq:dispersion-fun} and the temporal Fourier modes given by~\eqref{eq:t-ft-modes-under}, \eqref{eq:t-ft-modes-over}, \eqref{eq:t-ft-modes-crit}].  Exponential decline of the mode variance suppresses the high-$\kk$ modes and ensures integrability.

\item[(ii)] $B(\bfk)$ increases with $\kk$, implying an increase of the mode frequency and simultaneous decline of the damping time.  Hence, for large $\kk$ (small spatial scales) the temporal mode oscillation frequency is high but the oscillations are rapidly damped. This behavior is combined with the fast decline of the oscillation amplitude  due to $A(\bfk)$.  Linking the oscillation frequency and damping time via  $B(\bfk)$ is crucial for mathematical convenience [cf. the comment accompanying~\eqref{eq:wd-mode} below]. The reverse dependence, i.e., a damping time that increases with $\kk$ and concomitant decrease of the oscillation frequency,  complicates the explicit integration of the temporal Fourier modes.

\item[(iii)] Arbitrarily and without loss of generality, we assume that
$A(\bfo)=B(\bfo)=1$ so that $\tau_{c}(\bfo)=\tilde{\tau}_{c}$,  $\omega_{0}(\bfo)=\tilde{\omega}_{0}$ and
$\sigma(\bfo)=\sigma_{0}$.  Different values for $A(\bfo)$ and $B(\bfo)$ can be absorbed in $\sigma_{0}$ and $\tilde{\omega}_{0}$.
\end{enumerate}

\medskip
In Step~3 of Section~\ref{sec:hybrid-spectral},  $\bmthe_{0}$ is augmented by the vector of the dispersion hyperparameters
$\bmthe' \triangleq (b, \epsilon)^\top$ (see below).
According to Step~4, inserting the dispersion relations~\eqref{eq:dispersion} in the spectral density~\eqref{eq:spd-ldho-2} modifies the latter  as follows
\beq
\label{eq:ldho-dispersion}
\tC(\bfk,\omega)=  \frac{\sigma_{0}^{2} A(\bfk) }{ \omega^2  +  \left[\,\omega^2 - \tilde{\omega}_{0}^{2}\, B^{2}(\bfk) \,\right]^2 \tilde{\tau}_{c}^2 /B^{2}(\bfk) } \,.
\eeq
The function~\eqref{eq:ldho-dispersion} satisfies by construction $\tC(\bfk,\omega) \ge 0$ for all $\omega \in \R$ and $\bfk \in \Rd$.  Hence, to confirm that~\eqref{eq:ldho-dispersion}  is an admissible spectral density for a stationary process (according to Bochner's theorem) it suffices to provide conditions on $A(\bfk)$ and $B(\bfk)$  that render $\tC(\bfk,\omega)$  integrable over $\Rd \times \R$.   Integrability conditions for radial dispersion functions are formulated in Section~\ref{ssec:radial-dispersion}.

The dispersive relations~\eqref{eq:dispersion} lead to  scaling relations for  the damped natural frequency and for certain hyperparameter combinations that appear in the Fourier modes:
\begin{subequations}
\label{eq:scaling}
\begin{align}
\label{eq:wd}
& \omega^{2}_{0}(\bfk)\tau_{c}(\bfk)  = \tilde{\omega}_{0}^{2} \,\tilde{\tau}_{c} B(\bfk),
\\[1ex]
& \omega_{d}(\bfk)\tau_{c}(\bfk)  = \tilde{\tau}_{c} \tilde{\omega}_{d} ,
\\[1ex]
\label{eq:scaling-d}
& \frac{\sigma^{2}(\bfk)}{\omega^{2}_{0}(\bfk)\tau_{c}(\bfk)}= \frac{\sigma_{0}^{2}}{\tilde{\omega}_{0}^{2} \,\tilde{\tau}_{c}}\frac{A(\bfk) }{B(\bfk)} \, .
\end{align}
Tethering the dispersion relations for $\omega_{0}(\bfk)$ and  $\tau_{c}(\bfk)$ to the same dispersion function in~\eqref{eq:dispersion}, i.e.,
$\omega_{0}(\bfk) \propto B(\bfk)$ and $\tau_{c}(\bfk) \propto B^{-1}(\bfk)$, enforces the linear dependence of the damped frequency on $B(\bfk)$:
\begin{align}
\label{eq:wd-mode}
& \omega_{d}(\bfk)  = \tilde{\omega}_{d} B(\bfk), \, \text{where} \; \tilde{\omega}_{d} = \frac{1}{2\tilde{\tau}_{c}}\,\Big\lvert \sqrt{4\tilde{\tau}_{c}^2\tilde{\omega}_{0}^{2} - 1} \, \Big\rvert \,.
\end{align}
\end{subequations}
This is a key property, since  in combination with $B(\bfk)>0$ it ensures that the LDHO regime is  determined by $\tilde{\omega}_{d}$, and thus it is invariant for all $\bfk \in \Rd$.

\subsection{Radial dispersion functions}
\label{ssec:radial-dispersion}

\begin{definition}[Dispersion functions with $k^2$ dependence]
Let $A(\bfk)$ and $B(\bfk)$ be given by the following radial dispersion functions, where $k=\kk$ is the wavenumber:
\begin{subequations}
\label{eq:dispersion-fun}
\begin{align}
\label{eq:B1}
B(k)= & 1 + b k^{2}, \; b > 0,
\\
\label{eq:A1}
A(k) = & \E^{-\epsilon k^{2}} \, B(k), \; \epsilon > 0.
\end{align}
\end{subequations}
\end{definition}
The function $B(k)$ in~\eqref{eq:B1} implies that  $\tau_{c}(k) \sim k^{-2}$,  whereas  $\omega_{0}(k) \sim k^{2}$ for $k \to \infty$.  The function $A(k)$, as defined in~\eqref{eq:A1},   is dominated by the square exponential decay, implying a rapid decrease of the  modal variance $\sigma^{2}(\bfk)$ for $k \to \infty$.  The rapid decay (different forms than the square exponential law are possible)  ensures integrability of the spectral density.  The choice $A(k)=\E^{-\epsilon k^2}B(k)$ simplifies the scaling relation~\eqref{eq:scaling-d} because it leads to  $A(k)/B(k)=\E^{-k^2}$.

The spectral density of the spatiotemporal LDHO  kernel is obtained by inserting in~\eqref{eq:ldho-dispersion} the scaling relations~\eqref{eq:scaling} and the radial dispersion functions~\eqref{eq:dispersion-fun}, leading to
\beq
\label{eq:spd-ldho-st-1}
\tC(k,\omega)=  \frac{\sigma_{0}^{2} \, \left( 1 + b\, k^2\right) \,\E^{-\epsilon k^{2}}}{ \omega^2 + \left[\omega^2 - \tilde{\omega}_{0}^{2}\, \left( 1 + b k^{2} \right)^2 \right]^2 \, \frac{\tilde{\tau}_{c}^2}{(1+bk^2)^2}} \,.
\eeq
Recalling Step~5, a comparison of~\eqref{eq:spd-ldho-st-1} and~\eqref{eq:spd-ldho}  leads to the conclusion that $\tC_{-\omega}(k,\tau)$ is given by equations~\eqref{eq:cov-ldho} with the substitutions ${\omega}_{0}^{2} \to \tilde{\omega}_{0}^{2} \, (1+bk^{2})^{2}$, $\tau_{c}^{2} \to \tau_{c}^{2}/(1+bk^2)^2$, and $\sigma^{2} \to \sigma_{0}^{2} \,(1+bk^{2})\,\exp(-\epsilon \,k^{2})$.

\subsection{LDHO Covariance Kernels}
The temporal Fourier modes $\tC_{-\omega}(\bfk,\tau)$  are obtained from the respective temporal kernels~\eqref{eq:cov-ldho} by replacing the
LDHO hyperparameters with the dispersion relations~\eqref{eq:dispersion} and~\eqref{eq:dispersion-fun} (Step~5 in Section~\ref{sec:hybrid-spectral}). The radial functions $A(k), B(k)$  allow the evaluation of the IFT of $\tC_{-\omega}(k,\tau)$ (Step~6 in Section~\ref{sec:hybrid-spectral}) leading to isotropic LDHO covariance kernels. The latter are given by
\beq
\label{eq:ift-fourier-mode}
C(r,\tau)=\ift_{\bfk}[\tC_{-\omega}(\bfk,\tau)],
\eeq
according to the spatiotemporal Fourier transform decomposition property~\eqref{eq:ift-decomposition}.

\paragraph*{Spectral representation of radial functions}
For radial covariance functions, the pair of spatial
Fourier transforms  is expressed in terms of the following, one-dimensional, improper integrals~\cite[p. 353]{Yaglom87}
\begin{subequations}
\begin{equation}
    \label{eq:covft-iso}
    \tC_{-\omega}({k},\tau)= \frac{(2\pi)^{d/2}}{k^{\nu}}  \int_{0}^{\infty}  r^{d/2}
    {J_{\nu}(k  r)} \, C({r},\tau) \,\D r,
\end{equation}
\begin{equation}
    \label{eq:invcovft-iso}
    C({r},\tau)= \frac{1}{(2\pi)^{d/2} r^{\nu}} \int_{0}^{\infty}  k^{d/2}
    {J_{\nu}(k  r)}  \, \tC_{-\omega}({k},\tau)\, \D k,
\end{equation}
\end{subequations}
where $r=\rr$, $k=\kk$, $\nu = d/2-1$, and $J_{\nu}(\cdot)$ is the \emph{Bessel function of the first kind of
order} $\nu$~\cite{Schoenberg38}.

The resulting spatiotemporal LDHO covariance kernels for each regime are presented below.  The proofs as well as the respective expressions for the temporal Fourier modes are given in Appendix~\ref{app:kernel-ldho-squared}.  For notation  convenience the hyperparameter $c_{0} \triangleq {\sigma^{2}_{0}}/{2\tilde{\tau}_{c} \tilde{\omega}^{2}_{0}}$ is introduced.

\medskip

\subsubsection{Underdamping}
This regime  is obtained for $\tilde{\omega}_{0} \tilde{\tau}_{c}>1/2$.

\smallskip

\begin{theorem}[LDHO  kernel in underdamped regime]
\label{theo:ldho-kernel-st-under}
If $\tilde{\omega}_{0} \tilde{\tau}_{c}>1/2$, the LDHO spatiotemporal kernel generated by the radial spectral density~\eqref{eq:spd-ldho-st-1}  is given  by
the radial function $C(r,\tau)$:
\begin{subequations}
\label{eq:ldho-kernel-st-under}
\begin{align}
\label{eq:ldho-kernel-st-under-C}
& C(r,\tau)  =  c_{0}  \, \E^{-\frac{\lvert \tau\rvert}{2\tilde{\tau}_{c}}} \left[ F_{1}(r,\tau) + F_{2}(r,\tau)\right],
\\[1ex]
& F_{1}(r,\tau)   = \cos(\tilde{\omega}_{d}\tau) \, \Rei(r,\tau) - \sin(\tilde{\omega}_{d}\lvert\tau\rvert) \, \Imi(r,\tau),
\nonumber \\[1ex]
& F_{2}(r,\tau)  = \frac{\sin(\tilde{\omega}_{d}\lvert\tau\rvert )}{2\,\tilde{\omega}_{d} \tilde{\tau}_{c}} \, \Rei(r,\tau) + \frac{\cos(\tilde{\omega}_{d}\lvert\tau\rvert )}{2\,\tilde{\omega}_{d} \tilde{\tau}_{c}} \, \Imi(r,\tau)\, ,
\nonumber \\
& \Rei(r,\tau)=  \frac{\E^{-\lambda^2 r^2} \,  \cos\left(\kappa^{2}r^2 + \tfrac{d\phi}{2} \right)}{(4\pi)^{d/2}\, \left[ \left( \epsilon + \lvert \tau\rvert\,  b / 2\tilde{\tau}_{c}\right)^{2} + b^2\,\tilde{\omega}_{d}^{2}\,\lvert\tau\rvert^{2} \right]^{d/4}} ,
\nonumber \\[1ex]
& \Imi(r,\tau)= \frac{\E^{-\lambda^2 r^2} \,  \sin\left(\kappa^{2}r^2 + \tfrac{d\phi}{2} \right)}{(4\pi)^{d/2}\,  \left[ \left( \epsilon + \lvert \tau\rvert\,  b / 2\tilde{\tau}_{c}\right)^{2} + b^2\,\tilde{\omega}_{d}^{2}\,\lvert\tau\rvert^{2} \right]^{d/4}} \,,
\nonumber
\end{align}
where   $r, \tau$ are, respectively, the spatial and  temporal lags. The quantities $\kappa^2$, $\lambda^2$ and $\phi$ (the dependence on  $\tau$ is suppressed for brevity) are \emph{space-time interaction functions} given by
\beq
\label{eq:kappa2-under}
\kappa^2 =
\frac{b\,\tilde{\omega}_{d}\,\lvert\tau\rvert}{  \left( \frac{b\lvert\tau\rvert}{\tilde{\tau}_{c}} +2\epsilon  \right)^{2} + \left(2b\,\tilde{\omega}_{d}\,\lvert\tau\rvert\right)^{2}}\,,
\eeq

\beq
\label{eq:lambda2-under}
\lambda^{2} = \frac{\frac{b\lvert\tau\rvert}{2\tilde{\tau}_{c}}+\epsilon}{ \left( \frac{b\lvert\tau\rvert}{\tilde{\tau}_{c}} +2\epsilon  \right)^{2} + \left(2b\,\tilde{\omega}_{d}\,\lvert\tau\rvert\right)^{2}}\,,
\eeq

\beq
\label{eq:theta-under}
\phi = \atan\left( \frac{-2b\,\tilde{\omega}_{d}\,\lvert\tau\rvert\tilde{\tau}_{c}}{b\,\lvert\tau\rvert+
2\epsilon \tilde{\tau}_{c}} \right)\,.
\eeq

\end{subequations}

\end{theorem}

\smallskip

\begin{IEEEproof}
The proof is given in Appendix~\ref{app:under}.
\end{IEEEproof}
\medskip

\paragraph{The LDHO model hyperparameters}
The kernel function~\eqref{eq:ldho-kernel-st-under}  includes a hyperparameter vector with five independent  components: $\bmthe = \left( c_{0}, \tilde{\tau}_{c}, \tilde{\omega}_{d},  \epsilon, b \right)^\top$.
The physical significance of the hyperparameters is as follows:

\begin{itemize}
\item $c_{0}$ is a  scaling factor which has units  $[z]^{2}$ and determines the kernel's amplitude.
\item $\tilde{\tau}_{c}$ is a characteristic relaxation time that controls the temporal decay of the correlations.
\item $\tilde{\omega}_{d}$ is a cyclical frequency which controls the periodicity  of damped temporal oscillations.
\item $\epsilon$  is the variance decay scale; it has dimensions of square length and determines how fast the mode variance declines at large $k$.
\item $b$ controls the space-time interaction strength; it also has dimensions of square length and determines the rate at  which the non-damped resonance frequency increases  and the damping time drops with $k$.
\end{itemize}

The hyperparameters $\tilde{\tau}_{c}$, $\tilde{\omega}_{d}$ control the purely temporal dependence  of the LDHO kernel. The hyperparameters  $b, \epsilon$ enter, along with $\tilde{\tau}_{c}$ and $\tilde{\omega}_{d}$, in the three time-dependent functions $\kappa^2, \lambda^2, \phi$  that control the space-time interactions as follows:

\begin{itemize}

\item $\kappa$: wavenumber that controls the spatial  oscillations of the LDHO kernel;

\item $\lambda$:  inverse length controlling the decay of spatial correlations;

\item $\phi$: phase factor modulating the correlations at $r=0$.

\end{itemize}

 A preliminary discussion of the estimation of $\bmthe$ from data is given  in the Supplement (Section~S1).

\smallskip

\begin{rem}[Kernel dependence on $d$]
$C(r,\tau)$ depends on the spatial dimension $d$ via the scaling factor $(4\pi)^{-d/2}$, the phase factor $d\phi/2$, and the denominators in the damped oscillatory functions $\Rei(r,\tau)$ and $\Imi(r,\tau)$.
\end{rem}

\begin{figure}
\centering
\includegraphics[width=.99\linewidth]{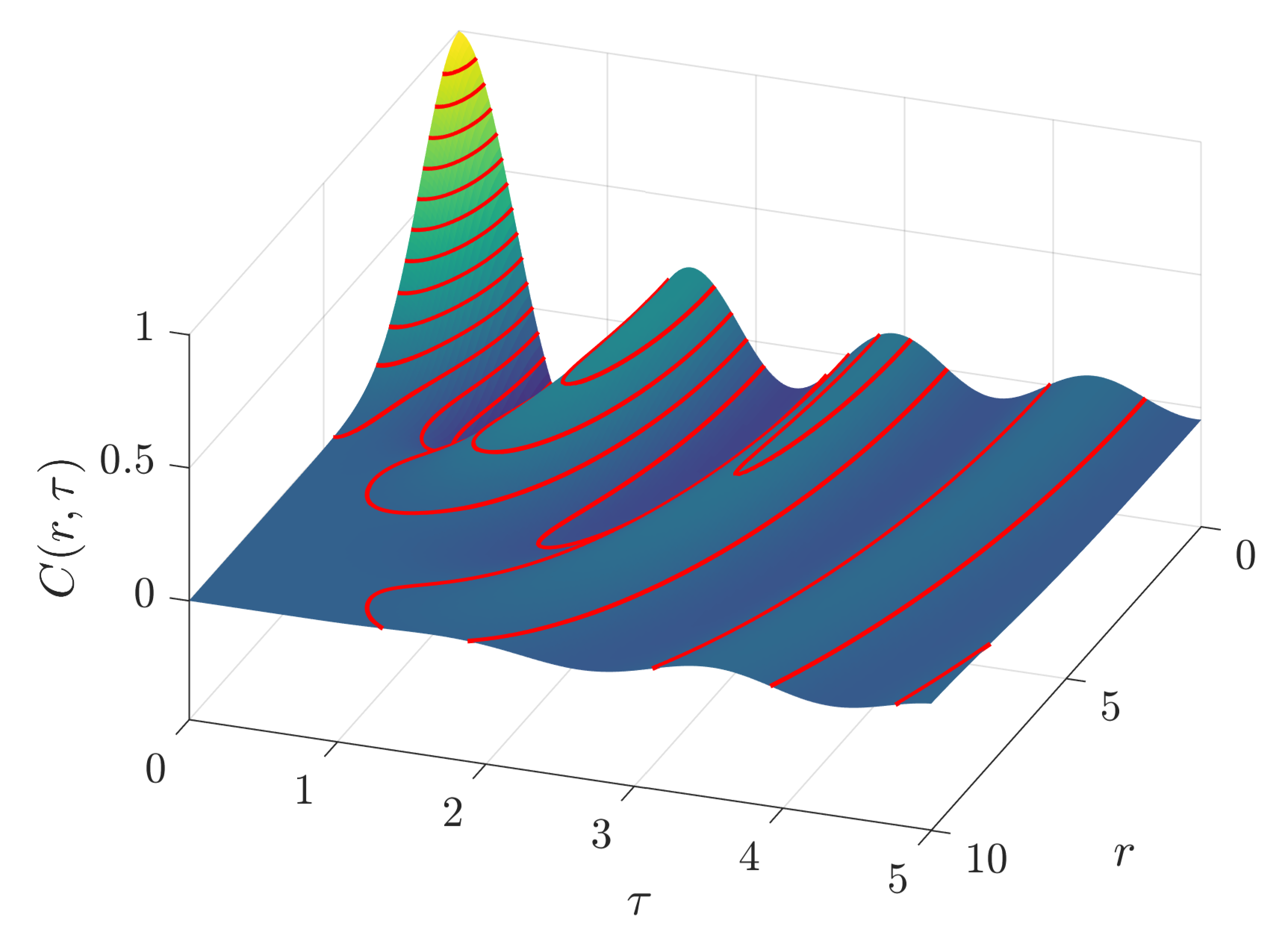}
\caption{Normalized $C(r,\tau)$ and isolevel contour lines (red online) in the underdamped regime, obtained from~\eqref{eq:ldho-kernel-st-under} using
$\tilde{\omega}_{d}=3\pi/2$, $\tilde{\tau}_{c}=3$, $b=0.4$,  $\epsilon=1$, and $d=2$.}
\label{fig:kernel_under_1}
\end{figure}

The LDHO covariance kernel is illustrated in Fig.~\ref{fig:kernel_under_1}.
A combination of a relatively slow damping time $\tilde{\tau}_{c}=3$ and a fast oscillation frequency, $\tilde{\omega}_{d}=3\pi/2$, generate four ridges with  decaying amplitude as $\tau$ increases. A smaller value of $\tilde{\tau}_{c}$ (not shown) leads to faster decay and fewer oscillation peaks. The plot also exhibits spatiotemporal interaction, i.e.,  spatial oscillation patterns that appear as ripples on the $(r,\tau)$ plane.

\begin{rem}[Variance decay scale]
The variance decay scale  $\epsilon$ crucially ensures that
the space-time interaction functions~\eqref{eq:kappa2-under}-\eqref{eq:theta-under} are stable at $\tau=0$: $\kappa^{2}(\tau=0)=0$, $\lambda^{2}(\tau=0)=1/4\epsilon$, and
 $\phi(\tau=0)=0$. These relations guarantee that $C(r, \tau=0)$ is finite.  In contrast, if $\epsilon=0$,  the limit of $C(r, \tau)$ as $\tau \to 0$ does not exist.
\end{rem}

\begin{rem}[Interaction strength]
The space-time interactions in~\eqref{eq:ldho-kernel-st-under} are controlled by the \emph{interaction hyperparameter} $b$.  Indeed, if $b=0$,
then~\eqref{eq:kappa2-under} implies that $\kappa=0$, $\lambda = 1/2\sqrt{\epsilon}$, and $\phi=0$. Thus, all three interaction functions become independent of $\tau$ for $b=0$.
\end{rem}

\paragraph{Zero-lag marginal covariances}  We evaluate the  marginal  kernels according to  Definition~\ref{defi:marginal}.

\begin{propo}[Spatial marginal covariance]
In the underdamped regime, the  spatial marginal covariance of the LDHO kernel~\eqref{eq:ldho-kernel-st-under}  at $\tau=0$ is given by the square exponential kernel
\beq
\label{eq:ldho-under-marginal-r}
\Cms(r) = c_{0}  \, \frac{\E^{-r^2/4\epsilon}}{(4\pi \epsilon)^{d/2}}.
\eeq
\end{propo}

\begin{IEEEproof}
From~\eqref{eq:ldho-kernel-st-under} for $\tau=0$ it follows that
$\Cms(r)= c_{0} \left[ F_{1}(r,0)+ F_{2}(r,0)\right]$. Furthermore,
$F_{1}(r,0)=\Rei(r,0)$ and $F_{2}(r,0)=\frac{1}{2\,\tilde{\omega}_{d} \tilde{\tau}_{c}}\Imi(r,0)$.
At zero time lag the space-time interaction functions take the following values:
\beq
\label{eq:st-interaction}
\kappa^{2}(\tau=0)=0, \; \lambda^{2}(\tau=0)=1/4\epsilon, \; \phi(\tau=0)=0.
\eeq
Hence, $\Rei(r,0)=\exp(-\lambda^{2}r^{2})/(4\pi \epsilon)^{d/2}$ while $\Imi(r,0)=0$. This concludes the proof.
\end{IEEEproof}

The result~\eqref{eq:ldho-under-marginal-r} shows that the hyperparameter $\epsilon$ can be viewed as
$\epsilon \triangleq \xi^{-2}/4$, where $\xi$ is the  correlation length of the square exponential kernel.

\begin{propo}[Temporal marginal covariance]
In the underdamped regime, the temporal marginal covariance of the LDHO kernel~\eqref{eq:ldho-kernel-st-under} at  $r=0$ is given by
the following damped harmonic expression---where $\phi$ is defined in~\eqref{eq:theta-under}:
\begin{align}
\label{eq:ldho-under-marginal-t}
& \Cmt(\tau) =  \frac{c_{0}\, \E^{-\frac{\lvert \tau\rvert}{2\tilde{\tau}_{c}}}}{(4\pi \epsilon)^{d/2}\, \left[ \left( \frac{b\lvert\tau\rvert}{2\tilde{\tau}_{c}\epsilon} + 1 \right)^{2} + \left(\frac{b\,\tilde{\omega}_{d}\,\lvert\tau\rvert}{\epsilon}\right)^{2} \right]^{d/4}}  \, 
\nonumber \\
& \quad \quad \times \left[  \cos\left(\tilde{\omega}_{d}\lvert \tau \rvert + \tfrac{d\phi}{2}\right)
+ \tfrac{1}{2\,\tilde{\omega}_{d} \tilde{\tau}_{c}}
\sin\left(\tilde{\omega}_{d}\lvert \tau \rvert + \tfrac{d\phi}{2}\right) \right].
\end{align}
\end{propo}

\begin{IEEEproof}
The functions $\Rei(0,\tau)$ and
$\Imi(0,\tau)$ are obtained from~\eqref{eq:ldho-kernel-st-under} by setting $r=0$. The rest follows from the definitions of $F_{1}(r,\tau)$ and $F_{2}(r,\tau)$, see~\eqref{eq:ldho-kernel-st-under},  using the trigonometric identities $\cos(\alpha+\beta)=\cos \alpha \cos \beta-\sin \alpha \sin \beta$ and
$\sin(\alpha+\beta)=\sin \alpha \cos \beta +\sin \beta \cos \alpha$.
\end{IEEEproof}

\medskip
\begin{propo}[Very large relaxation time limit]
Let $C_{\ast}(r,\tau)=\lim_{\tau_{c} \to \infty} C(r,\tau)$ be  the very-large-relaxation-time (VLRT) limit $\tau_c \to \infty$ of the underdamped LDHO kernel. Then, $C_{\ast}(r,\tau)$ is given by
\begin{subequations}
\beq
\label{eq:ldho-kernel-st-under-tauc-infty}
C_{\ast}(r,\tau)  =  \frac{c_{0} \E^{-\lambda_{0}^{2} \,r^2} \, \cos\left( \tilde{\omega}_{0}\tau + \kappa_{0}^{2} \,r^2 + \tfrac{d\phi_{0}}{2}\right)}{(4\pi)^{d/2}\, \left(  \epsilon^{2} + b^2\,\tilde{\omega}_{0}^{2}\,\lvert\tau\rvert^{2} \right)^{d/4}}
\eeq
\beq
\label{eq:kappa2-under-0}
\kappa_{0}^{2} =
\frac{b\,\tilde{\omega}_{0}\,\lvert\tau\rvert}{  4 \left(\epsilon^{2} +  b^{2}\,\tilde{\omega}_{0}^{2}\,\lvert\tau\rvert^{2}\right)} \,,
\eeq

\beq
\label{eq:lambda2-under-0}
\lambda^{2}_{0} = \frac{\epsilon}{  4 \left(\epsilon^{2} +  b^{2}\,\tilde{\omega}_{0}^{2}\,\lvert\tau\rvert^{2}\right)}\,,
\eeq

\beq
\label{eq:theta-under-0}
\phi_{0} = \atan\left( \frac{-b\,\tilde{\omega}_{0}\,\lvert\tau\rvert}{\epsilon} \right)\,.
\eeq

\end{subequations}

\end{propo}

\begin{IEEEproof}
Based on~\eqref{eq:wd} it holds that $\lim_{\tau_c \to \infty}\tilde{\omega}_{d} = \tilde{\omega}_{0}$. Equations~\eqref{eq:kappa2-under-0}-\eqref{eq:theta-under-0} are obtained from~\eqref{eq:kappa2-under}-\eqref{eq:theta-under} at the VLRT limit $\tau_{c} \to \infty$.  The kernel $C(r,\tau)$ is given by~\eqref{eq:ldho-kernel-st-under}; it is obvious that $\lim_{\tau_c \to \infty} F_{2}(r,\tau)=0$. Evaluating the VLRT limit of the first term, $F_{1}(r,\tau)$ at $\tau \to \tau_c$, the following is obtained
\begin{align*}
& C_{\ast}(r,\tau)  =  c_{0}  \,  \left[ \cos(\tilde{\omega}_{0}\tau) \, \Rei^\ast(r,\tau) - \sin(\tilde{\omega}_{0}\lvert\tau\rvert) \, \Imi^\ast(r,\tau)\right],
\\
& \Rei^{\ast}(r,\tau)=  \frac{\E^{-\lambda_{0}^{2} \,r^2} \,  \cos\left(\kappa_{0}^{2} \,r^2 + \tfrac{d\phi_{0}}{2} \right)}{(4\pi)^{d/2}\, \left(  \epsilon^{2} + b^2\,\tilde{\omega}_{0}^{2}\,\lvert\tau\rvert^{2} \right)^{d/4}} ,
\nonumber \\[1ex]
& \Imi^{\ast}(r,\tau)= \frac{\E^{-\lambda_{0}^2 \, r^2} \,  \sin\left(\kappa_{0}^{2} \,r^2 + \tfrac{d\phi_{0}}{2} \right)}{(4\pi)^{d/2}\,  \left(  \epsilon^{2} + b^2\,\tilde{\omega}_{0}^{2}\,\lvert\tau\rvert^{2} \right)^{d/4}} \,,
\nonumber
\end{align*}
where $\Rei^{\ast}(r,\tau)$ and $\Imi^{\ast}(r,\tau)$ are respectively the VLRT limits of $\Rei(r,\tau)$ and $\Imi(r,\tau)$ as $\tau_c \to \infty$, while $\kappa_{0}^{2}, \lambda_{0}^{2}, \phi_{0}$ in~\eqref{eq:kappa2-under-0}-\eqref{eq:theta-under-0}, are the limits of the respective functions as $\tau_c \to \infty$. Finally, the VLRT limit  $C_{\ast}(r,\tau)$ is obtained from the above equations using the trigonometric identity
$\cos(u+v)=\cos u \cos v - \sin u \sin v$, where $u,v \in \R$.
\end{IEEEproof}

\smallskip

\begin{rem}[Persistence of quasi-periodicity]
Even at the VLRT limit, the kernel $C_{\ast}(r,\tau)$ is not purely periodic due to the space-time interaction parameter $b$. The dispersion relation $\tau_{c}(\bfk)=\tilde{\tau}_{c}/(1 + bk^2)$ implies that even for large $\tilde{\tau}_{c}$, there exist $k \in \R$ such that $\tau_{c}(\bfk)$ is finite.  However, if $b=0$, i.e., for constant $\tau_{c}(\bfk)$, the LDHO kernel decouples in the VLRT limit into a product of separable spatial and temporal components; the latter is given by the purely periodic function $\cos(\tilde{\omega}_{0} \tau)$.
\end{rem}

\medskip

\subsubsection{Overdamping}
This regime  is obtained for $\tilde{\omega}_{0} \tilde{\tau}_{c}<1/2$.

\begin{theorem}[LDHO kernel in overdamped regime]
\label{theo:ldho-kernel-st-over}
If $\tilde{\omega}_{0} \tilde{\tau}_{c}<1/2$, the LDHO spatiotemporal kernel generated by the radial spectral density~\eqref{eq:spd-ldho-st-1}   is given by
\begin{align}
\label{eq:ldho-kernel-st-over}
C(r,\tau)= &  \frac{c_{0}\tilde{\tau}_{c}^{d/2-1}}{4\tilde{\omega}_{d}  } \left[ \frac{\beta_{f} \, \E^{-\frac{\beta_{s} \lvert \tau \rvert}{2\tilde{\tau}_{c}}-\frac{r^{2}\tilde{\tau}_{c} }{2b\lvert \tau \rvert \, \beta_{s}+ 4 \epsilon \tilde{\tau}_{c}}}}{ \left(2\pi b \beta_{s} \lvert \tau \rvert + 4\pi \epsilon \tilde{\tau}_{c}\right)^{d/2}} \right.
\nonumber \\
 &  \left. \quad\quad\quad -   \frac{\beta_{s} \, \E^{-\frac{\beta_{f} \lvert \tau \rvert}{2\tilde{\tau}_{c}}-\frac{r^{2}\tilde{\tau}_{c} }{2b\lvert \tau \rvert \beta_{f}+ 4 \epsilon \tilde{\tau}_{c}}}}{ \left(2\pi b \, \beta_{f} \lvert \tau \rvert + 4\pi \epsilon \tilde{\tau}_{c}\right)^{d/2}} \,  \right],
 \\[1ex]
\text{where} & \;  \beta_{s}=1- 2\tilde{\tau}_{c} \tilde{\omega}_{d}, \; \beta_{f}=1 + 2\tilde{\tau}_{c} \tilde{\omega}_{d}. \nonumber
\end{align}
\end{theorem}

\begin{figure}
\centering
\includegraphics[width=.99\linewidth]{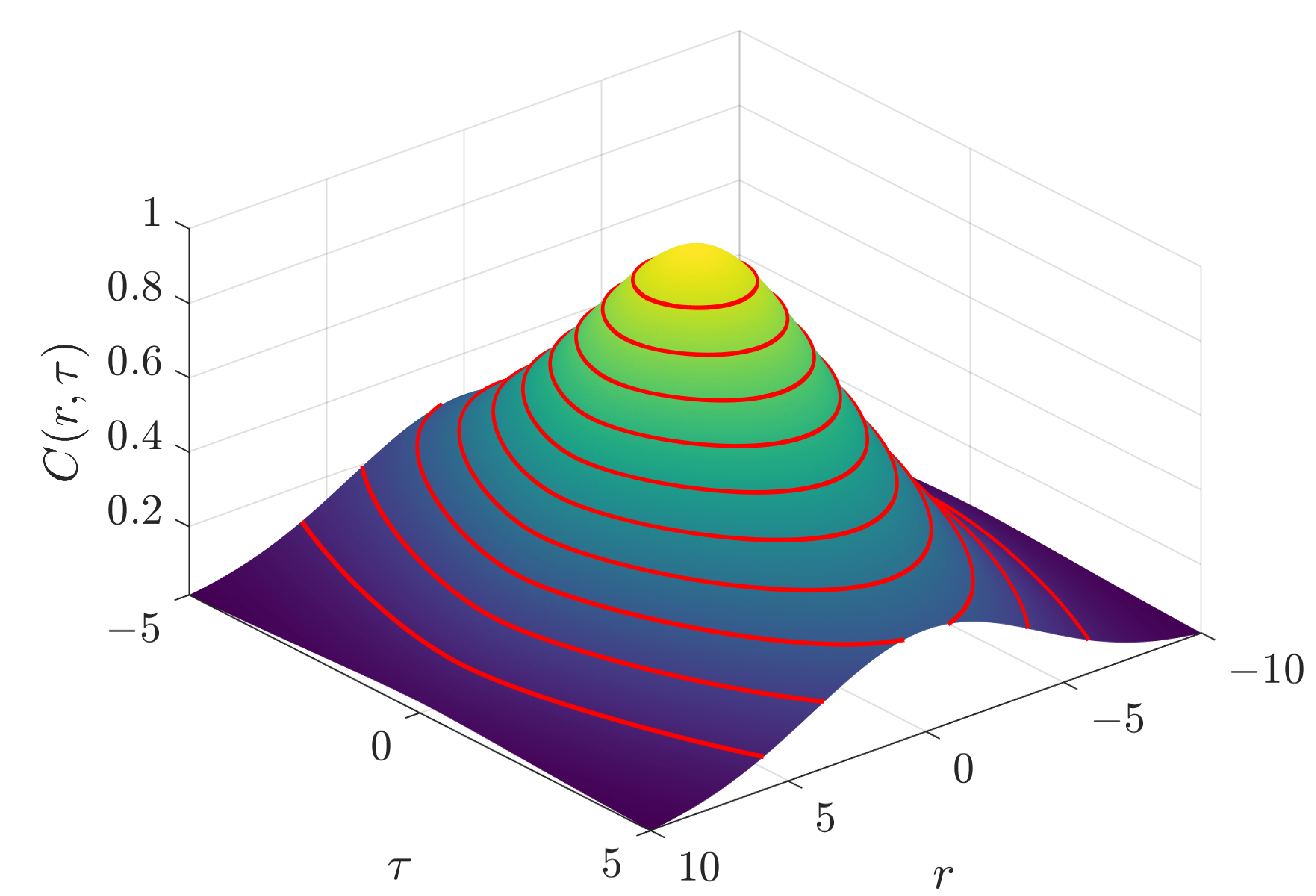}
\caption{Normalized $C(r,\tau)$ and isolevel contour lines (red online) in the overdamped regime, obtained from~\eqref{eq:ldho-kernel-st-over} using
$\tilde{\omega}_{d}=\pi/10$, $\tilde{\tau}_{c}=0.8$, $b=0.4$,  $\epsilon=8$, and $d=2$.}
\label{fig:kernel_over_1}
\end{figure}

\begin{IEEEproof}
The proof is given in Appendix~\ref{app:over}.
\end{IEEEproof}
A plot of the overdamped kernel $C(r,\tau)$  is shown in Fig.~\ref{fig:kernel_over_1}.
\smallskip
\begin{rem}[Variance stablilization]
\label{rem:over}
As in the underdamped case,  the spectral decay hyperparameter $\epsilon$ stabilizes the variance (i.e., the behavior at $\tau=0$),  and $b$ adjusts the space-time interaction since for $b=0$ the space and time dependence in~\eqref{eq:ldho-kernel-st-over} decouple.
\end{rem}
\smallskip

\paragraph{Zero-lag marginal covariances}
The spatial and temporal marginal kernels of Definition~\ref{defi:marginal} are  obtained from~\eqref{eq:ldho-kernel-st-over} by setting $\tau=0$ and $r=0$ respectively, following simple algebraic calculations.
\begin{align}
\label{eq:ldho-over-marginal-r}
& \Cms(r)=  \frac{c_{0}\tilde{\tau}_{c}^{d/2-1}}{4\tilde{\omega}_{d} }  \frac{(\beta_{f}-\beta_{s}) \, \E^{-\frac{r^{2} }{ 4 \epsilon }}}{ \left(4\pi \epsilon \tilde{\tau}_{c}\right)^{d/2}} = \frac{c_{0}\, \E^{-\frac{r^{2} }{ 4 \epsilon }}}{\left(4\pi \epsilon \right)^{d/2} }  ,
\end{align}
\begin{align}
\label{eq:ldho-over-marginal-t}
 \Cmt(\tau) = & \frac{c_{0}\tilde{\tau}_{c}^{d/2-1}}{4\tilde{\omega}_{d}  } \left[ \frac{\beta_{f} \, \E^{-\frac{\beta_{s} \lvert \tau \rvert}{2\tilde{\tau}_{c}}}}{ \left(2\pi b \beta_{s} \lvert \tau \rvert + 4\pi \epsilon \tilde{\tau}_{c}\right)^{d/2}}
\right. \nonumber \\
 & \left. \quad\quad\quad  - \frac{\beta_{s} \, \E^{-\frac{\beta_{f} \lvert \tau \rvert}{2\tilde{\tau}_{c}}}}{\left(2\pi b \beta_{f} \lvert \tau \rvert + 4\pi \epsilon \tilde{\tau}_{c}\right)^{d/2}}
\right]\,.
\end{align}
The spatial marginal kernel~\eqref{eq:ldho-over-marginal-r} is given by the square exponential function as in the underdamped case~\eqref{eq:ldho-under-marginal-r}.  The temporal marginal kernel~\eqref{eq:ldho-over-marginal-t} comprises a combination of slow and fast  exponential kernels. This is analogous to the purely temporal case~\eqref{eq:cov-ldho-o}, albeit  the coefficients of the exponentials are renormalized and include temporal dependence.

\medskip
\subsubsection{Critical damping}

In this regime it holds that $\tilde{\omega}_{0} \tilde{\tau}_{c}=1/2$.

\begin{theorem}[LDHO  kernel in critical damping regime]
\label{theo:ldho-kernel-st-crit} If $\tilde{\omega}_{0} \tilde{\tau}_{c}=1/2$, the LDHO spatiotemporal kernel generated by the radial spectral density~\eqref{eq:spd-ldho-st-1}
 is given by
\begin{align}
\label{eq:ldho-kernel-st-crit}
C(r,\tau) & =  c_{0} \left(\frac{ \tilde{\tau}_{c}}{2\pi \left( b\lvert \tau \rvert + 2 \epsilon \tilde{\tau}_{c}\right)} \right)^{d/2} \E^{-\frac{\lvert \tau \rvert}{2\tilde{\tau}_{c}} - \frac{r^{2}  \tilde{\tau}_{c}}{ 2 b\lvert \tau \rvert + 4 \epsilon \tilde{\tau}_{c}} } \nonumber\\
&  \times
 \left[ 1 + \frac{\lvert \tau \rvert}{2\tilde{\tau}_{c}} -\left( \frac{r^{2} \, b \lvert\tau \rvert \tilde{\tau}_{c}}{2(b\,\lvert \tau \rvert + 2 \epsilon\tilde{\tau}_{c})^{2}}   - \frac{d\, b\lvert\tau \rvert}{2b\lvert \tau \rvert + 4 \epsilon\tilde{\tau}_{c}}\right) \right].
\end{align}
\end{theorem}
\smallskip
\begin{IEEEproof}
The proof is given in Appendix~\ref{app:critical}.
\end{IEEEproof}

\smallskip

\begin{rem}[Hyperparameters at critical damping]
The critically damped LDHO kernel~\eqref{eq:ldho-kernel-st-crit}  includes four independent  hyperparameters: $c_{0}, \tilde{\tau}_{c}, \epsilon, b$.  The fifth hyperparameter, $\tilde{\omega}_{d}$, is not meaningful since  $\tilde{\omega}_{d}=0$ at critical damping.  The critical-damping kernel~\eqref{eq:ldho-kernel-st-crit} can be viewed as the limit of the overdamped kernel~\eqref{eq:ldho-kernel-st-over} for $\tilde{\omega}_{d} \to 0$, which implies $\beta_{s} \to 1$,  $\beta_{f} \to 1$.  The comments in Remark~\ref{rem:over} regarding the role of $b$ and $\epsilon$ also hold for the critically damped case.
\end{rem}

\paragraph{Zero-lag marginal covariances}
The spatial and temporal marginal kernels of Definition~\ref{defi:marginal} are  obtained from~\eqref{eq:ldho-kernel-st-crit} by setting $\tau=0$ and $r=0$ respectively. We thus obtain
\begin{align}
\label{eq:ldho-crit-marginal-r}
 \Cms(r)  & =  \frac{c_{0}\, \E^{-\frac{r^{2} }{ 4 \epsilon }}}{\left(4\pi \epsilon \right)^{d/2} } ,
\end{align}
\begin{align}
\label{eq:ldho-crit-marginal-t}
 \Cmt(\tau)& =  c_{0} \left(\frac{ \tilde{\tau}_{c}}{2\pi \left( b\lvert \tau \rvert + 2 \epsilon \tilde{\tau}_{c}\right)} \right)^{d/2} \E^{-\frac{\lvert \tau \rvert}{2\tilde{\tau}_{c}} }
 \nonumber \\
 & \quad \quad \quad
 \left[ 1 + \frac{\lvert \tau \rvert}{2\tilde{\tau}_{c}} + \frac{d\, b\lvert\tau \rvert}{2b\lvert \tau \rvert + 4 \epsilon\tilde{\tau}_{c}} \right].
\end{align}

\medskip

\section{Properties of LDHO Kernels}
\label{sec:properties}

\paragraph{Full symmetry}
A stationary covariance kernel is \emph{fully symmetric} if the following equalities hold for all $\bfr \in \Rd, \tau \in \R$~\cite{Gneiting02}:
\[
C(\bfr,\tau)=C(\bfr,-\tau)=C(-\bfr,\tau)=C(-\bfr,-\tau)\,.
\]
Since the LDHO spatiotemporal kernels depend on $\bfr$ and $\tau$ only via $r=\rr$ and $\lvert \tau \rvert$, they are fully symmetric.

Full symmetry is not a suitable assumption for transport processes with a dominant advection velocity~\cite{Gneiting02,Salvana20}. For such phenomena,  the LDHO covariance kernels can be extended by invoking Taylor's frozen field hypothesis~\cite{Taylor38,Lvov99}, according to which a non-symmetric spatiotemporal covariance $C(\bfr,\tau)$ can be obtained from a purely spatial model $C_{s}(\bfr)$  by means of
$C(\bfr,\tau)=C_{s}(\bfr- \mathbf{v}\tau)$, where $\mathbf{v}$ is the uniform advection velocity, e.g.~\cite{Salvana20}. Full symmetry is broken in the frozen-field model because, except for $\tau =0$ and $\mathbf{v} =\bfo$, it holds  that  $\bfr- \mathbf{v}\tau \neq \bfr+ \mathbf{v}\tau$; therefore, there exist $\bfr$ and $\tau$ such that $C(\bfr,-\tau) \neq C(\bfr,\tau)$. The ``frozen field'' assumption means that  the correlation between two points at the same location separated by a time distance $\tau$ is the same as the synchronous correlation  between two points that lie apart by $\bfr=\mathbf{v}\tau$. If the spatial distance $\bfr$ is replaced with the
composite space-time distance $\bfr- \mathbf{v}\tau$ in the LDHO kernels, models that are not fully symmetric are generated.  These models, however, do not respect the frozen-field condition, since they  depend  on $\tau$ in addition to  $\bfr- \mathbf{v}\tau$.

\paragraph{Hole effect}
Commonly used isotropic covariance kernels, such as the exponential (Ornstein-Uhlenbeck), square exponential (Gaussian), and Whittle-Mat\'{e}rn models are admissible  for input spaces of any dimension $d \in \Na$.  They can be extended to space-time by means of a composite space-time distance $u=\sqrt{r^{2}+a^{2}\tau^{2}}$, where $a>0$, and $\mathbf{u}=(\bfr, a\tau) \in \Rd \times \R$ is the composite lag vector.  Isotropic kernels satisfy  the inequality $C(u) \ge -C(0)/d$~\cite[p.~34]{Adler81}.  Hence, if the same functional form $C(\cdot)$ is valid for all $d \in \Na$, by taking the limit of the lower bound as $d \to \infty$, it follows that  $C(\cdot)$  is non-negative everywhere, and therefore the hole effect is prohibited.

Gneiting's non-separable kernels  are fully-symmetric and expressed as~\cite{Gneiting02}:
\beq
\label{eq:gneiting}
C(r,\tau)=\frac{\sigma^2}{\left(\psi(\tau^2)\right)^{d/2}}\,
\phi\left( \frac{r^{2}}{\psi(\tau^2)}\right), \;
(\bfr,\tau) \in \Rd \times \R,
\eeq
where $\phi(\cdot)$ is a completely monotone function and $\psi(\cdot)$ is a positive function with a completely monotone derivative (i.e., a Bernstein function).
Hence, this kernel family excludes negative values (i.e., the ``hole effect'').

\begin{figure}
\centering
\includegraphics[width=.99\linewidth]{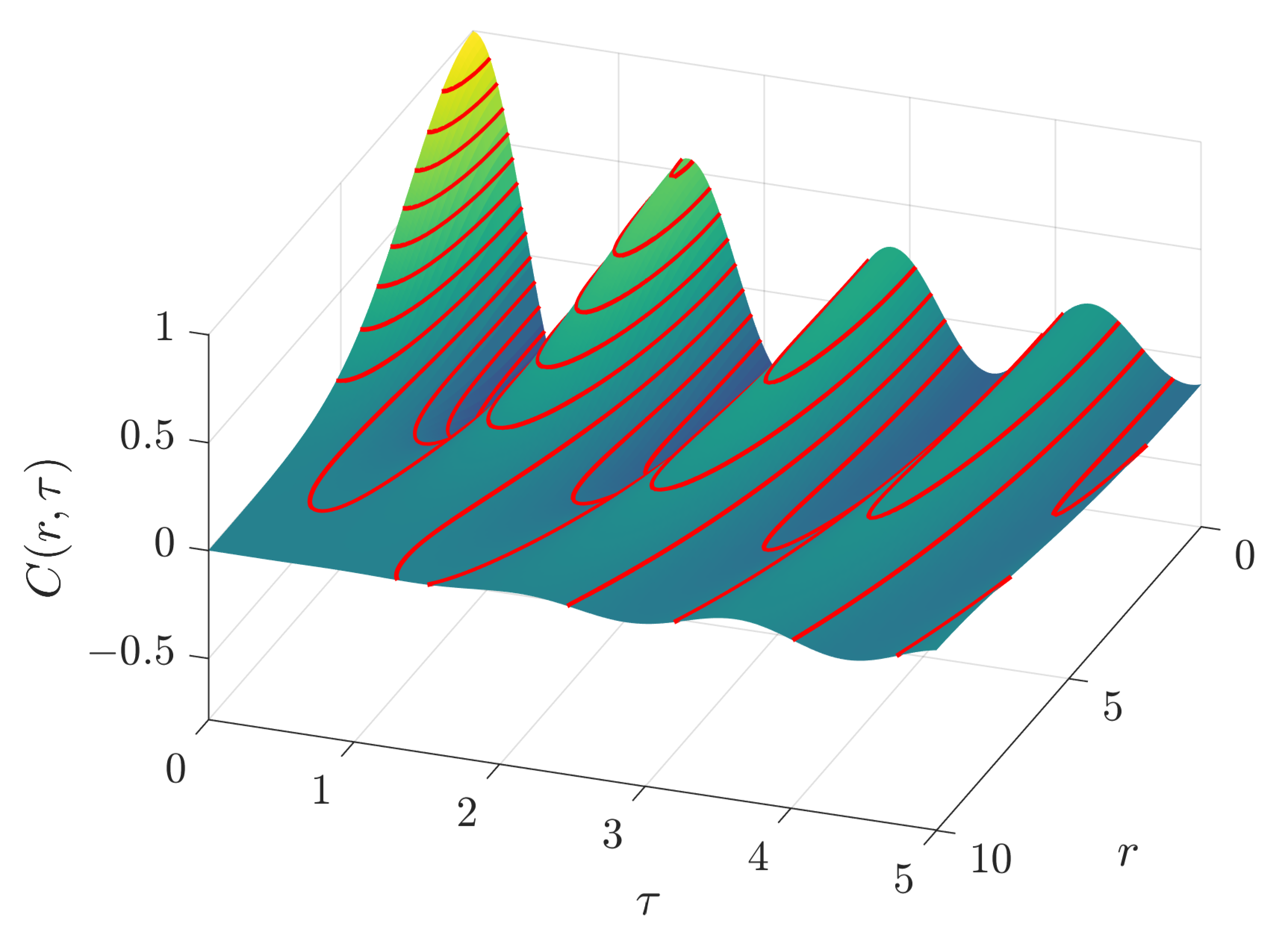}
\caption{Normalized $C(r,\tau)$ and isolevel contours (red online) in the underdamped regime, obtained from~\eqref{eq:ldho-kernel-st-under} using
$\tilde{\omega}_{d}=3\pi/2$, $\tilde{\tau}_{c}=3$, $b=0.4$,  $\epsilon=3$, and $d=2$.}
\label{fig:kernel_under_2}
\end{figure}

In contrast,  the LDHO kernel in the underdamped regime~\eqref{eq:ldho-kernel-st-under} allows negative correlations at certain spatial and temporal lags even for large $d$.  As evidenced in~\eqref{eq:ldho-under-marginal-r} and~\eqref{eq:ldho-under-marginal-t}, oscillations are favored by  (i) $\tau_c \uparrow$ and (ii) $\epsilon \uparrow$, since (i) slows down the temporal  and (ii) slows down the spatial decay. On the other hand,  large values of $b$ tend to suppress correlations as $\tau \uparrow$ and thus also suppress oscillations. Fig.~\ref{fig:kernel_under_2} illustrates a kernel with a ``deep hole effect'': the normalized $C(r,\tau)$ is plotted for the same hyperparameters as in Fig.~\ref{fig:kernel_under_1} except that $\epsilon=3$ instead of $\epsilon=1$. In this case,  the most negative peak of $C(r,\tau)$ is $C(0,0.7538)=-0.7853$.

\paragraph{Marginal kernels}
All three LDHO \emph{spatial} marginal kernels exhibit square exponential decay of the correlations  [cf.~\eqref{eq:ldho-under-marginal-r}, \eqref{eq:ldho-over-marginal-r},~\eqref{eq:ldho-crit-marginal-r}].  This  is inherited from the square exponential spectral decay of the dispersion function $A(k)$~\eqref{eq:dispersion}  which is transferred to real space via the  spectral integral~\eqref{eq:Jnu}.
The \emph{temporal} marginal kernels in all three regimes behave as the respective purely temporal LHDO kernels~\eqref{eq:cov-ldho} with renormalized coefficients.

\paragraph{Interactions}
The LDHO kernels exhibit space-time interactions enabled by the hyperparameter $b$ which determines (in all three regimes) to what extent the spatial and temporal lags are coupled in non-separable expressions [cf.~\eqref{eq:ldho-kernel-st-under}, \eqref{eq:ldho-kernel-st-over}, \eqref{eq:ldho-kernel-st-crit}]. If $b=0$ separable space-time covariance models are obtained. The presence of interactions in any given dataset can be tested using   statistical separability tests which are based on the \emph{interaction ratio}~\cite{Mitchell05,Chen21}:
\begin{equation}
\label{eq:interaction-ratio}
\rint(r,\tau) \triangleq \frac{C(0,0) C(r,\tau)}{C(r,0)C(0,\tau)} = \frac{C(0,0) C(r,\tau)}{\Cms(r)\,\Cmt(\tau)}\,.
\end{equation}
The interaction ratio is equal to one for separable models~\cite{Rodrigues10}. From~\eqref{eq:interaction-ratio} it also follows that $\rint(r,0)=\rint(0,\tau)=1$.  Within the framework of LDHO models, an initial estimate of $b$ can be obtained by fitting the marginal temporal covariance kernels to the data, exploiting the fact that   $b>0$ modifies the  $\tau$ dependence of the temporal marginal kernels.
For nonzero $r, \tau$, it is shown empirically (cf. plots in Section~S2 of the Supplement) that $\rint(r,\tau)$  can take both positive and negative values depending on the hyperparameter values and the space-time lags.

On the other hand, the LDHO temporal Fourier modes  $\tC_{-\omega}(\bfk,\tau)$ satisfy the linear generative ODE~\eqref{eq:ldho-cov-ode}, albeit with $\bfk$-dependent coefficients.  The linearity of the  generative ODE implies a lack of interaction between different Fourier modes.  Hence, the LDHO spatiotemporal kernel cannot capture nonlinear interactions that emerge in turbulent flows~\cite{Yakhot89} and cosmological structure formation~\cite{He19}. Nonetheless, this does not preclude the use of LDHO kernels as approximations, especially in  the framework of variational Gaussian processes~\cite{Titsias09,Damianou11,Tran15}.

\paragraph{Periodicity} Periodic kernels are suitable for physical processes that exhibit regular or  quasi-regular variation in time such as stellar activity~\cite{Nicholson22}. The MacKay periodic kernel~\cite{MacKay98} $C(t,t')= \sigma^{2} \, \exp\left[-a \sin^{2}\left(\pi(t-t')/T\right) \right]$, where $T$ is the period, is used in such cases~\cite{Rasmussen06}.  The periodic kernel is often multiplied with a square exponential, in order to model quasi-periodic behavior~\cite{Rasmussen06,Roberts13,Nicholson22}.  In spatial processes, the product of cosine, $\cos(\bfk_{0}\cdot \bfr)$, and exponential, $\exp(-\rr/\xi)$  kernels is used for the same purpose~\cite[p.~97]{Chiles12}.  The MacKay  kernel can also be used to construct products of separable terms which are periodic along each direction of the input space~\cite[p.~120]{dth20}. Separable scale mixtures that comprise products of periodic components with squared exponential functions at multiple scales have also been considered~\cite{Wilson13}.
The LDHO covariance~\eqref{eq:ldho-kernel-st-under} is, by construction, a \emph{quasi-periodic spatiotemporal} kernel.  In contrast with the MacKay kernel which only takes positive values, the LDHO functions incorporates both positive and negative correlations.  Furthermore, at the limit $\tau_{c} \to \infty$, the LDHO kernel~\eqref{eq:ldho-kernel-st-under} tends to a non-damped periodic function.

\paragraph{Connection with other models} The Mat\'{e}rn model comprises a family of flexible spatial kernels with  smoothness properties controlled by a hyperparameter $\nu \in \R_{+}$ ~\cite{Stein99,Rasmussen06,Rue11}.  Spatiotemporal extensions of the Mat\'{e}rn model have been derived~\cite{Gneiting02}. The LDHO temporal kernel~\eqref{eq:cov-ldho-c} at critical damping  is equivalent to the temporal Mat\'{e}rn model with $\nu=3/2$. However, the associated spatiotemporal LDHO kernel~\eqref{eq:ldho-kernel-st-crit} exhibits different space-time interactions,  inherited by the dispersion functions, than  the spatiotemporal Mat\'{e}rn model. LDHO models with temporal smoothness orders $\nu=p+1/2$, $p \in \{2, 3, \ldots\}$ can be obtained using higher-order generative ODEs (see Section~\ref{sec:conclusions}).

\section{Other Kernels based on the Hybrid Spectral Method}
\label{sec:other-kernels}
In this section we derive  additional kernels using the hybrid spectral method.  The first family comprises LDHO kernel functions obtained from a linear in $k$ dispersion function $B(k)$ and an $A(k)$ that decays exponentially with $k$.
The second family employs temporal Fourier modes derived from the first-order Ornstein-Uhlenbeck ODE using two different dispersion function pairs.

\subsection{LDHO Covariance Kernels with Linear-$k$ Dependence of Dispersion Relations}
\label{sec:ldho-exponential}
We use the radial dispersion functions $B(k)=1+\xi\,k$, $\xi>0$, and $A(k)=\E^{-\epsilon k} B(k)$,  $\epsilon>0$. These functions comply with the general principles laid out in Section~\ref{ssec:dispersion}. The spectral density of the respective LDHO  kernel is obtained by inserting in~\eqref{eq:ldho-dispersion} the scaling relations~\eqref{eq:scaling} and the above radial dispersion functions, leading to
\beq
\label{eq:spd-ldho-st-2}
\tC(k,\omega)=  \frac{\sigma_{0}^{2} \, \left( 1 + \xi\, k\right) \,\E^{-\epsilon k}}{ \omega^2 + \left[\omega^2 - \tilde{\omega}_{0}^{2}\, \left( 1 + \xi k \right)^2 \right]^2 \, \frac{\tilde{\tau}_{c}^2}{(1+\xi k)^2}} \,.
\eeq

The  spatiotemporal kernels are obtained following the same mathematical steps as those described in Appendix~\ref{app:kernel-ldho-squared} for $k^{2}$ dependence of the dispersion functions.  Below we present the main results, while the details of the derivations are given in  the Supplement (Section~S3).

\begin{theorem}[LDHO kernel in underdamped regime]
\label{theo:ldho-kernel-st-under-linear}
If $\tilde{\omega}_{0} \tilde{\tau}_{c}>1/2$, the LDHO spatiotemporal kernel generated by the radial spectral density~\eqref{eq:spd-ldho-st-2}   is given by
\begin{subequations}
\label{eq:ldho-kernel-st-under-linear}
\begin{align}
C(r,\tau) & = c_{0}\, \E^{-\frac{\lvert \tau \rvert}{2\tau_c}}\, \left[ F_{1}(r,\tau)+ F_{2}(r,\tau)\right]\,,
\\
F_{1}(r,\tau) & = g_{0}(r,\tau)\, \cos\left(\tilde{\omega}_{d}\tau + \frac{(d+1)\gamma}{2} -\phi\right)\,,
\nonumber \\
F_{2}(r,\tau) & = \frac{g_{0}(r,\tau)}{2\tilde{\omega}_{d}\tilde{\tau}_{c}}\, \sin\left(\tilde{\omega}_{d}\tau + \frac{(d+1)\gamma}{2} -\phi\right)
\end{align}
where the functions $g_{0}(r,\tau)$, $\gamma(r,\tau)$, and $\phi(\tau)$ (the $r,\tau$ dependence of $\gamma$ and $\phi$ was dropped in the above for brevity) are as follows
\begin{align}
    g_{0}(r,\tau) & = \frac{\Gamma(\frac{d+1}{2})}{\pi^{(d+1)/2}} \, \frac{\left(a^{2}_{\Re}+ a^{2}_{\Im}\right)^{1/2}}{ \left[\left(a^{2}_{\Re}+ a^{2}_{\Im}+ r^{2}\right)^{2} - 4a_{\Im}^{2} r^2 \right]^{(d+1)/4}} \,,
\\
\tan\gamma(r,\tau) & = \frac{2a_{\Im}a_{\Re}}{a_{\Re}^2-a_{\Im}^{2} + r^2 } \,, \;
\tan\phi(\tau) = \frac{a_{\Im}}{a_{\Re}} \,,
\\
a_{\Re}  & = \frac{\xi \lvert \tau \rvert}{2\tilde{\tau}_{c}}+ \epsilon\,, \; a_{\Im} = \xi \lvert \tau \rvert \tilde{\omega}_{d}\,.
\end{align}

\end{subequations}

\end{theorem}

\begin{theorem}[LDHO kernel in overdamped regime]
\label{theo:ldho-kernel-st-over-linear}
If $\tilde{\omega}_{0} \tilde{\tau}_{c}<1/2$, the LDHO spatiotemporal kernel generated by the radial spectral density~\eqref{eq:spd-ldho-st-2}   is given by
\begin{align}
\label{eq:ldho-kernel-st-over-linear}
C(r,\tau)= &   \frac{c_{0}^\ast \,\beta_{f} \, \E^{-\frac{\beta_{s} \lvert \tau \rvert}{2\tilde{\tau}_{c}}}}{ \left[ \left(\frac{\xi \beta_{s} \lvert \tau \rvert}{2\tilde{\tau}_{c}} + \epsilon\right)^2 + r^2 \right]^{(d + 1)/2}}
\nonumber \\
&   -   \frac{c_{0}^\ast \, \beta_{s} \, \E^{-\frac{\beta_{f} \lvert \tau \rvert}{2\tilde{\tau}_{c}}}}{ \left[ \left(\frac{\xi \beta_{f} \lvert \tau \rvert}{2\tilde{\tau}_{c}} + \epsilon\right)^2 + r^2 \right]^{(d + 1)/2}} \,  ,
\end{align}
where $c_{0}^\ast = \frac{c_{0}\,\Gamma(\frac{d+1}{2})}{4\tilde{\omega}_{d} \,\tilde{\tau}_{c} \,\pi^{(d+1)/2}}$,  $\beta_{s}=1- 2\tilde{\tau}_{c} \tilde{\omega}_{d}$,  $\beta_{f}=1 + 2\tilde{\tau}_{c} \tilde{\omega}_{d}$.

\end{theorem}

\medskip

\begin{theorem}[LDHO kernel in critically damped regime]
\label{theo:ldho-kernel-st-crtical-linear}
If $\tilde{\omega}_{0} \tilde{\tau}_{c}=1/2$, the LDHO spatiotemporal kernel generated by the radial spectral density~\eqref{eq:spd-ldho-st-2}   is given by the following expression, where $a_{\Re}$ is defined in~\eqref{eq:ldho-kernel-st-under-linear}:
\begin{subequations}
\label{eq:ldho-kernel-st-critical-linear}
\begin{align}
C(r,\tau) & = c_{0}\, \E^{-\frac{\lvert \tau \rvert}{2\tilde{\tau}_{c}}}\, \left[ \left( 1 + \frac{\lvert \tau \rvert}{2\tilde{\tau}_{c}}\right) C_{1}(r,\tau)+ \frac{\xi\lvert \tau \rvert}{2\tilde{\tau}_{c}} \, C_{2}(r,\tau)\right]\,,
\\
C_{1}(r,\tau) & = \frac{\Gamma(\frac{d+1}{2})}{\pi^{(d+1)/2}}\frac{1}{\left( r^{2} + a^{2}_{\Re}\right)^{(d+1)/2}}\,,
 \\
C_{2}(r,\tau) & = \frac{(d+1)\Gamma(\frac{d+1}{2})}{\pi^{(d+1)/2}}\, \frac{a_{\Re}}{\, \left( r^{2} + a^{2}_{\Re}\right)^{(d+3)/2}}\,.
\end{align}

The properties (a), (b), (d), (e) described in Section~\ref{sec:properties} are also valid for the above LDHO kernels with linear-$k$ dispersion relations.  The marginal spatial kernels $\Cms(r)$ obtained from~\eqref{eq:ldho-kernel-st-under-linear}-\eqref{eq:ldho-kernel-st-critical-linear} for $\tau=0$ decay as power laws with a dominant term which behaves as $\Or(r^{-(d+1)})$ as $r\to \infty$, while the decay of the temporal marginal kernels $\Cmt(\tau)$ is dominated by  exponential terms  (for more details, see Supplement, Section~S3).

\end{subequations}

\end{theorem}

\subsection{Covariance Kernels based on the Ornstein-Uhlenbeck Model}
\label{sec:o-u-kernel}
In this section we investigate the application of hybrid spectral matching to the Ornstein-Uhlenbeck (O-U) process which satisfies the first-order  SODE:
\beq
\label{eq:o-u}
\frac{\d {z}(t;\om)}{\d t}+\frac{1}{\tau_c}{z}(t;\om) = \sigma_{\eta}\eta(t;\om)\,,
\eeq
where   $\eta(t;\om)$ is the standard Gaussian white noise.  The covariance of the O-U process is given by
$C(\tau)=\sigma^2\, \exp(-\lvert \tau \rvert/\tau_c)$ where $\sigma^2 =\sigma_{\eta}^{2}\tau_{c}/{2}$~\cite[p.~448]{Papoulis02}. The dispersion relations are given by
$\tau_{c} \to \tau_{c}(\bfk)= \tilde{\tau}_{c}/B(\bfk), \, \sigma^{2} \to \sigma^{2}(\bfk) = \sigma_{0}^{2} A(\bfk)$.  The O-U temporal Fourier modes for radial dispersion functions  thus become
\begin{align}
\label{eq:t-ft-modes-O-U}
\tC_{-\omega}(k,\tau) &  = \sigma^{2}_{0}  \, A(k) \, \exp\left[ -\lvert \tau\rvert\,  B(k)/\tilde{\tau}_{c}\right]\,.
\end{align}
Since $A(k)$ is dimensionless, $[\sigma_{0}^{2}]=[\mathrm{X}]^{2}[\mathrm{L}]^d$, where $\mathrm{L}$ represents length, to ensure correct  dimensionality of the FT~\eqref{eq:t-ft-modes-O-U}.
Based on the IFT~\eqref{eq:invcovft-iso}, the O-U covariance kernel is given by the integral $(\nu=d/2-1)$:
\begin{equation}
    \label{eq:invcovft-O-U}
    C({r},\tau)= \frac{\sigma^{2}_{0}}{(2\pi)^{d/2} r^{\nu}} \int_{0}^{\infty}  k^{d/2}\,
    {J_{\nu}(k  r)}  A(k) \, \E^{ -\lvert \tau\rvert\,  B(k)/\tilde{\tau}_{c}} \, \D k\,.
\end{equation}

Below, we derive spatiotemporal  kernel expressions for two different choices of dispersion functions.

\textbf{(1)}  $A(k)=\E^{-\beta k^2}, \; B(k)=a+bk^2$ where $a,b, \beta>0$ are hyperparameters with units $[b]=[\beta]=[L]^2$, $[a]=[L]^{0}$.    In this case, the spectral integral~\eqref{eq:invcovft-O-U} becomes
\[
C({r},\tau)= \frac{\sigma^{2}_{0}\,
\E^{- a\,\lvert \tau\rvert/\tilde{\tau}_{c}}}{(2\pi)^{d/2}\, r^{\nu}}  \int_{0}^{\infty}  k^{d/2}    {J_{\nu}(k  r)} \, \E^{ -b k^{2} \lvert \tau\rvert\,  /\tilde{\tau}_{c} - \beta k^2} \, \D k\,.
\]
Using the table of integrals~\cite[6.631.4, p.~706]{Gradshteyn07} it follows that
\beq
\label{eq:o-u-k-square}
C(r,\tau)= \frac{\sigma^{2}_{0}\, \E^{- a\,\lvert \tau\rvert/\tilde{\tau}_{c}}}{(2\pi)^{d/2}}\,\frac{\E^{- r^{2} / 4\left( \beta + b\,\lvert \tau \rvert/\tilde{\tau}_c \right) }}{\left( \frac{2b\lvert \tau\rvert}{\tilde{\tau}_{c}}+2\beta\right)^{d/2}} \,.
\eeq
The kernel~\eqref{eq:o-u-k-square} involves four free hyperparameters: $\sigma_{0}$, $\tilde{\tau}_{c}/a$, $\tilde{\tau}_{c}/b$ and $\beta$. The spatial and temporal marginal kernels are given respectively by
\begin{subequations}
\label{eq:O-U-kernel-square-marginals}
\begin{align}
\Cms(r) = & \frac{\sigma^{2}_{0}\, }{(4\pi\, \beta)^{d/2}}\,\E^{-r^{2}/4\beta}\,,
    \\
\Cmt(\tau) = & \frac{\sigma^{2}_{0}\, \left(  1 + b\,\lvert \tau \rvert/\beta \tilde{\tau}_c\right)^{-d/2}}{(4\pi\, \beta)^{d/2}}\,\E^{-a\,\lvert \tau\rvert/\tilde{\tau}_{c}}\,  \,.
\end{align}
\end{subequations}
Hence, the spatial marginal is the square exponential  while the temporal marginal is a modified exponential kernel.

\textbf{(2)} $A(k)=\E^{-\beta k}$, $B(k)=a + \xi\,k$, where $a,\xi,\beta>0$  are hyperparameters with units $[b]=[\beta]=[L]$, $[a]=[L]^{0}$.
The spectral  integral~\eqref{eq:invcovft-O-U} becomes ($\nu=d/2-1$):
\[
C({r},\tau)= \frac{\sigma^{2}_{0}\,
\E^{- a\,\lvert \tau\rvert/\tilde{\tau}_{c}}}{(2\pi)^{d/2}\, r^{\nu}}  \int_{0}^{\infty}  k^{\nu+1}    {J_{\nu}(k r)} \, \E^{ - k \left(\beta +\xi \,  \lvert \tau\rvert\,  /\tilde{\tau}_{c} \right)} \, \D k\,.
\]
Using the table of integrals~\cite[6.623.2, p.~702]{Gradshteyn07} we obtain
\beq
\label{eq:o-u-k-linear}
C(r,\tau)= \frac{\sigma^{2}_{0}\, \Gamma(\frac{d+1}{2})\,}{\pi^{(d+1)/2}}\frac{\left( \beta \tilde{\tau}_{c} +\xi \,\lvert \tau\rvert \, \right) \,\E^{- a\,\lvert \tau\rvert/\tilde{\tau}_{c}}}{\tilde{\tau}_{c} \,\left[ r^{2} + \left( \beta + \frac{\xi\lvert \tau\rvert}{\tilde{\tau}_{c}}\right)^{2} \right]^{(d+1)/2}} \,.
\eeq
The kernel~\eqref{eq:o-u-k-linear} involves four free hyperparameters: $\sigma_{0}$, $\tilde{\tau}_{c}/a$, $\tilde{\tau}_{c}/\xi$ and $\beta$. The spatial and temporal marginal kernels are given respectively by the following functions
\begin{subequations}
\label{eq:O-U-kernel-linear-marginals}
\begin{align}
\Cms(r) = & \frac{\sigma^{2}_{0}\, \Gamma(\frac{d+1}{2})\,}{\pi^{(d+1)/2}}\frac{ \beta }{\left( r^{2} +  \beta^2 \right)^{(d+1)/2}} \,\,,
    \\
\Cmt(\tau) = & \frac{\sigma^{2}_{0}\, \Gamma(\frac{d+1}{2})\,}{\pi^{(d+1)/2}}\frac{\,\E^{- a\,\lvert \tau\rvert/\tilde{\tau}_{c}}}{\left( \beta + \frac{\xi\lvert \tau\rvert}{\tilde{\tau}_{c}}\right)^{d}} \,.
\end{align}
\end{subequations}

Plots of the O-U kernels~\eqref{eq:o-u-k-square} and~\eqref{eq:o-u-k-linear} derived above are shown in the Supplement (Section~S4).

\section{Conclusions}
\label{sec:conclusions}

This paper responds to the need for non-separable covariance kernels that incorporate space-time interactions~\cite{Chen21} by proposing a  hybrid spectral approach. New spatiotemporal covariance kernels are then derived which can be used for  regression and classification tasks involving Gaussian processes.

Our results are of interest for the analysis of spatiotemporal data obtained from  processes whose evolution  is governed by SPDEs.  The covariance kernels in such cases should ideally be derived by solving a generative PDE associated with the SPDE that represents the EOM of the process. However, the explicit solution of PDEs is impossible except in simplified cases (e.g., linear models, constant coefficients, tractable initial/boundary conditions), thus hindering the development of physically consistent covariance kernels. In many cases, the generative PDE is not even known.

The novel hybrid spectral approach  proposed herein overcomes the kernel solvability problem. It is based on the Cressie-Huang idea~\cite{Cressie99}, i.e., the construction of non-separable covariance kernels by inverting permissible spectral densities $\tC(\bfk,\omega)$.  However, we introduce two important novel elements:
\begin{enumerate}[wide, labelwidth=!, labelindent=0pt]
\item Instead of $\tC(\bfk,\omega)$, our starting point involves time-dependent kernels $C(\tau;\bmthe)$ which are  derived from a generative ODE. Herein, we focus on the ODE associated with the stochastic, linear, damped harmonic oscillator.
\item The $\bfk$-dependence is inserted by means of suitable dispersion relations that modulate the coefficients of the temporal kernels and lead to admissible Fourier modes $\tC_{-\omega}(\bfk,\tau) \triangleq C\left(\tau;\boldsymbol{\theta}(\bfk)\right)$; these are ultimately integrated (by means of the inverse Fourier transform) to obtain non-separable covariance kernels with space-time interactions.
\end{enumerate}

The spatiotemporal interactions are thus inserted in the kernel by means of the dispersion relations which modify the oscillator hyperparameters at different spatial frequencies.  A judicious choice of the dispersion relations leads to exactly solvable expressions for the LDHO spatiotemporal covariance kernels in the three oscillator regimes (underdamping, critical damping, and overdamping).

The LDHO kernel functions developed herein provide the first, to our knowledge, non-separable covariance kernels that exhibit  both space-time interactions and consistent (that is, not subject to the ``shallow hole'' effect) oscillatory behavior in time, independently of the input space dimension $d$.  In addition, the LDHO kernels have their underpinnings in the  paradigmatic harmonic oscillator model and physically meaningful dispersion functions. The derived isotropic LDHO covariance kernels  involve five hyperparameters (four in the critical regime). This already rich parametric dependence can be extended by  means of scaling factors along each  input dimension in the spirit of automatic relevance determination~\cite{Neal96}.

The hybrid spectral approach  can be investigated  with generative ODEs other than LDHO for the temporal Fourier modes and different dispersion functions.  For example, herein we also developed space-time kernels based on the first-order generative ODE that corresponds to the Ornstein-Uhlenbeck process~\cite{dth21}.  Higher-order generative ODEs are also useful, e.g. in calculations of background-error correlations in variational data assimilation~\cite{Yaremchuk11}.  The spatial roughness of the covariance kernels can be controlled by the asymptotic decay of the mode variance.  This was demonstrated by deriving LDHO kernels based on dispersion functions with square exponential as well as exponential decay.    Methods for  consistent estimation of the LDHO kernel hyperparameters as well as novel algorithms and computational strategies for the efficient simulation of spatiotemporal LDHO Gaussian processes on large spatiotemporal grids, are needed.   Particularly interesting is the extension of the hybrid spectral approach to multi-output (multivariate) Gaussian processes as well as Gaussian processes on manifolds (e.g., spherical surfaces).

\appendices

\section{Proof of Generative  ODE for LDHO Covariance}
\label{app:ldho-ode}

\begin{IEEEproof}
The EOM for the expectation of the harmonic oscillator's displacement is expressed, based on~\eqref{eq:ldho-sode}, as follows
\beq
\label{eq:ldho-mean}
	\EE\left[\frac{\d^{2}{z}(t;\om)}{\d t^{2}} \right] + \frac{1}{\tau_{c}} \, \EE\left[\frac{\d{z}(t;\om)}{\d t} \right] + \frac{1}{\omega_{0}^{2}} \, \EE[z(t;\om)] = 0.
\eeq
Assume that the time derivatives and the expectation  operator commute, being both linear operators~\cite[p.~398]{Papoulis02}.  Then,
if $\overline{z}(t) \triangleq \EE[z(t;\om)]$, the expectation $\overline{z}(t)$ obeys the linear ODE
\beq
\frac{\d^{2}}{\d t^{2}}\overline{z}(t) + \frac{1}{\tau_{c}} \,\frac{\d}{\d  t}\overline{z}(t) + \frac{\overline{z}(t)}{\omega_{0}^{2}}  \,  =0\,.
\eeq
The solution of the ODE is the following damped harmonic function~\cite[Chap.~24]{Feynman11}
\[
\overline{z}(t) = A\, \E^{-t/2\tau_{c}}\,\sin\left( \omega_{d}\, t + \phi_{0} \right),
\]
where the constants $A$ and $\phi_{0}$ are determined by initial conditions. Without loss of generality we assume that $\overline{z}(t=0)= \d \overline{z}(t=0)/\d t=0$. Then it follows that $\overline{z}(t) =0$ for all $t \ge 0$ and $C(\tau)=\EE[z(t+\tau;\om)\,z(t;\om)]$.

\paragraph{LDHO covariance} To derive the EOM for the covariance function we use the SODE~\eqref{eq:ldho-sode} at two different times, $t, t'=t-\tau$, (i) we duplicate~\eqref{eq:ldho-sode} for $t$ and $t'$; (ii) we multiply the respective sides of the two equations, and (iii) we calculate the expectation on both sides of the resulting EOM.  The right-hand side includes the term $\sigma^{2}_{\eta} \, \eta(t;\om)\,\eta(t';\om)$; upon calculating the expectation this leads to $\sigma^{2}_{\eta} \, \delta(t-t')$ in light of the noise covariance in~\eqref{eq:whitenoise}.  For brevity we use $z_{t}$ and $\dot{z}_{t}$ and $\ddot{z}_{t}$ for  $z(t;\om)$, and its first and second derivatives, respectively. The term on the left-hand side contains a sum of  nine product pairs:
\begin{align}
\label{eq:eom-terms}
& \ddot{z}_{t} \ddot{z}_{t'} + \frac{1}{\tau_{c}} \ddot{z}_{t} \dot{z}_{t'} + \frac{1}{\omega^{2}_{0}} \ddot{z}_{t} z_{t'} + \frac{1}{\tau_{c}} \dot{z}_{t} \ddot{z}_{t'} + \frac{1}{\tau_{c}^{2}} \dot{z}_{t} \dot{z}_{t'}
\nonumber \\
& + \frac{1}{\tau_{c}\omega^{2}_{0}} \dot{z}_{t} {z}_{t'} + \frac{1}{\omega^{2}_{0}} z_{t}\ddot{z}_{t'} +  \frac{1}{\omega^{2}_{0}\tau_{c}} z_{t}\dot{z}_{t'}+ \frac{1}{\omega^{4}_{0}} z_{t}\,{z}_{t'} \,.
\end{align}

The following lemma is used to evaluate the oscillator's covariance EOM.

\smallskip

\begin{lemma}[Covariance of process derivatives]
\label{lemma:derivs}
Let $z(t;\om)$ be a stationary,  stochastic process which admits derivatives  up to order $n \in \Na$ in the mean-square sense. Then, the following
identity holds for the covariance of the derivatives of order $k$ and $l$, where $\max(k,l) \le n$~\cite[p.~407,417]{Papoulis02}, \cite[p.~187]{dth20}:
\[
\EE\left[\frac{\d^{k}z(t;\om)}{\d t^{k}}\, \frac{\d^{l}z(t';\om)}{\d {t'}^{l}}\right]= (-1)^{l}\,\left. \frac{d^{k+l} C(\tau)}{d\tau^{k+l}} \right|_{\tau= t-t'} \,.
\]

\end{lemma}

Lemma~\ref{lemma:derivs} is used to evaluate the expectation of the summation~\eqref{eq:eom-terms} using $k, l=0, 1, 2$ according to the order of derivatives in each product.  The expectations of the following terms then   cancel out: the second with  the fourth  and the sixth with the eighth. The remaining terms in the covariance EOM then include
\begin{align}
\label{eq:eom-ldho-cov}
& \EE\left[ \ddot{z}_{t} \ddot{z}_{t'} + \frac{\ddot{z}_{t} z_{t'}+ z_{t} \ddot{z}_{t'}}{\omega^{2}_{0}}  + \frac{\dot{z}_{t} \dot{z}_{t'}}{\tau_{c}^{2}}
+ \frac{z_{t}\,{z}_{t'}}{\omega^{4}_{0}}   \right] =\delta(t-t')\,.
\end{align}

Finally, the LDHO covariance EOM~\eqref{eq:ldho-cov-ode} is obtained by applying Lemma~\ref{lemma:derivs} to the left-hand side of~\eqref{eq:eom-ldho-cov}.

\end{IEEEproof}

\smallskip

\section{Proofs of LDHO covariance kernel expressions}
\label{app:kernel-ldho-squared}

\subsection{LDHO kernel in underdamping regime}
\label{app:under}

\begin{IEEEproof}
The temporal Fourier modes of the LDHO kernel are obtained from~\eqref{eq:cov-ldho-u} by inserting the dispersion functions~\eqref{eq:dispersion} and~\eqref{eq:dispersion-fun}
and using the scaling relations~\eqref{eq:scaling}.
\begin{align}
\label{eq:t-ft-modes-under}
\tC_{-\omega}(k,\tau) &  = c_{0}  \, \E^{ -\frac{\lvert \tau\rvert\,  B(k)}{2\tilde{\tau}_{c}} -\epsilon k^2 } \, \Big[  \cos\big( \tilde{\omega}_{d} B(k) \tau \big)
\nonumber \\
& \quad + \frac{1}{2\tilde{\omega}_{d}\tilde{\tau}_{c}}\ \sin\big( \tilde{\omega}_{d} B(k) \lvert \tau \rvert \big) \Big].
\end{align}

\noindent This can also be expressed in terms of the radial functions $\tilde{F}_{1}(k)$, $\tilde{F}_{2}(k)$ as follows
\begin{subequations}
\label{eq:tC-under}
\begin{align}
\tC_{-\omega}(k,\tau) &  = c_{0}  \, \E^{-\frac{\lvert \tau\rvert}{2\tilde{\tau}_{c}}} \left[ \tilde{F}_{1}(k)+\tilde{F}_{2}(k)\right],
\\[1ex]
\tilde{F}_{1}(k) & = \E^{ -k^{2} \left(\frac{\lvert\tau\rvert\,  b}{2\tilde{\tau}_{c}} +\epsilon \right) } \, \cos\big( \tilde{\omega}_{d} (1+bk^2) \lvert\tau\rvert\, \big),
\\[1ex]
\tilde{F}_{2}(k) & = \E^{ -k^{2} \left(\frac{\lvert \tau\rvert\,  b}{2\tilde{\tau}_{c}} +\epsilon \right) } \frac{\sin\big( \tilde{\omega}_{d} (1+bk^2) \lvert\tau\rvert \big)}{2\tilde{\omega}_{d}\tilde{\tau}_{c}}\,.
\end{align}
\end{subequations}
In~\eqref{eq:tC-under} we replaced $\tau$ with $\lvert\tau\rvert$ in the cosine term; this is allowed due to the symmetry of $\cos(\cdot)$ under sign changes.  The functions $\tilde{F}_{i}(\cdot)$ depend on both $k$ and $\lvert \tau \rvert$, while the $F_{i}(\cdot)$ depend on $r$ and $\tau$ (where $i=1,2$). For reasons of brevity, in the following  only the dependence of $\tilde{F}_{i}(\cdot)$ on $k$ (in the Fourier domain) and of $F_{i}(\cdot)$ on $r$  are shown explicitly.  Based on~\eqref{eq:ift-fourier-mode} and~\eqref{eq:tC-under}, $C(r,\tau)$ is given by
\beq
\label{eq:C-real-under}
C(r,\tau) =  c_{0}  \, \E^{-\frac{\lvert \tau\rvert}{2\tilde{\tau}_{c}}} \left[ F_{1}(r) + F_{2}(r)\right],
\eeq
where $F_{i}(r) = \ift_{\bfk} [\tilde{F}_{i}(k)]$, $i=1,2$. In order to evaluate $F_{i}(r)$  we express the harmonic terms as linear combinations of $\E^{\pm \I x}$
using  Euler's formula $\E^{\I x}= \cos x + \I \sin x$, for $x\in \R$. By defining $a_{r} \triangleq \frac{\lvert \tau\rvert\,  b}{2\tilde{\tau}_{c}} +\epsilon$, it follows that $a_{r} >0$ and the $\tilde{F}_{1}(k)$, $\tilde{F}_{2}(k)$ are given by
\begin{align*}
\tilde{F}_{1}(k) & = \frac{\E^{ -a_{r}  k^{2} }}{2} \, \left[ \E^{\I \left(  \tilde{\omega}_{d}\lvert\tau\rvert + b   \tilde{\omega}_{d}\lvert\tau\rvert\, k^{2} \right) } + \E^{-\I \left(  \tilde{\omega}_{d}\lvert\tau\rvert + b   \tilde{\omega}_{d}\lvert\tau\rvert k^{2} \right) } \right],
\\[1ex]
\tilde{F}_{2}(k) & = \frac{\E^{ -a_{r}  k^{2} }}{4\I \,\tilde{\omega}_{d} \tilde{\tau}_{c}} \, \left[ \E^{\I \left(  \tilde{\omega}_{d}\lvert\tau\rvert + b   \tilde{\omega}_{d}\lvert \tau\rvert k^{2} \right) } - \E^{-\I \left(  \tilde{\omega}_{d}\lvert \tau\rvert + b   \tilde{\omega}_{d}\lvert\tau\rvert k^{2} \right) } \right].
\end{align*}

\noindent Hence, the $k$-dependent parts of $\tilde{F}_{1}(k)$, $\tilde{F}_{2}(k)$ comprise the functions  $\tilde{I}_{\pm}(k)=\exp\left[ - \left( a_{r} \pm \I \, b\,\tilde{\omega}_{d}\,\lvert\tau\rvert \right) k^{2} \right]$:
\begin{align*}
\tilde{F}_{1}(k) & = \frac{1}{2} \, \left[ \E^{\I  \tilde{\omega}_{d}\lvert\tau\rvert } \, \tilde{I}_{+}(k) + \E^{-\I   \tilde{\omega}_{d}\lvert\tau\rvert }  \, \tilde{I}_{-}(k) \right],
\\[1ex]
\tilde{F}_{2}(k) & = \frac{1}{4\I \,\tilde{\omega}_{d} \tilde{\tau}_{c}} \, \left[ \E^{\I  \tilde{\omega}_{d}\lvert\tau \rvert} \, \tilde{I}_{+}(k) - \E^{-\I   \tilde{\omega}_{d}\lvert\tau\rvert }  \, \tilde{I}_{-}(k) \right].
\end{align*}

\noindent Using the identities $\tilde{I}_{-}(k)=I_{+}^{\dagger}(k)$, $\E^{-\I  \tilde{\omega}_{d}\lvert\tau\rvert }=\left( \E^{\I  \tilde{\omega}_{d}\lvert\tau\rvert }\right)^{\dagger}$, and $z_{1}^{\dagger}z_{2}^{\dagger}=(z_{1}z_{2})^{\dagger}$ for any $z_{1}, z_{2} \in \Co$, the functions $\tilde{F}_{1}(k)$, $\tilde{F}_{2}(k)$ are expressed in terms of the real and imaginary parts of the function $\E^{\I  \tilde{\omega}_{d}\tau } \, \tilde{I}_{+}(k)$, i.e.,

\begin{subequations}
\label{eq:F1k-F2k}
\begin{align}
\label{eq:F1}
\tilde{F}_{1}(k) & = \Re \left[ \E^{\I  \tilde{\omega}_{d}\lvert\tau\rvert } \, \tilde{I}_{+}(k)  \right] = \cos(\tilde{\omega}_{d}\lvert\tau\rvert) \, \Re\left[ \tilde{I}_{+}(k)\right]
\nonumber \\
& \quad \quad \quad - \sin(\tilde{\omega}_{d}\lvert\tau\rvert)\, \Im\left[ \tilde{I}_{+}(k)\right],
\\[1ex]
\label{eq:F2}
\tilde{F}_{2}(k) & = \frac{1}{2\,\tilde{\omega}_{d} \tilde{\tau}_{c}} \, \Im \left[ \E^{\I  \tilde{\omega}_{d} \lvert\tau\rvert } \, \tilde{I}_{+}(k)  \right]
\nonumber \\
& \quad  =  \frac{\sin(\tilde{\omega}_{d}\lvert\tau\rvert )}{2\,\tilde{\omega}_{d} \tilde{\tau}_{c}} \Re\left[ \tilde{I}_{+}(k)\right] \nonumber \\
& \quad + \frac{\cos(\tilde{\omega}_{d}\lvert\tau\rvert )}{2\,\tilde{\omega}_{d} \tilde{\tau}_{c}} \Im\left[ \tilde{I}_{+}(k)\right] .
\end{align}
\end{subequations}

Furthermore, since both $\Re\left[ \tilde{I}_{+}(k)\right]$ and $\Im\left[ \tilde{I}_{+}(k)\right]$ are radial functions of $k$, their inverse Fourier transforms are real-valued, radial functions of $r$ according to~\eqref{eq:invcovft-iso}.
Let $I_{+}(r)\triangleq \ift_{\bfk}\left[ \tilde{I}_{+}(k)\right]$ denote the inverse Fourier transform of $\tilde{I}_{+}(k)$.  $I_{+}(r)$ comprises real and imaginary parts denoted by $\Rei(r) \triangleq \Re[I_{+}(r)]$ and
$\Imi(r)\triangleq  \Im[I_{+}(r)]$.  Then,
\begin{align*}
& \ift\left\{\Re[\tilde{I}_{+}(k)]\right\}= \Re\left\{\ift[\tilde{I}_{+}(k)]\right\} = \Rei(r),
\\
& \ift\left\{\Im[\tilde{I}_{+}(k)]\right\}= \Im\left\{\ift[\tilde{I}_{+}(k)]\right\}= \Imi(r)\, .
\end{align*}
Based on  the above IFTs and the spectral functions~\eqref{eq:F1k-F2k}, the inverse Fourier transforms $F_{i}(r)=\ift_{\bfk}[\tilde{F}_{i}(k)], \, i=1, 2$ are given by
\begin{subequations}
\label{eq:F1-F2-ift}
\begin{align}
\label{eq:F1-ift}
F_{1}(r) &  = \cos(\tilde{\omega}_{d}\tau) \, \Rei(r) - \sin(\tilde{\omega}_{d}\lvert\tau\rvert) \, \Imi(r),
\\[1ex]
\label{eq:F2-ift}
F_{2}(r) & = \frac{\sin(\tilde{\omega}_{d}\lvert\tau\rvert )}{2\,\tilde{\omega}_{d} \tilde{\tau}_{c}} \, \Rei(r) + \frac{\cos(\tilde{\omega}_{d}\lvert\tau\rvert )}{2\,\tilde{\omega}_{d} \tilde{\tau}_{c}} \, \Imi(r)\, .
\end{align}
\end{subequations}

The function $I_{+}(r)$ is evaluated by means of the spectral representation~\eqref{eq:invcovft-iso} which involves the following integral
\beq
\label{eq:Iplus}
I_{+}(r) \triangleq \frac{r}{(2\pi r)^{d/2}} \int_{0}^{\infty} \, k^{d/2}
    {J_{d/2-1}(k  r)} \,\E^{- \left( a_{r} + \I \, b\,\tilde{\omega}_{d}\,\lvert\tau\rvert \right) k^{2}} \D k \,.
\eeq
Hence, $I_{+}(r)$ can be evaluated using the following lemma~\cite[Eq.~(6.631.4)]{Gradshteyn07}.

\smallskip

\begin{lemma}[Spectral integral for radial functions]
\label{lemma:Jnu-integral}
Let $J_{\nu}(x)$ represent the Bessel function of the first kind of order $\nu \in \Co$, where $\Re(\nu)>-1$.  Furthermore, let $a \in \Co$ be a constant coefficient with $\Re(a)>0$. Then, the
following is true:
 \begin{align}
\label{eq:Jnu}
& \int_{0}^{\infty} k^{\nu+1}\, J_{\nu}(r k)\,\E^{-a k^{2}} \D k= \frac{r^{\nu}}{(2a)^{\nu+1}}\, \E^{-r^2/4a}.
\end{align}
\end{lemma}

\smallskip

 Hence,  in light of Lemma~\eqref{lemma:Jnu-integral} and by setting $\nu = d/2-1$, the function $I_{+}(r)$ defined in~\eqref{eq:Iplus} is given by the following complex-valued expression
\beq
\label{eq:Ir}
I_{+}(r)=\frac{\E^{-r^2/4a}}{(4\pi a)^{d/2}}, \; \text{where} \;a = a_{r} + \I \, b\,\tilde{\omega}_{d}\,\lvert\tau\rvert\, .
\eeq
If we define $\beta  \triangleq 1/4a$, since $1/a = a^{\dagger}/\lvert a \rvert^{2}$, it follows that $\beta = \beta_{r} + \I \beta_{i}$, where
\begin{subequations}
\label{eq:beta}
\begin{align}
\label{eq:beta-xy}
\beta_{r}= \frac{a_{r}}{4( a_{r}^{2} + b^2\,\tilde{\omega}_{d}^{2}\,\lvert\tau\rvert^{2})}, \;
\beta_{i}= - \frac{b\,\tilde{\omega}_{d}\,\lvert\tau\rvert}{4( a_{r}^{2} + b^2\,\tilde{\omega}_{d}^{2}\,\lvert\tau\rvert^{2})}.
\end{align}
In polar representation, $\beta$ is expressed as $\beta = \lvert \beta \rvert \, \E^{\,\I \phi}$, where
\begin{align}
\label{eq:beta-polar}
\lvert \beta \rvert = \frac{1}{4\sqrt{a^{2}_{r} + b^2\,\tilde{\omega}_{d}^{2}\,\lvert\tau\rvert^{2}}},
 \;
\phi = \atan\left( \frac{-b\,\tilde{\omega}_{d}\,\lvert\tau\rvert}{a_{r}}\right).
\end{align}
\end{subequations}
In light of~\eqref{eq:beta}, the function $I_{+}(r)$ in~\eqref{eq:Ir} is expressed as
\[
I_{+}(r) = \E^{-\beta r^{2}} \left( \frac{\beta}{\pi} \right)^{d/2}  = \left( \frac{\lvert\beta\rvert}{\pi} \right)^{d/2}  \E^{-\beta_{r} r^{2} - \I \left( \beta_{i} r^{2} - \d\phi/2\right)}\,.
\]

\noindent Using the expressions for $\beta_{r}$, $\beta_{i}$ given by~\eqref{eq:beta-xy}  and for  $\lvert \beta \rvert$ given by~\eqref{eq:beta-polar}, we obtain the following expressions for the real and imaginary parts of $I_{+}(r)$
\begin{subequations}
\label{eq:Re-Im-Iplus}
\begin{align}
\Rei(r)= & \frac{\E^{-\lambda^2 r^2} \,\cos\left(\kappa^{2}r^2 + \tfrac{d\phi}{2}\right)}{\left(4\pi \sqrt{a^{2}_{r} + b^2\,\tilde{\omega}_{d}^{2}\,\lvert\tau\rvert^{2}}\right)^{d/2}} ,
\\
\Imi(r)= & \frac{\E^{-\lambda^2 r^2} \,\sin\left(\kappa^{2}r^2 + \tfrac{d\phi}{2}\right)}{\left(4\pi \sqrt{a^{2}_{r} + b^2\,\tilde{\omega}_{d}^{2}\,\lvert\tau\rvert^{2}}\right)^{d/2}},
\end{align}
\end{subequations}
where $\kappa^2 = b\,\tilde{\omega}_{d}\,\lvert\tau\rvert /4( a_{r}^{2} + b^2\,\tilde{\omega}_{d}^{2}\,\lvert\tau\rvert^{2})$, $\lambda^{2} = a_{r}\kappa^2 / b\,\tilde{\omega}_{d}\,\lvert\tau\rvert$.

Finally, the LDHO kernel~\eqref{eq:ldho-kernel-st-under} is obtained by combining~\eqref{eq:C-real-under}, \eqref{eq:F1-F2-ift} and~\eqref{eq:Re-Im-Iplus}.
\end{IEEEproof}

\subsection{LDHO kernel in overdamping regime}
\label{app:over}
\begin{IEEEproof}
The temporal Fourier modes of the LDHO kernel are obtained from~\eqref{eq:cov-ldho-o}-\eqref{eq:fast-slow} by inserting the dispersion functions~\eqref{eq:dispersion} and~\eqref{eq:dispersion-fun}
and using the scaling relations~\eqref{eq:scaling}. This leads to
\beq
\label{eq:t-ft-modes-over}
\tC_{-\omega}(k,\tau) = \frac{c_{0}\, \E^{-\epsilon k^2}}{2\tilde{\omega}_{d} B(k)} \left[ \frac{\E^{ -\frac{\lvert \tau \rvert }{\tau_{s}(k)}}}{\tau_{f}(k)} - \frac{\E^{ -\frac{\lvert \tau \rvert }{\tau_{f}(k)}}}{\tau_{s}(k)}\right].
\eeq
In light of~\eqref{eq:fast-slow} and taking account of the dispersion relations, the fast and slow decay times transform as follows
\begin{subequations}
\label{eq:ts-tf}
\begin{align}
\tau_{s}(k) = & \frac{2\tilde{\tau}_{c}}{B(k) \left(1- 2\tilde{\tau}_{c} \tilde{\omega}_{d} \right)}, \;
\\
\tau_{f}(k) = & \frac{2\tilde{\tau}_{c}}{B(k) \left(1 + 2\tilde{\tau}_{c} \tilde{\omega}_{d} \right)}\,.
\end{align}
\end{subequations}
Based on~\eqref{eq:ts-tf},  the functions $B(k)$ in~\eqref{eq:t-ft-modes-over} cancel out, and  the temporal Fourier modes are given by
\beq
\tC_{-\omega}(k,\tau) = \frac{c_{0}\, \E^{-\epsilon k^2}}{4\tilde{\omega}_{d} \tilde{\tau}_{c} } \left[ \beta_{f} \,\E^{ -\frac{\beta_{s} \lvert \tau \rvert B(k)  }{2\tilde{\tau}_{c}}}  - \beta_{s} \, \E^{ -\frac{\beta_{f} \lvert \tau \rvert B(k) }{2\tilde{\tau}_{c}}} \right],
\eeq
where $\beta_{s}=1- 2\tilde{\tau}_{c} \tilde{\omega}_{d}$ and $\beta_{f}=1 + 2\tilde{\tau}_{c} \tilde{\omega}_{d}$.
Recalling~\eqref{eq:B1} for $B(k)$, the IFT expression~\eqref{eq:ift-fourier-mode} for the LDHO kernel, and the linearity of the IFT, it follows that
\begin{align}
\label{eq:C-over}
C(r,\tau)= & \frac{c_{0}}{4\tilde{\omega}_{d} \tilde{\tau}_{c} } \left[ \beta_{f}  C_{s}(r,\tau)
-  \beta_{s}  C_{f}(r,\tau)\,  \right],
\end{align}
where
\beq
\label{eq:Cj-r-o}
C_{j}(r,\tau) = \E^{ -\frac{\beta_{j} \lvert \tau \rvert}{2\tilde{\tau}_{c}}} \, \ift\left[\,\E^{ -\frac{b\,\beta_{j} \lvert \tau \rvert \,k^2 }{2\tilde{\tau}_{c}}  -\epsilon k^{2}}  \right], \; j=s, f\,.
\eeq

The IFTs in~\eqref{eq:Cj-r-o} can be evaluated using the following lemma.

\begin{lemma}[Inverse Fourier Transform of a Gaussian]
\label{lemma:Gaussian}
The IFT of a square exponential (Gaussian), radial spectral function $\exp(-c k^{2})$ where $c>0$ and $k=\kk$ for $\bfk \in \Rd$ ($d \in \Na$), is  a Gaussian radial function of $r=\rr$ where $\bfr \in \Rd$ is a vector in  the direct space. More precisely, for $c=a^{2}/4$ where $a>0$ it holds that~\cite[p.~160]{dth20}

\beq\label{eq:ift-Gaussian}
\ift[\,\E^{-a^{2}\,k^2/4}]=\E^{-(r/a)^2} / \left( a\sqrt{\pi}\right)^{d}.
\eeq
\end{lemma}

\noindent Hence, setting $a^{2} = 2b\lvert \tau \rvert \beta_{j}/\tilde{\tau}_{c} + 4\epsilon$ for
$j=s,f$, it follows from Lemma~\ref{lemma:Gaussian} that the functions $C_{j}(r,\tau)$ in~\eqref{eq:Cj-r-o} are given by
\beq
\label{eq:Cj-r-over}
C_{j}(r,\tau) = \frac{\tilde{\tau}_{c}^{d/2} \, \E^{-\frac{\beta_{j} \lvert \tau \rvert}{2\tilde{\tau}_{c}}} \,
\E^{-\frac{r^{2}\tilde{\tau}_{c} }{2b\lvert \tau \rvert \beta_{j}+ 4 \epsilon \tilde{\tau}_{c}}}}{\left(2\pi b \, \beta_{j} \lvert \tau \rvert + 4\pi \epsilon \tilde{\tau}_{c}\right)^{d/2}} .
\eeq

Finally, the overdamped LDHO covariance kernel~\eqref{eq:ldho-kernel-st-over} is obtained from~\eqref{eq:C-over} and~\eqref{eq:Cj-r-over}.
\end{IEEEproof}


\subsection{LDHO kernel in  critical-damping regime}
\label{app:critical}
\begin{IEEEproof}
The LDHO temporal Fourier modes  are obtained  from~\eqref{eq:cov-ldho-c} by inserting the dispersion functions~\eqref{eq:dispersion} and~\eqref{eq:dispersion-fun} and using the scaling relations~\eqref{eq:scaling}.  Thus, we get
\beq
\label{eq:t-ft-modes-crit}
\tC_{-\omega}(k,\tau) = c_{0} \E^{ -\frac{\lvert \tau \rvert B(k)}{2\tilde{\tau}_{c}} -\epsilon k^{2}} \left[ 1 + \frac{\lvert \tau \rvert \,B(k)}{2\tilde{\tau}_c }\right].
\eeq
Recalling~\eqref{eq:B1} for $B(k)$, it follows that
\beq
\tC_{-\omega}(k,\tau) =c_{0} \E^{ -\frac{\lvert \tau \rvert (1+bk^{2})}{2\tilde{\tau}_{c}} -\epsilon k^{2}}\, \left( 1+ \frac{\lvert \tau \rvert}{2\tilde{\tau}_{c}} + \frac{b\,k^{2}\lvert \tau \rvert}{2\tilde{\tau}_{c}}\right).
\eeq
Based on the linearity of the IFT we obtain
\begin{align}
\label{eq:C-crit}
C(r,\tau)= & c_{0}\E^{ -\frac{\lvert \tau \rvert}{2\tilde{\tau}_{c}}}\, \left( 1+ \frac{\lvert \tau \rvert}{2\tilde{\tau}_{c}}\right) C_{1}(r,\tau)
\nonumber \\
& + \frac{c_{0} b  \,\lvert \tau \rvert}{2\tilde{\tau}_{c}} \, \E^{ -\frac{\lvert \tau \rvert}{2\tilde{\tau}_{c}}}\,  C_{2}(r,\tau),\end{align}
where
\begin{subequations}
\label{eq:C-crit-ift}
\begin{align}
\label{eq:C1-crit-ift}
C_{1}(r,\tau)= & \ift\left[ \, \E^{-k^{2}\left(\epsilon + \frac{b\lvert \tau \rvert}{2\tilde{\tau}_{c}} \right)} \right],
\\
\label{eq:C2-crit-ift}
C_{2}(r,\tau)= & \ift\left[ \, k^{2} \,\E^{-k^{2}\left(\epsilon + \frac{b\lvert \tau \rvert}{2\tilde{\tau}_{c}} \right)} \right].
\end{align}
\end{subequations}

The IFTs in~\eqref{eq:C-crit-ift} can be calculated using Lemma~\ref{lemma:Gaussian}; more precisely, by setting $a^{2} = 4\epsilon + \frac{2b\lvert \tau \rvert}{\tilde{\tau}_{c}}$ it follows that
\beq
\label{eq:C1-crit}
C_{1}(r,\tau)
=\left(\frac{\tilde{\tau}_{c}}{2\pi \, b\lvert \tau \rvert + 4\pi \epsilon \tilde{\tau}_{c}} \right)^{d/2}
\, \E^{-r^{2}\tilde{\tau}_{c} / \left( 2\, b\lvert \tau \rvert + 4\epsilon \tilde{\tau}_{c}\right)}.
\eeq

\smallskip
To calculate $C_{2}(r,\tau)$ we use the fact that $\kk^{2}$ is the image of the Laplace operator $-\nabla^{2}=\sum_{i=1}^{d} \partial^{2}/\partial r_{i}^{2}$ under the Fourier transform~\cite{Trefethen05}.  Hence, it follows from~\eqref{eq:C-crit} that $C_{2}(r,\tau) = - \nabla^{2} C_{1}(r,\tau)$.

\smallskip

\begin{lemma}[The Laplacian of Gaussian]
\label{lemma:nabla2-C1}
Let $r=\rr, \, \bfr \in \Rd$ and assume that $C_{1}(r,\tau) =\E^{-c(\tau) r^{2}}$, where $c(\tau)  >0$ for all $\tau \in \R$.  Then, the Laplacian of $C_{1}(r,\tau)$ is given by
\beq
\label{eq:nabla2-C1}
\nabla^{2} \,\E^{-c(\tau) r^{2}} = \left[ 4 r^{2}  \,c^{2}(\tau) - 2d \, c(\tau)\, \right] \, \E^{-c(\tau) r^{2}}.
\eeq
\end{lemma}

\begin{IEEEproof}
The Laplacian of a radial function, i.e., $C_{1}(r,\tau)$, is given by~\cite[p.~190]{dth20}
\[
\nabla^{2} C_{1}(r,\tau) = \frac{\D^{2}C_{1}(r,\tau)}{\D r^{2}} + \left(\frac{d-1}{r}\right)\frac{\D C_{1}(r,\tau)}{\D r}.
\]

The Laplacian of the square exponential~\eqref{eq:nabla2-C1} follows from  the above and the independence of $c(\tau)$  on $r$.
\end{IEEEproof}
\medskip

Next, we apply Lemma~\ref{lemma:nabla2-C1} to $C_{2}(r,\tau) = - \nabla^{2} C_{1}(r,\tau)$,  where $c(\tau) \to \tilde{\tau}_{c} / \left( 2\, b\lvert \tau \rvert + 4\epsilon \tilde{\tau}_{c}\right)$,  to obtain
\beq
\label{eq:C2-crit}
C_{2}(r,\tau)= \left(\frac{d \tilde{\tau}_{c}}{ b\lvert \tau \rvert + 2\epsilon \tilde{\tau}_{c}} - \frac{r^{2}\tilde{\tau}_{c}^{2}}{ (b\lvert \tau \rvert + 2\epsilon \tilde{\tau}_{c})^2} \right)\,
\E^{- \frac{r^{2}\tilde{\tau}_{c}}{ 2\, b\lvert \tau \rvert + 4\epsilon \tilde{\tau}_{c}}}.
\eeq

Finally, by combining~\eqref{eq:C-crit} with~\eqref{eq:C1-crit} and \eqref{eq:C2-crit} the
critically damped LDHO kernel~\eqref{eq:ldho-kernel-st-crit} is obtained.
\end{IEEEproof}

\section*{Acknowledgment}
I would like to thank my colleagues, professors Athanasios Liavas and Aggelos Bletsas (ECE, Technical University of Crete) for their helpful input.

\Urlmuskip=0mu plus 1mu\relax


\onecolumn
\renewcommand\thesection{S.\arabic{section}}
\setcounter{subsection}{0}
		\renewcommand\thesubsection{\thesection.\arabic{subsection}}

\begin{center}\Large\textbf{Supplementary Information}\end{center}

\section*{S1. Estimation of LDHO model from synthetic data}
\label{sec:app-data-anal}

In this section we focus on estimating the hyperparameters of the LDHO covariance kernel in the underdamped regime using a synthetic dataset.

\subsection{Data}
\begin{figure}[!hb]
\centering
\includegraphics[width=0.8\textwidth]{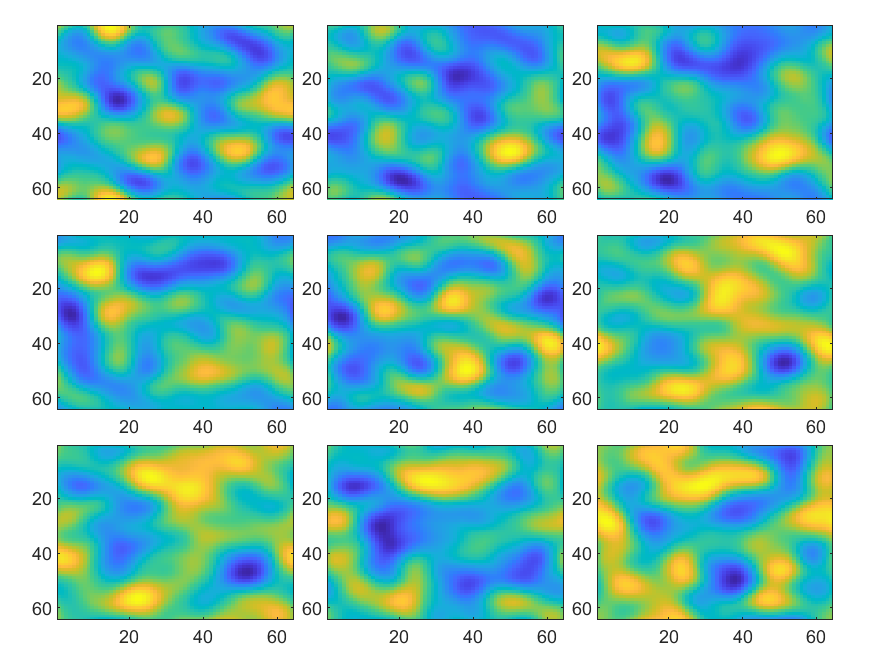}
\caption{First nine time slices of the synthetic data corresponding to $z(\bfs, t)$, where $t=1, \ldots, 9$.}
\label{fig:synthetic-data-space}
\end{figure}

\begin{figure}[!ht]
\centering
\includegraphics[width=0.8\textwidth]{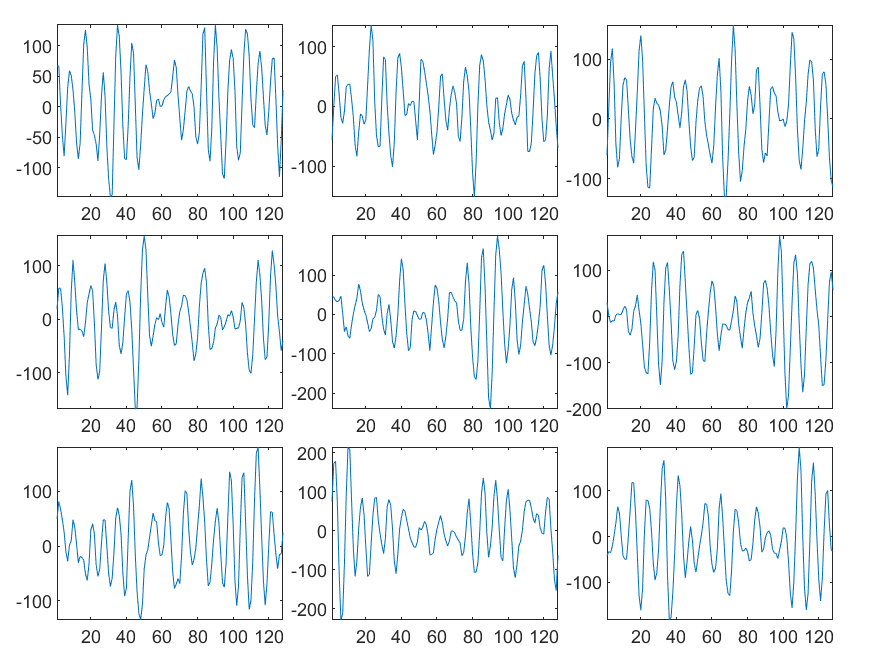}
\caption{Time series of the synthetic data corresponding to $z(s_{i}, s_{i}, t)$ where
$s_{i}=7\,(i-1)+ 1$, and $i=1, 2, \ldots, 9$.}
\label{fig:synthetic-data-time}
\end{figure}

The synthetic data are simulated on a space-time grid $(\bfs_{i},t_{j})$ with dimensions $64\times 64 \times 128$ where $i=1, \ldots, 64$ and $j=1, \ldots, 128$. The data are drawn from a joint Gaussian distribution with an underdamped LDHO covariance kernel using  the Fast Fourier transform spectral simulation method~\cite{dth20}.
In Fig.~\ref{fig:synthetic-data-space} we illustrate the spatial distribution of the data for the first nine time slices.  Fig.~\ref{fig:synthetic-data-time} shows nine time series of the data drawn from the nodes with coordinates $\left(s_{i}, s_{i}, \{t_{j} \}_{j=1}^{N_t}\right)$, where $i \in \{1, 8, 15, 22, 29, 36, 43, 50, 57 \}$, where $N_{t}=128$ is the number of times and $N_{s}=64\times 64$ is the total number of spatial nodes.

\subsection{Estimation Method}
The method of maximum likelihood estimation (MLE) is commonly used to determine kernel hyperparameters~\cite{Rasmussen06} since it is computationally efficient~\cite{Chen21}. However,  MLE is computationally intensive and thus impractical for the current dataset which contains $N \gtrapprox 10^5$ sampling points. Hence, for large datasets approximations based on composite likelihoods are often used~\cite{Varin11}.

Herein we opt for the method of moments in which the  hyperparameters are fitted by fitting the ``theoretical'' kernel function to  sample-based kernel estimates~\cite{Cressie91}. The method of moments is computationally more efficient and provides an easy visual assessment of the quality of the fit. More specifically, instead of the covariance  we use the semi-variogram function (also known as structure function, henceforward ``variogram'' for short) defined by~\cite{Cressie91,Chiles12}
\beq
\label{eq:variogram}
\gamma(\bfr,\tau) = \frac{1}{2}\, \var \left[  z(\bfs+\bfr, t + \tau)- z(\bfs, t)\right]\,,
\eeq
where $\var[\cdot] \triangleq \EE[\cdot^{2}] - \EE^{2}[\cdot]$ is the variance operator.  For stationary processes, it holds that
\[
\gamma(\bfr, \tau) = C(\bfo, 0) - C(\bfr, \tau)\,.
\]
Hence, the variogram is equivalent to the covariance for stationary processes. However, estimation of the former  is often preferred because (i) the variogram is purely a function of the space-time lags  for processes that are non-stationary but satisfy the \emph{intrinsic hypothesis}---the process $z(\bfs,t)$ is non-stationary but the increments $z(\bfs+\bfr, t + \tau)- z(\bfs, t)$ are stationary---and (ii)  if the mean of $z(\bfs,t)$ is constant but unknown, the sample-based variogram is an unbiased estimator while the covariance is not.

The variogram fitting is performed by means of the approximate weighted least squares method~\cite[Eq.~(2.6.12]{Cressie91}. The  Matlab$\circledR$ constrained minimization function \texttt{fmincon} is employed using the interior point algorithm, a maximum of $10^4$ function evaluations and iterations, and  tolerances equal to $10^{-4}$.
We first fit the marginal spatial and temporal variograms~\cite{Cesare01,dth17} to obtain initial estimates of the hyperparameters followed by a fit of the full LDHO kernel to the space-time variogram.  The main steps of the estimation procedure are as follows:

\begin{enumerate}

\item The \emph{spatial omnidirectional marginal variogram} is estimated from the data for $N_{c;S}$ spatial classes by averaging the spatial variograms obtained for each time instant as follows:
\begin{subequations}
\label{eq:spatial-vario-estim}
\begin{align}
\hat{\gamma}_{t}(r_{k}) & = \frac{1}{2N(r_{k})}\, \sum_{j=1}^{N_{S}}\, \sum_{j=1}^{N_{S}}\,  \mathbb{I}\left( A_{r_{k};i,j} \right)  \,
 \left[ z(\bfs_{i}, t) -  z(\bfs_{j},t) \right]^2\,,\; k=1, \ldots, N_{c;S}\,, \;\; t=1, \ldots, N_{T}\,,
 \\
 A_{r;i,j} & = \mathrm{true}\,   \; \textrm{if} \; r - \delta r \le  \lVert\bfs_{i} - \bfs_{j} \rVert \le r + \delta r\, \textrm{and} \; A_{r;i,j}= \mathrm{false}\, \; \textrm{otherwise},
 \\
\hgms(r_{k}) & = \frac{1}{N_{T}} \, \sum_{t=1}^{N_{T}}\hat{\gamma}_{t}(r_{k}) \; .
\end{align}
\end{subequations}
$\mathbb{I}(A)$ is the indicator function: $\mathbb{I}(A)=1$ if A is true and $\mathbb{I}(A)=0$ if A is false, while $\delta r$ is the tolerance of the spatial lag (all lags in $[r-\delta r, r+\delta r]$ are considered in the lag bin associated with $r$).  $N_{S}= 64^{2}$ is the number of grid nodes per time slice, while $N_{T}=128$ is the number of time slices. $N(\bfr_{k})=\sum_{j=1}^{N_{S}}\, \sum_{j=1}^{N_{S}}\,  \mathbb{I}\left( A_{r_{k};i,j} \right)$ is the number of node pairs that contribute to the lag $r_{k}$, $k=1, \ldots, N_{c;S}$.

\item The spatial marginal variogram is fitted to the respective LDHO marginal model using the method of weighted least squares. This leads to estimates for the hyperparameters $\hat{c}_{0}$, $\hat{\epsilon}$, and an uncorrelated noise variance, $\hat{\sigma}^{2}_{\eta;S}$.

\item Next, we estimate the  \emph{temporal  marginal variogram} for $N_{c;T}$ temporal classes based on a spatial averaging of the temporal variograms per location, i.e.,
\begin{subequations}
\label{eq:temporal-vario-estim}
\begin{align}
\hat{\gamma}_{i}(\tau_m) & =  \frac{1}{2 (N_{T}-m)} \, \sum_{\ell=1}^{N_{T}-m} \,  \left[   z(\bfs_{i}, t_{\ell+m}) -  z(\bfs_{i}, t_{\ell}) \right]^{2}, \;  i=1, \ldots, N_{S}; \; m=1, \ldots N_{c;T}\,,
\\
{\hgmt}(\tau_m) & =   \frac{1}{N_{S}} \,\sum_{i=1}^{N_{S}} \hat{\gamma}_{i}(\tau_{m})\,,
\end{align}
\end{subequations}
where $N_S$ and $N_T$ are respectively the number of sampling points per time slice, and the number of sampling times at each location.

\item The temporal marginal variogram is fitted to the respective LDHO marginal model.   This leads to estimates for the hyperparameters $b$, $\omega_{d}$, $\tau_{c}$ and an uncorrelated noise variance, $\hat{\sigma}^{2}_{\eta;T}$.

\item We then estimate the space-time variogram based on
\beq
\label{eq:estim-st-variogram}
\hat{\gamma}(\bfr_{k}, \tau_{m}) = \frac{1}{N_{k,m}}
\sum_{i=1}^{N_{S}}\sum_{j=1}^{N_{S}}\sum_{\ell=1}^{N_{T}-\ell} \mathbb{I}\left( A_{r_{k};i,j} \right)  \, \left[   z(\bfs_{i}, t_{\ell+m}) -  z(\bfs_{j}, t_{\ell}) \right]^{2}, \;
\eeq
where $N_{k,m}=(N_{T}-m) \,N(r_{k})$.

\item We fit the estimate $\hat{\gamma}(\bfr_{k}, \tau_{m})$ to the LDHO model using as initial estimates for the hyperparameters the estimates derived from the marginal variograms---for the noise variance which is estimated from both the temporal and spatial marginal variograms, we use $\min(\hat{\sigma}^{2}_{\eta;T}\, , \hat{\sigma}^{2}_{\eta;S})$.

\end{enumerate}

\subsection{Parameter estimation for the synthetic data}
For the underdamped LDHO model, the marginal variograms are given by the equations below.

\underline{Spatial marginal variogram:}
\beq
\label{eq:vario-ldho-under-marginal-s}
\gms(r) = c_{1}  \, \left( 1 - \E^{-r^2/4\epsilon} \right) + \hat{\sigma}^{2}_{\eta;S}\,\mathbb{I}(r>0),  \quad  c_{1} = \frac{c_{0}}{(4\pi \epsilon)^{d/2}}\,,
\eeq

\underline{Temporal marginal variogram:}
\begin{align}
\label{eq:vario-ldho-under-marginal-t}
& \gmt(\tau) =  \frac{c_{1}\, \left\{ 1 - \E^{-\frac{\lvert \tau\rvert}{2\tilde{\tau}_{c}}}\, \left[  \cos\left(\tilde{\omega}_{d}\lvert \tau \rvert + \tfrac{d\phi}{2}\right)
+ \tfrac{1}{2\,\tilde{\omega}_{d} \tilde{\tau}_{c}}
\sin\left(\tilde{\omega}_{d}\lvert \tau \rvert + \tfrac{d\phi}{2}\right) \right] \right\}}{ \left[ \left( \frac{b\lvert\tau\rvert}{2\tilde{\tau}_{c}\epsilon} + 1 \right)^{2} + \left(\frac{b\,\tilde{\omega}_{d}\,\lvert\tau\rvert}{\epsilon}\right)^{2} \right]^{d/4}}  + \hat{\sigma}^{2}_{\eta;T}\,\mathbb{I}( \lvert\tau \rvert > 0) \,,
\end{align}
where $\phi = \atan\left( \frac{-2b\,\tilde{\omega}_{d}\,\lvert\tau\rvert\tilde{\tau}_{c}}{b\,\lvert\tau\rvert+
2\epsilon \tilde{\tau}_{c}} \right)\,$ and $\hat{\sigma}^{2}_{\eta;T}$, $\hat{\sigma}^{2}_{\eta;S}$ are noise variances. The  hyperparameters are defined in Section~V.C.1 of the main manuscript.

\begin{figure}
\centering
\includegraphics[width=0.49\textwidth]{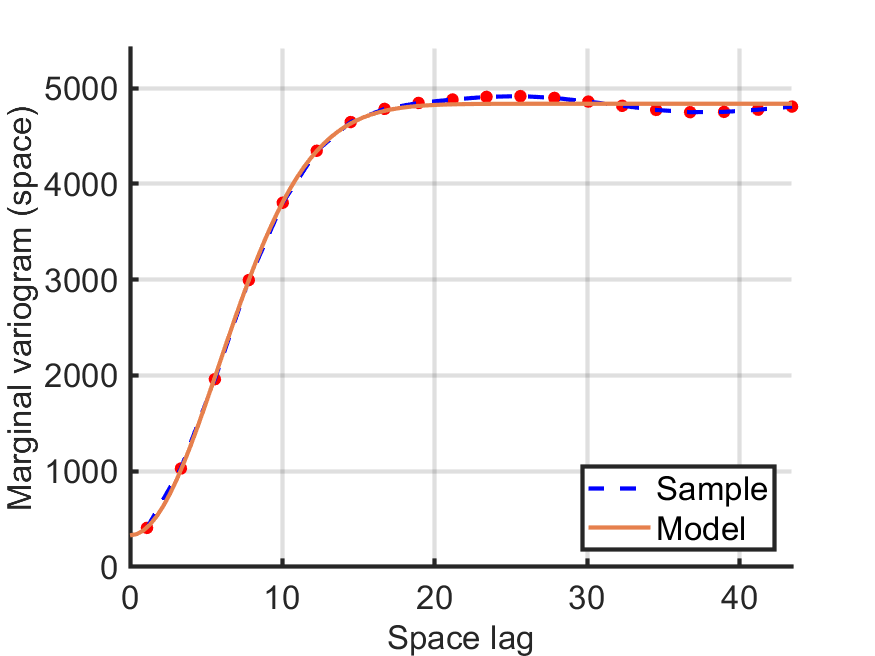}
\includegraphics[width=0.49\textwidth]{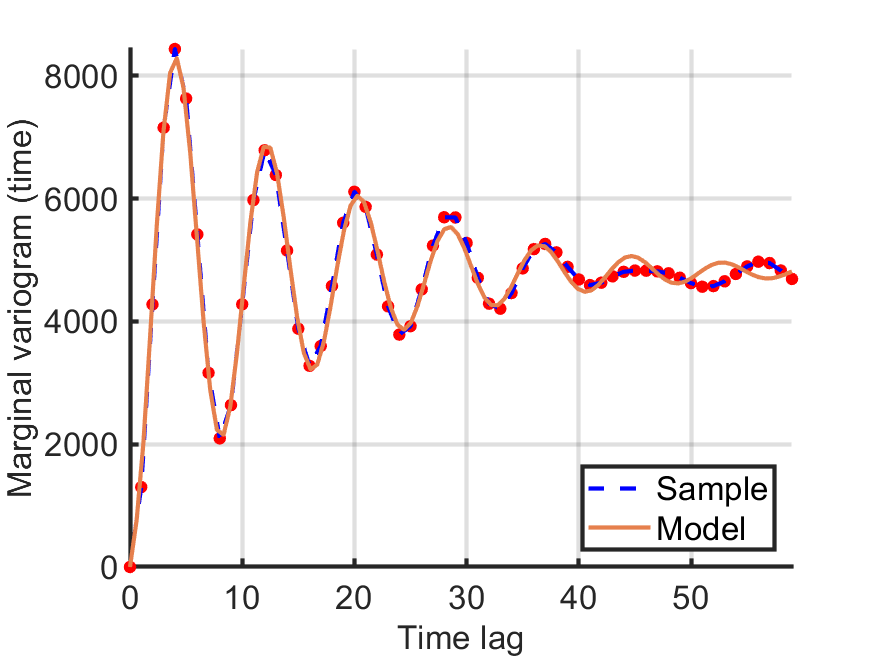}  \caption{Marginal spatial (left) and temporal (right) variograms. The broken line (blue online) and circle markers denote the sample-based variograms while the continuous line (orange online) denotes the optimal fit to the LDHO marginal variograms. }
\label{fig:marginal-variograms}
\end{figure}

The fits between the sample-based marginal variograms obtained from~\eqref{eq:spatial-vario-estim}-\eqref{eq:temporal-vario-estim} and the respective theoretical models~\eqref{eq:vario-ldho-under-marginal-s}-\eqref{eq:vario-ldho-under-marginal-t} are shown in Fig.~\ref{fig:marginal-variograms}. The plots reveal excellent agreement between the theoretical values  and the sample estimates. The estimates of the LDHO kernel hyperparameter vector $\bmthe_{0}$ based on the marginal kernels is:
\beq
\label{eq:theta0}
\hat{c}_{1}=450.75\times 10^3,  \,
\hat{\epsilon}=17.0, \, \hat{\sigma}^{2}_{\eta;S}=330.4,
\hat{\omega}_{d}=0.78, \, \hat{\tau}_{c}=8.04, \, \hat{b}=0.19, \,
\hat{\sigma}^{2}_{\eta;S}=305.59\,.
\eeq
The space-time sample-based variogram is then estimated using~\eqref{eq:estim-st-variogram}.   Using $\bmthe_0$  as initial values in the constrained minimization procedure, we obtain the following estimates for the LDHO kernel hyperparameter vector $\bmthe^\ast$:
\beq
\label{eq:theta}
{c}_{1}^{\ast}=444.19\times 10^3,  \,
{\epsilon}^{\ast}=14.51, \,
{\omega}_{d}^{\ast}=0.75, \, {\tau}_{c}^{\ast}=5.24, \, {b}^{\ast}=0.44, \,
{\sigma_{\eta}^{\ast}}^{2}=366.71\,.
\eeq

\begin{figure}
\centering
\includegraphics[width=0.49\textwidth]{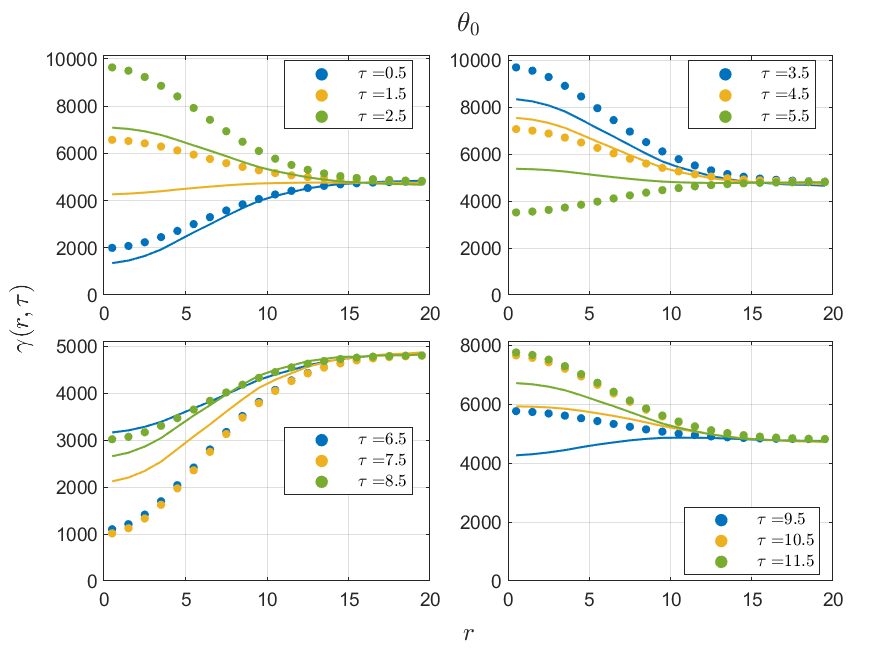}
\includegraphics[width=0.49\textwidth]{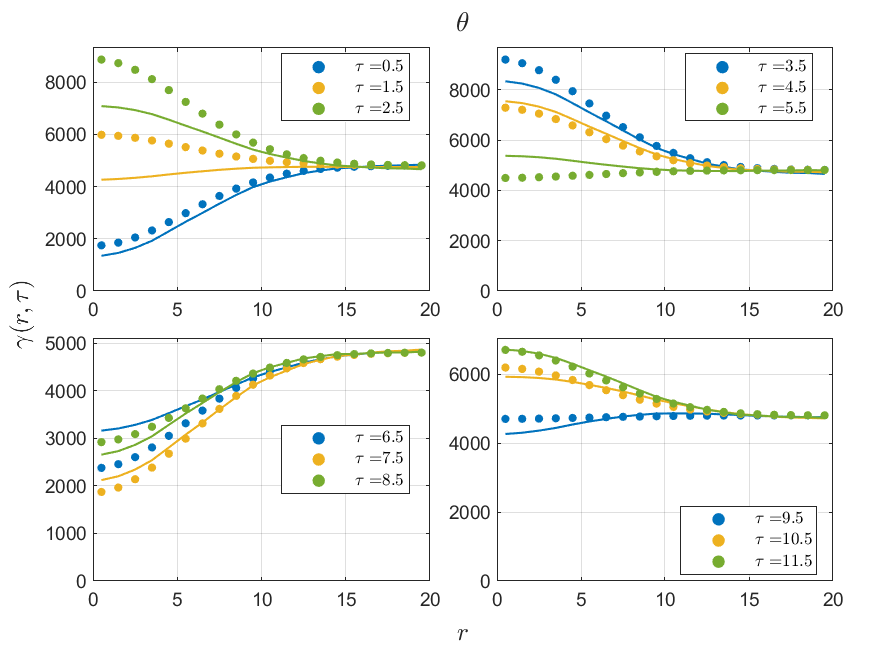}
\caption{Space-time variogram curves for fixed temporal lags. The two columns on the left compare the sample-based variograms (continuous lines) with the respective curves obtained from the theoretical expression (circle markers) using the initial hyperparameter estimates (based on fitting the marginal functions). The respective columns on the right represent the same comparison, but the theoretical expressions are used with the hyperparameters obtained by fitting the full space-time variogram obtained by~\eqref{eq:estim-st-variogram}. The horizontal axis in the plots represents the spatial lag.}
\label{fig:st-variograms-constant-tau}
\end{figure}

\begin{figure}
\centering
\includegraphics[width=0.49\textwidth]{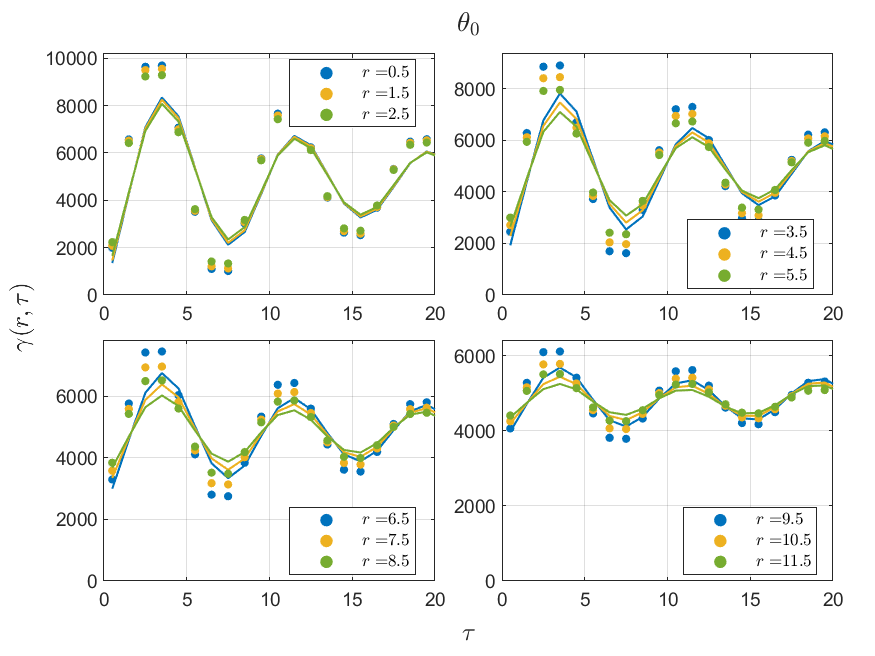}
\includegraphics[width=0.49\textwidth]{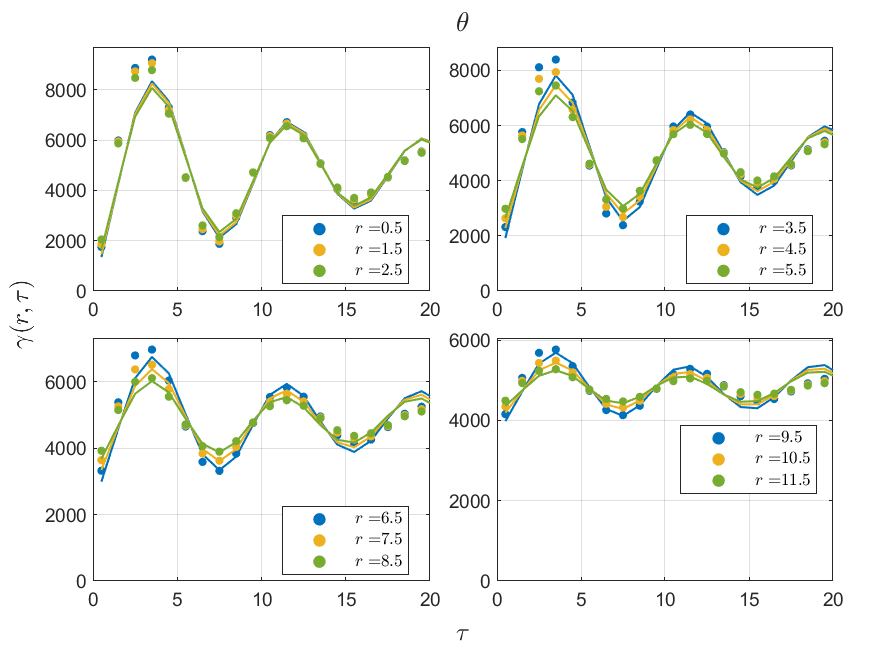}
\caption{Space-time variogram curves for fixed spatial lags. The two columns on the left compare the estimated variograms (continuous lines) with the respective curves obtained from the theoretical expression (circle markers) using the initial hyperparameter estimates (derived by fitting the marginal functions). The respective columns on the right represent the same comparison, but the theoretical expressions are used with the hyperparameters obtained by fitting the full space-time variogram. The horizontal axis in the plots represents time lags. }
\label{fig:st-variograms-constant-r}
\end{figure}

We illustrate the fitness of the kernel hyperparameters by plotting the sample-based variogram~\eqref{eq:estim-st-variogram} against the theoretical expression corresponding to~\eqref{eq:ldho-kernel-st-under}.  To ease the comparison, we use parametric plots obtained first by keeping $\tau$ fixed (cf. Fig.~\ref{fig:st-variograms-constant-tau}) and then by keeping $r$ fixed (cf. Fig.~\ref{fig:st-variograms-constant-r}). The first two columns of each figure compare the estimated space-time variogram curves with the respective theoretical expressions derived from~\eqref{eq:ldho-kernel-st-under} equipped with the initial parameter vector $\bmthe_{0}$~\eqref{eq:theta0}. The last two columns repeat the comparison using $\bmthe$ from~\eqref{eq:theta0}.  Each curve in Fig.~\ref{fig:st-variograms-constant-r} corresponds to a different fixed $\tau$, while the curves in Fig.~\ref{fig:st-variograms-constant-tau} correspond to different fixed $r$. As evidenced in these plots, improved agreement between the sample-based and the theoretical variogram is obtained by using $\bmthe$ instead of $\bmthe_{0}$.
Overall, there are more discrepancies between the theoretical model and the sample-based variogram than in the case of the marginal variograms.  This is due to the fact that sample-based marginal variograms are subject to smoothing caused by averaging over multiple time slices or spatial locations.

\section*{S2. The role of the interaction ratio on prediction}
\label{sec:qinteraction}
In Gaussian process regression, the prediction at an unmeasured point $(\bfs_{\ast} , t_{\ast})$ is determined by the conditional mean~\cite{Rasmussen06,dth20} which involves the covariance kernel.  Let us consider the following thought experiment: We assume a stationary process $z(\bfs,t)$ with known constant mean $m$ and covariance model $C(r,\tau)$. We aim to predict the values of the process $(\bfs_{\ast} , t_{\ast})$ assuming that there is only one sampling point, $(\bfs_{0} , t_{0})$, in the ``neighborhood'' of $(\bfs_{\ast} , t_{\ast})$ (this means that other sampling points are sufficiently far in space-time to consider their influence negligible.  Then, the forecast can be expressed as~\cite{Chiles12,dth20}
\beq
\hat{z}_{\ast}= m + \frac{C(\lVert \bfs_{\ast} - \bfs_{0}\rVert, \tau_{\ast}-\tau_0)}{C(0,0)}\,\, \left( z_{0}  - m \right)\,,
\eeq
where $\hat{z}_{\ast}$ is the conditional mean at $(\bfs_{\ast} , t_{\ast})$, $z_0$ is the sample value at $(\bfs_{0} , t_{0})$, and  $C(\lVert \bfs_{\ast} - \bfs_{0}\rVert, \tau_{\ast}-\tau_0)$ is the non-separable LDHO kernel.

The predictive equation  can also be expressed as
\beq
\label{eq:condm-nonsep}
\hat{z}'_{\ast}= \frac{C(\lVert \bfs_{\ast} - \bfs_{0}\rVert, \tau_{\ast}-\tau_0)}{C(0,0)}\,\,  {z'}_{0}\,,
\eeq
where $\hat{z}'_{\ast}=\hat{z}_{\ast}-m$ and ${z'}_{0}={z}_{0}-m$ are fluctuations of the process around  the global mean.

Let us now consider that instead of using the non-separable covariance kernel, we use the separable model
$K(r,\tau)=K_{1}(r)\,K_{2}(\tau)$. Furthermore, we assume that the separable model has the same marginal kernels as the non-separable model, i.e., $K_{S}(r)=\Cms(r)$ and $K_{T}(\tau)=\Cmt(\tau)$.  The following relations hold between  the separable model and the marginals of the non-separable model:
\[
\Cms(r) \triangleq K(r,0) = K_{2}(0)\, K_{1}(r)\,,
\]
\[
\Cmt(\tau) \triangleq K(0,\tau) = K_{1}(0)\, K_{2}(\tau)\,,
\]
Thus, the product of the non-separable model's marginal kernels is given by
\beq
\label{eq:marginals-sep}
\Cms(r) \Cmt(\tau) =K_{1}(0)\,K_{2}(0)  K_{1}(r)\,K_{2}(\tau) = C(0,0)\,K_{1}(r)\,K_{2}(\tau)\,,
\eeq
where in deriving the above we took into account that $C(0,0)=K(0,0)$.

The predictive equation for the separable model becomes
\beq
\tilde{z}'_{\ast}= \frac{K(\lVert \bfs_{\ast} - \bfs_{0}\rVert, \tau_{\ast}-\tau_0)}{K(0,0)}\,\,  {z'}_{0} = \frac{K_{1}(\lVert \bfs_{\ast} - \bfs_{0}\rVert)\, K_{2}(\tau_{\ast}-\tau_0)}{C(0,0)}\,\,  {z'}_{0}\,,
\eeq
where $\tilde{z}'_{\ast}$ is the conditional mean under the separable model.
Then, using~\eqref{eq:marginals-sep} it follows that
\beq
\label{eq:condm-sep}
\tilde{z}'_{\ast}=  \frac{\Cms(\lVert \bfs_{\ast} - \bfs_{0}\rVert)\, \Cmt(\tau_{\ast}-\tau_0)}{C^{2}(0,0)}\,\,  {z'}_{0}\,.
\eeq
Finally, assuming that ${z'}_{0} \neq 0$, it follows from~\eqref{eq:condm-nonsep} and~\eqref{eq:condm-sep} that the ratio of the two predictors (i.e., $\hat{z}'_{\ast}$ for the non-separable model over $\tilde{z}'_{\ast}$ for the separable model) is given by
\beq
\label{eq:pred-ratio}
\frac{\hat{z}'_{\ast}}{\tilde{z}'_{\ast}} =
\frac{C(\lVert \bfs_{\ast} - \bfs_{0}\rVert, \tau_{\ast}-\tau_0)\, C(0,0)}{\Cms(\lVert \bfs_{\ast} - \bfs_{0}\rVert)\, \Cmt(\tau_{\ast}-\tau_0)}\, = Q_{\mathrm{int}}(\lVert \bfs_{\ast} - \bfs_{0}\rVert, \tau_{\ast}-\tau_0)\,.
\eeq
Hence, the \emph{interaction ratio}  $Q_{\mathrm{int}}(\lVert \bfs_{\ast} - \bfs_{0}\rVert, \tau_{\ast}-\tau_0)$ determines the relative change obtained by introducing a non-separable kernel with space-time interactions to the prediction obtained from the separable kernel. Note that the LDHO marginal temporal kernel given by~(26) in the main text has a harmonic dependence that   involves $\tau$ but not $r$, while the space-time kernel given by~(23) in the main manuscript involves a harmonic term that depends on both $r$ and $\tau$.

\begin{figure}
\centering
\includegraphics[width=0.9\textwidth]{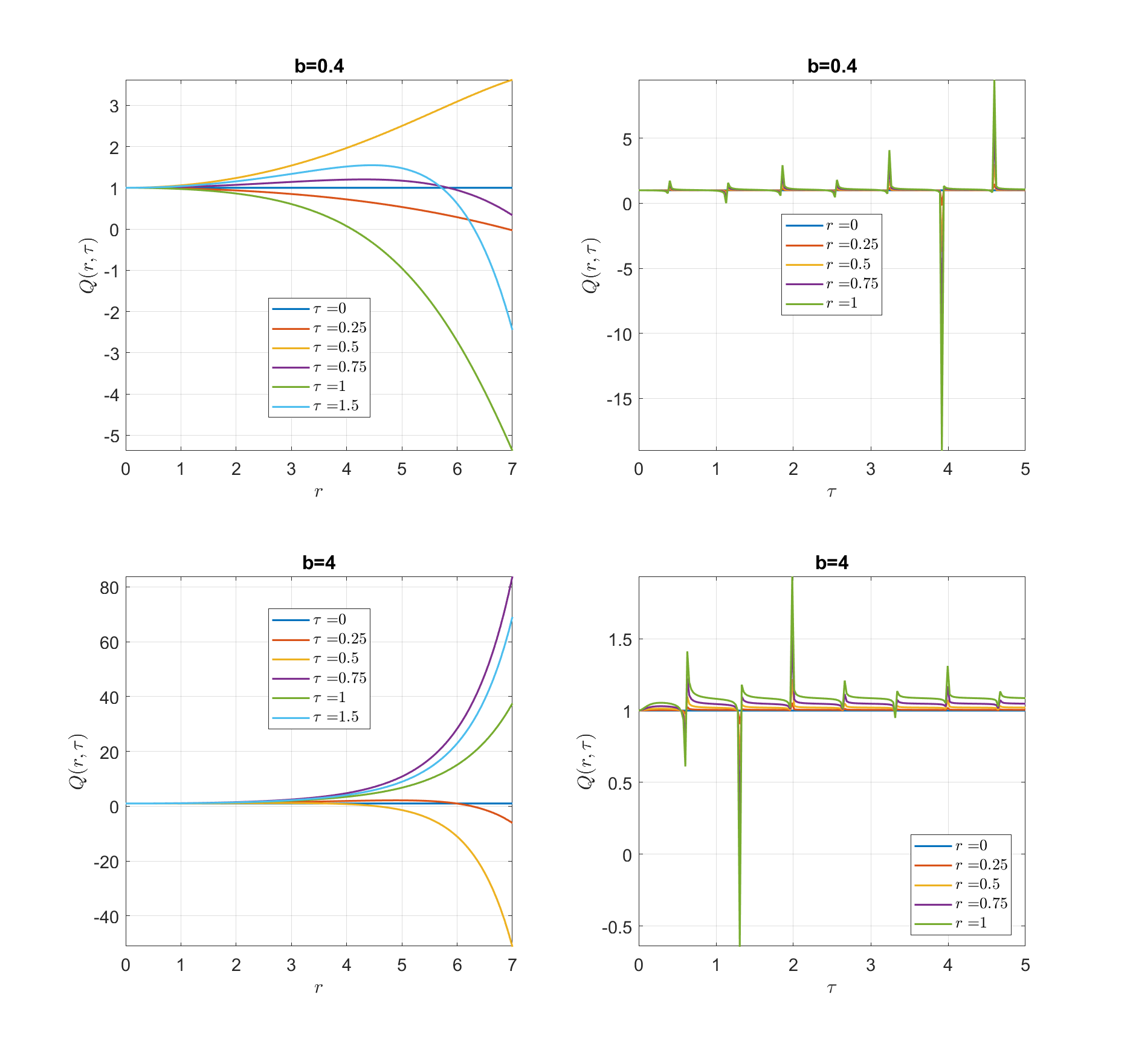}
\caption{Parametric plots of the interaction ratio for the LDHO kernel in the underdamped regime with two different values of the interaction parameter $b=0.4$ (top) and $b=4$ (bottom). The left column presents $Q_{\mathrm{int}}(r,\tau)$ as functions of $r$ for fixed $\tau$, while the right column presents plots of $Q_{\mathrm{int}}(r,\tau)$ as functions of $\tau$ for fixed $r$. The other hyperparameter values are: $d=2$,  $\omega_d = 3\pi/2$, $\tau_c = 2$, $\epsilon = 3$. }
\label{fig:Qint-under}
\end{figure}

The dependence of the interaction ratio on the hyperparameter values and the space/time lags is illustrated in Fig.~\ref{fig:Qint-under} which shows parametric plots of  $Q_{\mathrm{int}}(r,\tau)$ for the LDHO kernel in the underdamped regime.  Considering the interaction ratio as a function of $r$ under constant $\tau$, we observe that values of larger magnitude tend to appear for increasing $r$; both the sign and magnitude of $Q_{\mathrm{int}}(r , \tau_{i})$ for fixed $\tau_{i}$ depend on the value of the interaction parameter $b$.    On the other hand, the plots $Q_{\mathrm{int}}(r_{i} , \tau)$ display characteristic peaks at certain times which are caused by near-zero values of the LDHO marginal temporal kernel.

\setcounter{subsection}{0}
\section*{S3. LDHO Covariance Kernel: Model with  Linear $B(k)$ and Exponential Decay of $A(k)$}
\label{sec:ldho-exponential-app}

\subsection{General}

In order to construct the spatiotemporal kernels by means of the hybrid spectral matching approach, we use temporal Fourier modes, $\tC_{-\omega}(\bfk,\tau;\bmthe)$ which are  obtained from the purely temporal kernels, derived from the respective ODE, by replacing the constant coefficients with suitable dispersion functions.

The spatiotemporal kernel is then given by means of the inverse Fourier transform~\eqref{eq:invcovft-iso}.
In the LDHO case the temporal kernels are given  in each regime by means of the functions~\eqref{eq:cov-ldho}.

The general form of the dispersion functions for radial dependence is given by
\begin{align}
\label{eq:dispersion-lin}
& \sigma^{2} \to \sigma^{2}(k) = \sigma_{0}^{2} A(k), \; \tau_{c} \to \tau_{c}(k)= \frac{\tilde{\tau}_{c}}{B(k)}, \; \omega_{0} \to \omega_{0}(k)=\tilde{\omega}_{0}\, B(k),
\end{align}
where $\tilde{\omega}_{0}, \tilde{\tau}_{c} \in \R_{+}$,  $A(k), B(k) >0$ for all $k>0$.

\begin{definition}[Dispersion functions with linear in $k$ dependence]
Let us assume that $A(\bfk)$ and $B(\bfk)$ are given by the following radial dispersion functions:
\begin{subequations}
\label{eq:dispersion-fun-lin}
\begin{align}
\label{eq:lin-B1}
B(k)= & 1 + \xi k, \; \xi > 0,
\\
\label{eq:lin-A1}
A(k) = & \E^{-\epsilon k} \, B(k), \; \epsilon > 0.
\end{align}
\end{subequations}
\end{definition}
The function $B(k)$ in~\eqref{eq:lin-B1} implies that  $\tau_{c}(k) \sim k$,  whereas  $\omega_{0}(k) \sim k$ for $k \to \infty$.  The function $A(k)$ defined in~\eqref{eq:lin-A1}   is dominated by the exponential decay, implying a rapid decrease of  $\sigma^{2}(k)$ for $k \to \infty$.    The choice $A(k)=\E^{-\epsilon k}B(k)$  leads to  $A(k)/B(k)=\E^{- \epsilon k}$.

The resulting spatiotemporal LDHO covariance kernels for each regime are derived below.  For notational convenience the hyperparameter $c_{0} \triangleq {\sigma^{2}_{0}}/{2\tilde{\tau}_{c} \tilde{\omega}^{2}_{0}}$ is introduced.

\medskip

\subsection{Underdamping}
This regime  is obtained for $\tilde{\omega}_{0} \tilde{\tau}_{c}>1/2$.
The temporal Fourier modes, based on~\eqref{eq:cov-ldho-u} and the dispersion relations, are given by
\beq
\label{eq:cov-ldho-tft-u}
\tC_{-\omega}(k,\tau) = {c_0}  \, \E^{ -\lvert \tau \rvert\, \left( 1+\xi k\right)/2\tilde{\tau}_{c} - \epsilon \,k}  \, \left[ \cos\left( \tilde{\omega}_{d}\left(1+\xi k\right) \tau\right)  + \frac{\sin\left(\tilde{\omega}_{d} \left( 1 + \xi k\right)\lvert \tau \rvert \right)}{2\tilde{\omega}_{d}\tilde{\tau}_{c}}  \right]\,.
\eeq
\smallskip

\begin{theorem}[LDHO  kernel in underdamped regime]
\label{theo:ldho-kernel-st-under-lin}
In this case, the LDHO spatiotemporal kernel generated by the IFT of the temporal Fourier modes~\eqref{eq:cov-ldho-tft-u}. This leads to the following expressions:
\begin{subequations}
\label{eq:ldho-kernel-st-under-lin}
\begin{align}
\label{eq:ldho-kernel-st-under-lin-C}
& C(r,\tau)  =  c_{0}  \,  \E^{-\frac{\lvert \tau\rvert}{2\tilde{\tau}_{c}}} \left[ F_{1}(r,\tau) + F_{2}(r,\tau)\right],
\\[1ex]
& F_{1}(r,\tau)   = \cos(\tilde{\omega}_{d}\tau) \, \Rei(r,\tau) - \sin(\tilde{\omega}_{d}\lvert\tau\rvert) \, \Imi(r,\tau),
 \\[1ex]
& F_{2}(r,\tau)  = \frac{\sin(\tilde{\omega}_{d}\lvert\tau\rvert )}{2\,\tilde{\omega}_{d} \tilde{\tau}_{c}} \, \Rei(r,\tau) + \frac{\cos(\tilde{\omega}_{d}\lvert\tau\rvert )}{2\,\tilde{\omega}_{d} \tilde{\tau}_{c}} \, \Imi(r,\tau)\, ,
 \\
  \label{eq:ldho-under-gre}
& \Rei(r,\tau)=  \frac{\Gamma(\frac{d+1}{2})}{\pi^{(d+1)/2}} \,\frac{ \sqrt{a_{\Re}^{2} +a_{\Im}^{2}}\, \cos\left( \frac{(d+1)\gamma}{2}-\phi\right)}{\left[\left(a^{2}_{\Re}+a_{\Im}^{2}+r^{2} \right)^{2} - 4 a_{\Im}^{2}\, r^2\right]^{(d+1)/4}} ,
 \\[1ex]
 \label{eq:ldho-under-gim}
& \Imi(r,\tau)= \frac{\Gamma(\frac{d+1}{2})}{\pi^{(d+1)/2}}\, \frac{ \sqrt{a_{\Re}^{2} +a_{\Im}^{2}}\, \sin\left( \frac{(d+1)\gamma}{2}-\phi\right)}{\left[\left(a^{2}_{\Re}+a_{\Im}^{2}+r^{2} \right)^{2} - 4 a_{\Im}^{2}\, r^2\right]^{(d+1)/4}} \,,
\end{align}
where   $r, \tau$ are, respectively, the spatial and  temporal lags. The quantities $a_{\Re}$, $a_{\Im}$, $\gamma$ and $\phi$ are \emph{space-time interaction functions} given by
\beq
\label{eq:ar}
a_{\Re} =
\frac{\xi\,\lvert\tau\rvert}{2\tilde{\tau}_{c}} +\epsilon  \,,
\eeq
\beq
\label{eq:ai}
a_{\Im} =
\xi\,\lvert\tau\rvert\,\tilde{\omega}_{d}  \,,
\eeq
\beq
\label{eq:gamma}
\tan \gamma =
\frac{2a_{\Im}\,a_{\Re}}{a^{2}_{\Re}-a_{\Im}^{2}+r^2}  \,,
\eeq
\beq
\label{eq:phi}
\tan \phi = \frac{a_{\Im}}{a_{\Re}}  \,.
\eeq

\end{subequations}

\end{theorem}

\smallskip

\begin{IEEEproof}
The temporal Fourier modes are obtained from~\eqref{eq:cov-ldho-tft-u}.
\begin{subequations}
\label{eq:tC-under-lin}
\begin{align}
\tC_{-\omega}(k,\tau) &  = c_{0}  \, \E^{-\frac{\lvert \tau\rvert}{2\tilde{\tau}_{c}}} \left[ \tilde{F}_{1}(k)+\tilde{F}_{2}(k)\right],
\\[1ex]
\tilde{F}_{1}(k) & = \E^{ -k\left(\frac{\lvert\tau\rvert\,  \xi}{2\tilde{\tau}_{c}} +\epsilon \right) } \, \cos\big( \tilde{\omega}_{d} (1+k\xi) \lvert\tau\rvert\, \big),
\\[1ex]
\tilde{F}_{2}(k) & =\E^{ -k\left(\frac{\lvert\tau\rvert\,  \xi}{2\tilde{\tau}_{c}} +\epsilon \right) }\, \frac{\sin\big( \tilde{\omega}_{d} (1+k\xi) \lvert\tau\rvert \big)}{2\tilde{\omega}_{d}\tilde{\tau}_{c}}\,.
\end{align}
\end{subequations}

For reasons of brevity, in the following  only the dependence of $\tilde{F}_{i}(\cdot)$ on $k$ (in the Fourier domain) and of $F_{i}(\cdot)$ on $r$  are shown explicitly.  Based on~\eqref{eq:tC-under-lin}, $C(r,\tau)$ is given by the following function
\beq
\label{eq:C-real-under-ok}
C(r,\tau) =  c_{0}  \, \E^{-\frac{\lvert \tau\rvert}{2\tilde{\tau}_{c}}} \left[ F_{1}(r) + F_{2}(r)\right],
\eeq
where $F_{i}(r) = \ift_{\bfk} [\tilde{F}_{i}(k)]$, $i=1,2$. In order to evaluate $F_{i}(r)$  we express the harmonic terms as linear combinations of $\E^{\pm \I x}$
using  Euler's formula $\E^{\I x}= \cos x + \I \sin x$, for $x\in \R$. By defining $a_{\Re} \triangleq \frac{\lvert \tau\rvert\,  \xi}{2\tilde{\tau}_{c}} +\epsilon$, it follows that $a_{\Re} >0$ and the $\tilde{F}_{1}(k)$, $\tilde{F}_{2}(k)$ are given by
\begin{align*}
\tilde{F}_{1}(k) & = \frac{\E^{ -a_{\Re}  k }}{2} \, \left[ \, \E^{\I \left(  \tilde{\omega}_{d}\lvert\tau\rvert + \xi   \tilde{\omega}_{d}\lvert\tau\rvert\, k \right) } + \E^{-\I \left(  \tilde{\omega}_{d}\lvert\tau\rvert + \xi  \tilde{\omega}_{d}\lvert\tau\rvert k \right) } \right],
\\[1ex]
\tilde{F}_{2}(k) & = \frac{\E^{ -a_{\Re}  k }}{4\I \,\tilde{\omega}_{d} \tilde{\tau}_{c}} \, \left[ \E^{\I \left(  \tilde{\omega}_{d}\lvert\tau\rvert + \xi   \tilde{\omega}_{d}\lvert \tau\rvert k \right) } - \E^{-\I \left(  \tilde{\omega}_{d}\lvert \tau\rvert + \xi  \tilde{\omega}_{d}\lvert\tau\rvert k \right) } \right].
\end{align*}

\noindent Hence, the $k$-dependent parts of $\tilde{F}_{1}(k)$, $\tilde{F}_{2}(k)$ comprise the functions  $\tilde{I}_{\pm}(k)=\exp\left[ - \left( a_{\Re} \pm \I \, \xi\,\tilde{\omega}_{d}\,\lvert\tau\rvert \right) k \right]\,$:
\begin{align*}
\tilde{F}_{1}(k) & = \frac{1}{2} \, \left[ \E^{\I  \tilde{\omega}_{d}\lvert\tau\rvert } \, \tilde{I}_{+}(k) + \E^{-\I   \tilde{\omega}_{d}\lvert\tau\rvert }  \, \tilde{I}_{-}(k) \right],
\\[1ex]
\tilde{F}_{2}(k) & = \frac{1}{4\I \,\tilde{\omega}_{d} \tilde{\tau}_{c}} \, \left[ \E^{\I  \tilde{\omega}_{d}\lvert\tau \rvert} \, \tilde{I}_{+}(k) - \E^{-\I   \tilde{\omega}_{d}\lvert\tau\rvert }  \, \tilde{I}_{-}(k) \right].
\end{align*}

\noindent Using the identities $\tilde{I}_{-}(k)=I_{+}^{\dagger}(k)$, $\E^{-\I  \tilde{\omega}_{d}\lvert\tau\rvert }=\left( \E^{\I  \tilde{\omega}_{d}\lvert\tau\rvert }\right)^{\dagger}$, and $z_{1}^{\dagger}z_{2}^{\dagger}=(z_{1}z_{2})^{\dagger}$ for any $z_{1}, z_{2} \in \Co$, the functions $\tilde{F}_{1}(k)$, $\tilde{F}_{2}(k)$ are expressed in terms of the real and imaginary parts of the function $\E^{\I  \tilde{\omega}_{d}\tau } \, \tilde{I}_{+}(k)$, i.e.,

\begin{subequations}
\label{eq:F1k-F2k-lin}
\begin{align}
\label{eq:F1-lin}
\tilde{F}_{1}(k) & = \Re \left[ \E^{\I  \tilde{\omega}_{d}\lvert\tau\rvert } \, \tilde{I}_{+}(k)  \right] = \cos(\tilde{\omega}_{d}\lvert\tau\rvert) \, \Re\left[ \tilde{I}_{+}(k)\right]
 - \sin(\tilde{\omega}_{d}\lvert\tau\rvert)\, \Im\left[ \tilde{I}_{+}(k)\right],
\\[1ex]
\label{eq:F2-lin}
\tilde{F}_{2}(k) & = \frac{1}{2\,\tilde{\omega}_{d} \tilde{\tau}_{c}} \, \Im \left[ \E^{\I  \tilde{\omega}_{d} \lvert\tau\rvert } \, \tilde{I}_{+}(k)  \right]
 =  \frac{\sin(\tilde{\omega}_{d}\lvert\tau\rvert )}{2\,\tilde{\omega}_{d} \tilde{\tau}_{c}} \Re\left[ \tilde{I}_{+}(k)\right] + \frac{\cos(\tilde{\omega}_{d}\lvert\tau\rvert )}{2\,\tilde{\omega}_{d} \tilde{\tau}_{c}} \Im\left[ \tilde{I}_{+}(k)\right] .
\end{align}
\end{subequations}

Since both $\Re\left[ \tilde{I}_{+}(k)\right]$ and $\Im\left[ \tilde{I}_{+}(k)\right]$ are radial functions of $k$, their inverse Fourier transforms are real-valued, radial functions of $r$ according to~\eqref{eq:invcovft-iso}.
Let $I_{+}(r)\triangleq \ift_{\bfk}\left[ \tilde{I}_{+}(k)\right]$ denote the inverse Fourier transform of $\tilde{I}_{+}(k)$.  $I_{+}(r)$ comprises real and imaginary parts denoted by $\Rei(r) \triangleq \Re[I_{+}(r)]$ and
$\Imi(r)\triangleq  \Im[I_{+}(r)]$.  Then,
\begin{align*}
& \ift\left\{\Re[\tilde{I}_{+}(k)]\right\}= \Re\left\{\ift[\tilde{I}_{+}(k)]\right\} = \Rei(r),
\\
& \ift\left\{\Im[\tilde{I}_{+}(k)]\right\}= \Im\left\{\ift[\tilde{I}_{+}(k)]\right\}= \Imi(r)\, .
\end{align*}
Based on  the above IFTs and the spectral functions~\eqref{eq:F1k-F2k-lin}, the inverse Fourier transforms $F_{i}(r)=\ift_{\bfk}[\tilde{F}_{i}(k)]$, where $i=1, 2$ are given by~\eqref{eq:F1-F2-ift}.

The function $I_{+}(r)$ is evaluated by means of the spectral representation~\eqref{eq:invcovft-iso} which involves the following integral
\beq
\label{eq:Iplus-lin}
I_{+}(r) \triangleq \frac{r}{(2\pi r)^{d/2}} \int_{0}^{\infty} \, k^{d/2}
    {J_{d/2-1}(k  r)} \,\E^{- \left( a_{\Re} + \I \, \xi\,\tilde{\omega}_{d}\,\lvert\tau\rvert \right) k} \,\D k \,.
\eeq
Hence, $I_{+}(r)$ can be evaluated using the  lemma~\cite[Eq.~(6.623.2)]{Gradshteyn07}.

\smallskip

\begin{lemma}[Spectral integral for radial functions]
\label{lemma:Jnu-integral-ok}
Let $J_{\nu}(x)$ represent the Bessel function of the first kind of order $\nu \in \Co$, where $\Re(\nu)>-1$.  Furthermore, let $a \in \Co$ be a constant coefficient with $\Re(a)>0$. Then, the
following is true
 \begin{align}
\label{eq:Jnu-lin}
& \int_{0}^{\infty} k^{\nu+1}\, J_{\nu}(r k)\,\E^{-a k} \D k= \frac{2a\,(2r)^{\nu}}{\left(a^{2} + r^{2} \right)^{\nu+3/2}}\,   \frac{\Gamma\left( \nu+\frac{3}{2}\right)}{\sqrt{\pi}}\,.
\end{align}
\end{lemma}

\smallskip

Hence,  in light of Lemma~\eqref{lemma:Jnu-integral-ok} and by setting $\nu = d/2-1$, the function $I_{+}(r)$ defined in~\eqref{eq:Iplus-lin} is given by the following complex-valued expression
\beq
\label{eq:Ir-lin}
I_{+}(r)=\frac{2a\,r\,(2r)^{\nu}}{(2\pi r)^{d/2}\left(a^{2} + r^{2} \right)^{\nu+3/2}}\,   \frac{\Gamma\left( \nu+\frac{3}{2}\right)}{\sqrt{\pi}} = \frac{\Gamma(\frac{d+1}{2})}{\pi^{(d+1)/2}}\, \frac{a}{\left(a^{2} + r^{2} \right)^{(d+1)/2}},
\eeq
where $a = a_{\Re} + \I \,\xi\,\tilde{\omega}_{d}\,\lvert\tau\rvert\ $.
Then, using $\Rei(r) \triangleq \Re[I_{+}(r)]$ and
$\Imi(r)\triangleq  \Im[I_{+}(r)]$ we obtain
\begin{subequations}
\label{eq:Re-Im-Iplus-ok}
\beq
\Rei(r,\tau)=  \frac{\Gamma(\frac{d+1}{2})}{\pi^{(d+1)/2}} \,\frac{ a_{\Re} \, \cos\left( \frac{(d+1)\gamma}{2}\right)+ a_{\Im} \, \sin\left( \frac{(d+1)\gamma}{2}\right)}{\left[\left(a^{2}_{\Re}+a_{\Im}^{2}+r^{2} \right)^{2} - 4 a_{\Im}^{2}\, r^2\right]^{(d+1)/4}}\,,
\eeq
\beq
\Imi(r,\tau)=  \frac{\Gamma(\frac{d+1}{2})}{\pi^{(d+1)/2}} \,\frac{ a_{\Re} \, \sin\left( \frac{(d+1)\gamma}{2}\right) - a_{\Im} \, \cos\left( \frac{(d+1)\gamma}{2}\right)}{\left[\left(a^{2}_{\Re}+a_{\Im}^{2}+r^{2} \right)^{2} - 4 a_{\Im}^{2}\, r^2\right]^{(d+1)/4}}\,,
\eeq
\end{subequations}
where $\tan\gamma = 2a_{\Re}a_{\Im}/(a^{2}_{\Re}- a_{\Im}^{2}+r^2)$.  Equations~\eqref{eq:ldho-under-gre} and~\eqref{eq:ldho-under-gim} follow by defining $\phi=\atan(a_{\Im}/a_{\Re})$.
Finally, the LDHO kernel~\eqref{eq:ldho-kernel-st-under-lin} is obtained by combining~\eqref{eq:C-real-under-ok}, \eqref{eq:F1-F2-ift} and~\eqref{eq:Re-Im-Iplus-ok}.
\end{IEEEproof}

\medskip
 The kernel equations can be cast in a different but equivalent form (see main text) by defining
\beq
\label{eq:g0}
g_{0}(r,\tau)  = \frac{\Gamma(\frac{d+1}{2})}{\pi^{(d+1)/2}} \, \frac{\left(a^{2}_{\Re}+ a^{2}_{\Im}\right)^{1/2}}{ \left[\left(a^{2}_{\Re}+ a^{2}_{\Im}+ r^{2}\right)^{2} - 4a_{\Im}^{2} r^2 \right]^{(d+1)/4}}
\eeq
and using trigonometric identities so that
\begin{subequations}
\label{eq:F1-2}
\begin{align}
F_{1}(r,\tau) & = g_{0}(r,\tau)\, \cos\left(\tilde{\omega}_{d}\tau + \frac{(d+1)\gamma}{2} -\phi\right)\,,
\\
F_{2}(r,\tau) & = \frac{g_{0}(r,\tau)}{2\tilde{\omega}_{d}\tilde{\tau}_{c}}\, \sin\left(\tilde{\omega}_{d}\tau + \frac{(d+1)\gamma}{2} -\phi\right)\,.
\end{align}
\end{subequations}

\medskip

\paragraph{Kernel hyperparameters}
The kernel function~\eqref{eq:ldho-kernel-st-under}  includes five independent  hyperparameters: $c_{0}, \tilde{\tau}_{c}, \tilde{\omega}_{d},  \epsilon, \xi$. The first three have the same physical significance as their counterparts of the LDHO model with $O(k^2)$ dependence of the dispersion relation. The hyperparameter  $\epsilon$ plays a similar role but has length dimensions (instead of length squared). The hyperparameter $\xi$, which replaces $b$,  has dimensions of  length and---like $b$---determines the rate at  which the non-damped resonance frequency increases  and the damping time drops with $k$.  The phase of the oscillatory kernel functions depends on the space-time phase $\gamma(r,\tau)$ and the time-dependent phase $\phi(\tau)$.
\smallskip

\begin{rem}[Kernel dependence on $d$]
$C(r,\tau)$ depends on the spatial dimension $d$ via the  factor ${\Gamma(\frac{d+1}{2})}/{\pi^{(d+1)/2}}$, the phase factor $(d+1)\gamma(r,\tau)/2$, and the denominators in the damped oscillatory functions $\Rei(r,\tau)$ and $\Imi(r,\tau)$---or equivalently of the non-oscillating function $g_{0}(r,\tau)$.
\end{rem}

\begin{figure}[!ht]
\centering
\includegraphics[width=.99\linewidth]{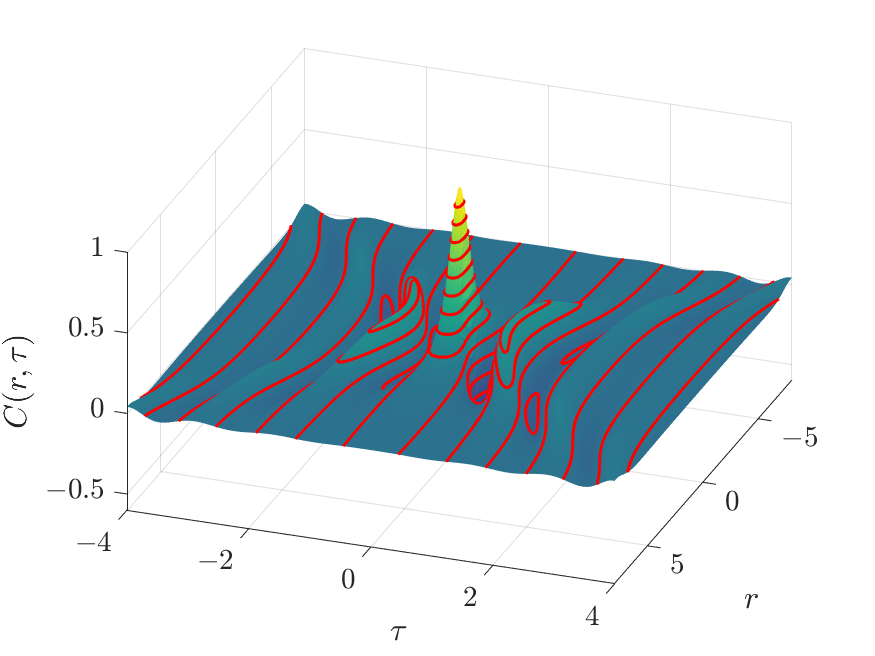}
\caption{Normalized $C(r,\tau)$ and isolevel contour lines (red online) in the underdamped regime, obtained from~\eqref{eq:ldho-kernel-st-under} using
$\tilde{\omega}_{d}=3\pi/2$, $\tilde{\tau}_{c}=3$, $\xi=0.4$,  $\epsilon=1$, and $d=2$.}
\label{fig:kernel_under_linear_1}
\end{figure}

The LDHO covariance kernel is illustrated in Fig.~\ref{fig:kernel_under_linear_1}.
A combination of a relatively slow damping time $\tilde{\tau}_{c}=3$ and a fast oscillation frequency, $\tilde{\omega}_{d}=3\pi/2$, generate four ridges with  decaying amplitude as $\tau$ increases. A smaller value of $\tilde{\tau}_{c}$ (not shown) leads to faster decay and fewer oscillation peaks. The plot also exhibits spatiotemporal interaction, i.e.,  spatial oscillation patterns that appear as ripples on the $(r,\tau)$ plane.

\begin{rem}[Variance decay scale]
The variance decay scale  $\epsilon$ ensures that
the variance $C(r=0,\tau=0)$ is stable.
For $\tau=0$ it holds that $\phi(0)=\gamma(r,0)=0$ because $a_{\Im}=0$.  This also leads to $F_{2}(r,0)=0.$  Hence, $C(r,0)=c_{0}\, F_{1}(r,0)=c_{0}\, g_{0}(r,0)$, where
\beq
\label{eq:g0-r-0}
g_{0}(r,0) = \frac{\Gamma(\frac{d+1}{2})}{\pi^{(d+1)/2}} \, \frac{\epsilon}{\left( \epsilon^{2} + r^{2}\right)^{(d+1)/2}}\,.
\eeq
Hence, $g_{0}(0,0) =\epsilon^{-d} \, \frac{\Gamma(\frac{d+1}{2})}{\pi^{(d+1)/2}}$.
If $\epsilon=0$,  the limit of $C(0, \tau)$ as $\tau \to 0$ does not exist.
\end{rem}

\medskip
\begin{propo}[Spatial marginal covariance]
In the underdamped regime, the  spatial marginal covariance of the LDHO kernel~\eqref{eq:ldho-kernel-st-under}  at $\tau=0$ is given by the square exponential kernel
\beq
\label{eq:ldho-under-marginal-r-ok}
\Cms(r) = c_{0}  \, \frac{\Gamma(\frac{d+1}{2})}{\pi^{(d+1)/2}} \, \frac{\epsilon}{\left( \epsilon^{2} + r^{2}\right)^{(d+1)/2}}.
\eeq
\end{propo}

\begin{IEEEproof}
From~\eqref{eq:ldho-kernel-st-under} for $\tau=0$ it follows that
$\Cms(r)= c_{0} \left[ F_{1}(r,0)+ F_{2}(r,0)\right]$. Furthermore,
$F_{1}(r,0)=g_{0}(r,0)$ and $F_{2}(r,0)=0$.
Using~\eqref{eq:g0-r-0} for $g_{0}(r,0)$ we obtain~\eqref{eq:ldho-under-marginal-r-ok}.
\end{IEEEproof}

\medskip

\begin{propo}[Temporal marginal covariance]
In the underdamped regime, the temporal marginal covariance of the LDHO kernel~\eqref{eq:ldho-kernel-st-under} at  $r=0$ is given by
\beq
\label{eq:C-r0-tau-under-marginal-r-ok}
\Cmt(\tau)= \frac{c_{0} \, \E^{-\frac{\lvert \tau\rvert}{2\tilde{\tau}_{c}}} \, \Gamma(\frac{d+1}{2})}{\pi^{(d+1)/2}} \, \left(a^{2}_{\Re}+ a^{2}_{\Im}\right)^{-d/2}\,
\left[ \cos\left(\tilde{\omega}_{d}\tau + \varphi(\tau)\right) + \frac{1}{2\tilde{\omega}_{d}\tilde{\tau}_{c}}\, \sin\left(\tilde{\omega}_{d}\tau + \varphi(\tau)\right) \right],
\eeq
where $\varphi(\tau)=\frac{(d+1)}{2} \gamma_{0}(\tau)-\phi(\tau)$,
$\tan \gamma_{0}(\tau)=\frac{2a_{\Im}(\tau)a_{\Re}(\tau)}{a^{2}_{\Re}(\tau) - a^{2}_{\Im}(\tau)}$; $a_{\Re}, a_{\Im}$ are defined in~\eqref{eq:ar} and~\eqref{eq:ai} respectively, $\gamma_{0}(\tau)$ in~\eqref{eq:gamma}, and $\phi(\tau)$ in~\eqref{eq:phi}.
\end{propo}

\begin{IEEEproof}
The result is obtained from~\eqref{eq:C-real-under-ok} using~\eqref{eq:g0} and~\eqref{eq:F1-2}.  Setting $r=0$ in $g_{0}(r,\tau)$, the following expression is obtained
\beq
\label{eq:g0-tau-0}
g_{0}(r=0,\tau)  = \frac{\Gamma(\frac{d+1}{2})}{\pi^{(d+1)/2}} \, \left(a^{2}_{\Re}+ a^{2}_{\Im}\right)^{-d/2} \,.
\eeq
Plugging this in~\eqref{eq:F1-2}, the equation~\eqref{eq:F1-2} leads to~\eqref{eq:C-r0-tau-under-marginal-r-ok}.
\end{IEEEproof}

\medskip

\subsection{Overdamping}
This regime  is obtained for $\tilde{\omega}_{0} \tilde{\tau}_{c}<1/2$. The temporal Fourier modes, based on~\eqref{eq:cov-ldho-o} and the dispersion relations, are given by
\beq
\label{eq:cov-ldho-tft-o}
\tC_{-\omega}(k,\tau) = \frac{c_{0}\, \E^{-\epsilon k}}{2\tilde{\omega}_{d} B(k)} \left[ \frac{\E^{ -\frac{\lvert \tau \rvert }{\tau_{s}(k)}}}{\tau_{f}(k)} - \frac{\E^{ -\frac{\lvert \tau \rvert }{\tau_{f}(k)}}}{\tau_{s}(k)}\right]\,.
\eeq

\begin{theorem}[LDHO kernel in overdamped regime]
\label{theo:ldho-kernel-st-over-lin}
The LDHO spatiotemporal kernel is given by
\begin{align}
\label{eq:ldho-kernel-st-over-linear-lin}
C(r,\tau)= &   \frac{c_{0}^\ast \,\beta_{f} \, \E^{-\frac{\beta_{s} \lvert \tau \rvert}{2\tilde{\tau}_{c}}}}{ \left[ \left(\frac{\xi \beta_{s} \lvert \tau \rvert}{2\tilde{\tau}_{c}} + \epsilon\right)^2 + r^2 \right]^{(d + 1)/2}}   -   \frac{c_{0}^\ast \, \beta_{s} \, \E^{-\frac{\beta_{f} \lvert \tau \rvert}{2\tilde{\tau}_{c}}}}{ \left[ \left(\frac{\xi \beta_{f} \lvert \tau \rvert}{2\tilde{\tau}_{c}} + \epsilon\right)^2 + r^2 \right]^{(d + 1)/2}} \,  ,
 \\[1ex]
\text{where} & \;  c_{0}^\ast = \frac{c_{0}\,\Gamma(\frac{d+1}{2})}{4\tilde{\omega}_{d} \,\tilde{\tau}_{c} \,\pi^{(d+1)/2}}, \; \beta_{s}=1- 2\tilde{\tau}_{c} \tilde{\omega}_{d}, \; \beta_{f}=1 + 2\tilde{\tau}_{c} \tilde{\omega}_{d}. \nonumber
\end{align}
\end{theorem}

\begin{figure}[!ht]
\centering
\includegraphics[width=.99\linewidth]{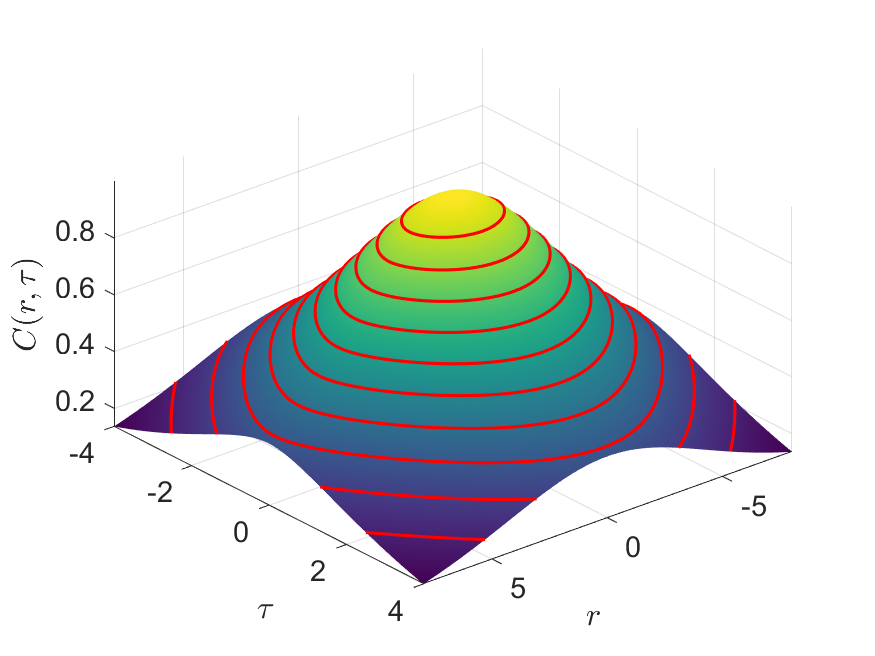}
\caption{Normalized $C(r,\tau)$ and isolevel contour lines (red online) in the overdamped regime, obtained from~\eqref{eq:ldho-kernel-st-over-linear} using
$\tilde{\omega}_{d}=\pi/10$, $\tilde{\tau}_{c}=0.8$, $\xi=0.4$,  $\epsilon=8$, and $d=2$.}
\label{fig:kernel_over_linear_1}
\end{figure}

\begin{IEEEproof}
The temporal Fourier modes of the LDHO kernel are obtained from~\eqref{eq:cov-ldho-o} leading to~\eqref{eq:cov-ldho-tft-o}.
In light of~\eqref{eq:fast-slow} and taking account  the dispersion relations, the fast and slow decay times transform as follows
\begin{subequations}
\label{eq:ts-tf-lin}
\begin{align}
\tau_{s}(k) = & \frac{2\tilde{\tau}_{c}}{B(k) \left(1- 2\tilde{\tau}_{c} \tilde{\omega}_{d} \right)}, \;
\\
\tau_{f}(k) = & \frac{2\tilde{\tau}_{c}}{B(k) \left(1 + 2\tilde{\tau}_{c} \tilde{\omega}_{d} \right)}\,.
\end{align}
\end{subequations}
Based on~\eqref{eq:ts-tf-lin},  the functions $B(k)$ in~\eqref{eq:cov-ldho-tft-o} cancel out, and  the temporal Fourier modes are given by
\beq
\tC_{-\omega}(k,\tau) = \frac{c_{0}\, \E^{-\epsilon k}}{4\tilde{\omega}_{d} \tilde{\tau}_{c} } \left[ \beta_{f} \,\E^{ -\frac{\beta_{s} \lvert \tau \rvert B(k)  }{2\tilde{\tau}_{c}}}  - \beta_{s} \, \E^{ -\frac{\beta_{f} \lvert \tau \rvert B(k) }{2\tilde{\tau}_{c}}} \right],
\eeq
where $\beta_{s}=1- 2\tilde{\tau}_{c} \tilde{\omega}_{d}$ and $\beta_{f}=1 + 2\tilde{\tau}_{c} \tilde{\omega}_{d}$.
Recalling~\eqref{eq:B1} for $B(k)$, the IFT expression~\eqref{eq:invcovft-iso}, and the linearity of the IFT, it follows that
\begin{align}
\label{eq:C-over-lin}
C(r,\tau)= & \frac{c_{0}}{4\tilde{\omega}_{d} \tilde{\tau}_{c} } \left[ \beta_{f}  C_{s}(r,\tau)
-  \beta_{s}  C_{f}(r,\tau)\,  \right],
\end{align}
where
\beq
\label{eq:Cj-r-o-lin}
C_{j}(r,\tau) = \E^{ -\frac{\beta_{j} \lvert \tau \rvert}{2\tilde{\tau}_{c}}} \, \ift\left[\,\E^{ -\frac{\xi\,\beta_{j} \lvert \tau \rvert \,k }{2\tilde{\tau}_{c}}  -\epsilon k}  \right] =
\frac{\E^{ -\frac{\beta_{j} \lvert \tau \rvert}{2\tilde{\tau}_{c}}}}{(2\pi)^{d/2}\, r^\nu} \int_{0}^{\infty}\D k k^{d/2}\,J_{\nu}(kr)\,\E^{-a_{j}k}\,, \; j=s, f\,,
\eeq
where $a_{j}=\frac{\beta_{j} \xi \lvert \tau \rvert}{2\tilde{\tau}_{c}} + \epsilon$.
The integral above can be calculated using~\cite[6.623.2]{Gradshteyn07}:
\beq
\label{eq:I-linear-over}
\int_{0}^{\infty}\D k \, k^{d/2}\,J_{\nu}(kr)\,\E^{-a_{j}k} =\frac{2a_{j}(2r)^{\nu}\,\Gamma(\nu+3/2)}{\sqrt{\pi}\left( a^{2}_{j}+ r^2\right)^{(d+1)/2}}\,, \; \textrm{where} \; \nu=d/2 -1\,.
\eeq

\noindent In view of the above, the functions $C_{j}(r,\tau)$ in~\eqref{eq:Cj-r-o-lin} are given by
\beq
\label{eq:Cj-r-over-lin}
C_{j}(r,\tau) =  \E^{ -\frac{\beta_{j} \lvert \tau \rvert}{2\tilde{\tau}_{c}}} \,
\frac{\Gamma(\frac{d+1}{2})}{\pi^{(d+1)/2}}\,
\frac{1}{\left( a^{2}_{j} + r^2\right)^{(d+1)/2}}
\,.
\eeq

Finally, the overdamped LDHO covariance kernel~\eqref{eq:ldho-kernel-st-over-linear} is obtained from~\eqref{eq:C-over-lin} and~\eqref{eq:Cj-r-over-lin}.
\end{IEEEproof}
An example of the overdamped kernel $C(r,\tau)$  is shown in Fig.~\ref{fig:kernel_over_linear_1}.
\smallskip
\begin{rem}[Variance stablilization]
\label{rem:over-lin}
As in the underdamped case,  the spectral decay hyperparameter $\epsilon$ stabilizes the variance (i.e., the behavior at $\tau=0$),  and $\xi$ adjusts the space-time interaction since for $\xi=0$ the space and time dependence in~\eqref{eq:ldho-kernel-st-over-linear} decouple.
\end{rem}
\smallskip

\paragraph{Zero-lag marginal covariances}
The spatial and temporal marginal kernels  are  obtained from~\eqref{eq:ldho-kernel-st-over-linear} by setting $\tau=0$ and $r=0$ respectively, following simple algebraic calculations. Thus we obtain

\begin{align}
\label{eq:ldho-over-linear-marginal-r}
 \Cms(r) =& \frac{c_{0}^\ast \,\left( \beta_{f} -\beta_{s} \right)\, }{ \left(  \epsilon^2 + r^2 \right)^{(d + 1)/2}} ,
\end{align}
\begin{align}
\label{eq:ldho-over-linear-marginal-t}
 \Cmt(\tau) = &  \frac{c_{0}^\ast \,\beta_{f} \, \E^{-\frac{\beta_{s} \lvert \tau \rvert}{2\tilde{\tau}_{c}}}}{ \left(\frac{\xi \beta_{s} \lvert \tau \rvert}{2\tilde{\tau}_{c}} + \epsilon\right)^{d + 1}}   -   \frac{c_{0}^\ast \, \beta_{s} \, \E^{-\frac{\beta_{f} \lvert \tau \rvert}{2\tilde{\tau}_{c}}}}{ \left(\frac{\xi \beta_{f} \lvert \tau \rvert}{2\tilde{\tau}_{c}} + \epsilon\right)^{d + 1}}  \,.
\end{align}
The spatial marginal kernel~\eqref{eq:ldho-over-linear-marginal-r} is proportional to $(\epsilon^2 + r^2)^{-(d+1)/2}$ as in the underdamped case~\eqref{eq:ldho-under-marginal-r-ok}.  The temporal marginal kernel~\eqref{eq:ldho-over-linear-marginal-t} comprises a combination of slow and fast  exponential kernels. This is analogous to the purely temporal case~\eqref{eq:cov-ldho-o}, albeit  the coefficients of the exponentials are renormalized and include temporal dependence.

\medskip
\subsection{Critical damping}

In this regime it holds that $\tilde{\omega}_{0} \tilde{\tau}_{c}=1/2$.  The temporal Fourier modes, based on~\eqref{eq:cov-ldho-c} and the dispersion relations, are given by
\beq
\label{eq:t-ft-modes-crit-linear}
\tC_{-\omega}(k,\tau) = c_{0} \E^{ -\frac{\lvert \tau \rvert B(k)}{2\tilde{\tau}_{c}} -\epsilon k} \left[ 1 + \frac{\lvert \tau \rvert \,B(k)}{2\tilde{\tau}_c }\right].
\eeq

\begin{theorem}[LDHO  kernel in critical damping regime]
\label{theo:ldho-kernel-st-crit-lin}
In this case, the LDHO spatiotemporal kernel generated by the IFT of the temporal Fourier modes~\eqref{eq:t-ft-modes-crit-linear}.
The spatiotemporal LDHO kernel in the critical damping regime is given by~\eqref{eq:ldho-kernel-st-critical-linear}.
\end{theorem}
\smallskip

\begin{IEEEproof}
The LDHO temporal Fourier modes  are obtained  from~\eqref{eq:cov-ldho-c} by inserting the dispersion functions~\eqref{eq:dispersion-lin} and~\eqref{eq:dispersion-fun-lin}.
Recalling~\eqref{eq:lin-B1} for $B(k)$, it follows that
\beq
\tC_{-\omega}(k,\tau) =c_{0} \E^{ -\frac{\lvert \tau \rvert (1+\xi k)}{2\tilde{\tau}_{c}} -\xi k}\, \left( 1+ \frac{\lvert \tau \rvert}{2\tau_c} + \frac{\xi\,k\lvert \tau \rvert}{2\tau_c}\right).
\eeq
Based on the linearity of the IFT we obtain
\begin{align}
\label{eq:C-crit-lin}
C(r,\tau)= & c_{0}\E^{ -\frac{\lvert \tau \rvert}{2\tilde{\tau}_{c}}}\, \left( 1+ \frac{\lvert \tau \rvert}{2\tau_c}\right) C_{1}(r,\tau)
+ \frac{c_{0} \xi  \,\lvert \tau \rvert}{2\tilde{\tau}_{c}} \, \E^{ -\frac{\lvert \tau \rvert}{2\tilde{\tau}_{c}}}\,  C_{2}(r,\tau),\end{align}
where $C_{1}(r,\tau)$ and $C_{2}(r,\tau)$ are given by
\begin{subequations}
\label{eq:C-crit-ift-lin}
\begin{align}
\label{eq:C1-crit-ift-lin}
C_{1}(r,\tau)= & \ift_{\bfk}\left[ \, \E^{-k\left(\epsilon + \frac{\xi\lvert \tau \rvert}{2\tilde{\tau}_{c}} \right)} \right],
\\
\label{eq:C2-crit-ift-lin}
C_{2}(r,\tau)= & \ift_{\bfk}\left[ \, k \,\E^{-k\left(\epsilon + \frac{\xi\lvert \tau \rvert}{2\tilde{\tau}_{c}} \right)} \right].
\end{align}
\end{subequations}
Based on~\eqref{eq:invcovft-iso}, the $\ift_{\bfk}$~\eqref{eq:C1-crit-ift-lin} can be expressed as
\[
C_{1}(r,\tau)=\frac{1}{(2\pi)^{d/2}\, r^\nu} \int_{0}^{\infty}\D k k^{d/2}\,J_{\nu}(kr)\,\E^{-a_{\Re}k}, \; \textrm{where} \; a_{\Re}=\epsilon + \frac{\xi\lvert \tau \rvert}{2\tilde{\tau}_{c}} \,.
\]
The integral in the above equation has been evaluated in~\eqref{eq:I-linear-over}, based on which we obtain
\beq
\label{eq:C1-crit-ift-linear}
C_{1}(r,\tau)= \frac{\Gamma(\frac{d+1}{2})}{\pi^{(d+1)/2} \left( r^{2} + a_{\Re}^{2}\right)^{(d+1)/2}}\,.
\eeq

Similarly, the IFT in~\eqref{eq:C2-crit-ift-lin} can be expressed as
\[
C_{2}(r,\tau)=\frac{1}{(2\pi)^{d/2}\, r^\nu} \int_{0}^{\infty}\D k k^{d/2 +1}\,J_{\nu}(kr)\,\E^{-a_{\Re}k}, \; \textrm{where} \; a_{\Re}=\epsilon + \frac{\xi\lvert \tau \rvert}{2\tilde{\tau}_{c}} \,.
\]
The latter can be expressed in terms of $C_{1}(r,\tau)$ by noticing that the only difference between the two is an extra factor of $k$ in the integral for $C_{2}(r,\tau)$. Therefore, by taking advantage that $C_{2}(r,\tau)$ depends on $a_{\Re}$ only via $\E^{-a_{\Re}k}$, it follows that
\[
C_{2}(r,\tau)= - \frac{\partial C_{1}(r,\tau)}{\partial a_{\Re}}\,.
\]
Thus, we obtain by differentiating $C_{1}(r,\tau)$ in~\eqref{eq:C1-crit-ift-linear}
\beq
\label{eq:C2-crit-ift-linear}
C_{2}(r,\tau)= \frac{\Gamma(\frac{d+1}{2})\, (d+1) \,a_{\Re}}{\pi^{(d+1)/2} \left( r^{2} + a_{\Re}^{2}\right)^{(d+3)/2}}\,.
\eeq
This concludes the proof.
\end{IEEEproof}

\smallskip

\begin{rem}[Hyperparameters at critical damping]
The critically damped LDHO kernel~\eqref{eq:ldho-kernel-st-critical-linear}  includes four independent  hyperparameters: $c_{0}, \tilde{\tau}_{c}, \epsilon, \xi$;  $\tilde{\omega}_{d}=0$ at critical damping.  The critical-damping kernel~\eqref{eq:ldho-kernel-st-critical-linear} can be viewed as the limit of the overdamped kernel~\eqref{eq:ldho-kernel-st-over-linear} for $\tilde{\omega}_{d} \to 0$, which implies $\beta_{s} \to 1$,  $\beta_{f} \to 1$.  The comments in Remark~\ref{rem:over} regarding the role of $\xi$ and $\epsilon$ also hold for the critically damped case.
\end{rem}

\paragraph{Zero-lag marginal covariances}
The spatial and temporal marginal kernels  are  obtained from~\eqref{eq:ldho-kernel-st-critical-linear} by setting $\tau=0$ and $r=0$ respectively. We thus obtain
\begin{align}
\label{eq:ldho-crit-marginal-r-linear}
 \Cms(r)  & =  c_{0}\, C_{1}(r, 0)=
 \frac{c_{0} \Gamma(\frac{d+1}{2})}{\pi^{(d+1)/2} \left( r^{2} + \epsilon^{2}\right)^{(d+1)/2}}\,,
\end{align}
\begin{align}
\label{eq:ldho-crit-marginal-t-linear}
\Cmt(\tau) & = c_{0}\, \frac{\,\Gamma(\frac{d+1}{2})}{ \pi^{(d+1)/2}\, a_{\Re}^{d+1}}\, \E^{-\frac{\lvert \tau \rvert}{2\tilde{\tau}_{c}}}\, \left[ \left( 1 + \frac{\lvert \tau \rvert}{2\tilde{\tau}_{c}}\right)   + \frac{\xi\lvert \tau \rvert}{2\tilde{\tau}_{c}} \, \frac{(d+1) }{a_{\Re}}\right]\,.
\end{align}

\medskip

\setcounter{subsection}{0}
\section*{S4. Kernels based on the Ornstein-Uhlenbeck ODE}
\label{S-sec:o-u-kernel}

The covariance of the O-U process is given by 
$C(\tau)=\sigma^2\, \exp(-\lvert \tau \rvert/\tau_c)$ where $\sigma^2 =\sigma_{\eta}^{2}\tau_{c}/{2}$~\cite[p.~448]{Papoulis02}. The radial dispersion relations are given by 
$\sigma^{2} \to \sigma^{2}(k) = \sigma_{0}^{2} A(k), \, \tau_{c} \to \tau_{c}(k)= \tilde{\tau}_{c}/B(k)$.  The O-U temporal Fourier modes for radial dispersion functions are thus given by 
\begin{align}
\label{eq:t-ft-modes-O-U}
\tC_{-\omega}(k,\tau) &  = \sigma^{2}_{0}  \, A(k) \, \exp\left[ -\lvert \tau\rvert\,  B(k)/\tilde{\tau}_{c}\right].
\end{align}
Since $A(k)$ is dimensionless, $[\sigma_{0}^{2}]=[\mathrm{X}]^{2}[\mathrm{L}]^d$, where $\mathrm{L}$ represents length, so that the FT~\eqref{eq:t-ft-modes-O-U} be dimensionally correct. 
Based on the IFT~\eqref{eq:invcovft-iso}, the O-U covariance kernel is given by the integral $(\nu=d/2-1)$:
\begin{equation}
    \label{eq:invcovft-O-U}
    C({r},\tau)= \frac{\sigma^{2}_{0}}{(2\pi)^{d/2} r^{\nu}} \int_{0}^{\infty}  k^{d/2}\,
    {J_{\nu}(k  r)}  A(k) \, \E^{ -\lvert \tau\rvert\,  B(k)/\tilde{\tau}_{c}} \, \D k\,.
\end{equation}

In the following, we derive spatiotemporal  kernel expressions for two different choices of dispersion functions. 

\medskip

\begin{enumerate}
\itemsep1em

\item  $A(k)=\E^{-\beta k^2}, \; B(k)=a+bk^2$ where $a,b, \beta>0$ are hyperparameters with units $[b]=[\beta]=[L]^2$, $[a]=[L]^{0}$:  

The spectral integral~\eqref{eq:invcovft-O-U} becomes
\[
C({r},\tau)= \frac{\sigma^{2}_{0}\, 
\E^{- a\,\lvert \tau\rvert/\tilde{\tau}_{c}}}{(2\pi)^{d/2}\, r^{\nu}}  \int_{0}^{\infty}  k^{d/2}    {J_{\nu}(k  r)} \, \E^{ -b k^{2} \lvert \tau\rvert\,  /\tilde{\tau}_{c} - \beta k^2} \, \D k\,.
\]
Using the table of integrals~\cite[6.631.4, p.~706]{Gradshteyn07} it follows that
\beq
\label{eq:o-u-k-square}
C(r,\tau)= \frac{\sigma^{2}_{0}\, \E^{- a\,\lvert \tau\rvert/\tilde{\tau}_{c}}}{(2\pi)^{d/2}}\,\frac{\E^{- r^{2} / 4\left( \beta + b\,\lvert \tau \rvert/\tilde{\tau}_c \right) }}{\left( \frac{2b\lvert \tau\rvert}{\tilde{\tau}_{c}}+2\beta\right)^{d/2}} \,.
\eeq
The kernel~\eqref{eq:o-u-k-square} involves four free hyperparameters: $\sigma_{0}$, $\tilde{\tau}_{c}/a$, $\tilde{\tau}_{c}/b$ and $\beta$. The spatial and temporal marginal kernels are given respectively by
\begin{subequations}
\label{eq:O-U-kernel-square-marginals}
\begin{align}
\Cms(r) = & \frac{\sigma^{2}_{0}\, }{(4\pi\, \beta)^{d/2}}\,\E^{-r^{2}/4\beta}\,,
    \\
\Cmt(\tau) = & \frac{\sigma^{2}_{0}\, \left(  1 + b\,\lvert \tau \rvert/\beta \tilde{\tau}_c\right)^{-d/2}}{(4\pi\, \beta)^{d/2}}\,\E^{-a\,\lvert \tau\rvert/\tilde{\tau}_{c}}\,  \,.  
\end{align}
\end{subequations}
Hence, the spatial marginal covariance is given by the square exponential kernel while the temporal marginal is a modified exponential kernel.   

\item $A(k)=\E^{-\beta k}$, $B(k)=a + \xi\,k$, where $a,\xi,\beta>0$  are hyperparameters with units $[b]=[\beta]=[L]$, $[a]=[L]^{0}$:  

The spectral  integral~\eqref{eq:invcovft-O-U} becomes ($\nu=d/2-1$):
\[
C({r},\tau)= \frac{\sigma^{2}_{0}\, 
\E^{- a\,\lvert \tau\rvert/\tilde{\tau}_{c}}}{(2\pi)^{d/2}\, r^{\nu}}  \int_{0}^{\infty}  k^{\nu+1}    {J_{\nu}(k r)} \, \E^{ - k \left(\beta +\xi \,  \lvert \tau\rvert\,  /\tilde{\tau}_{c} \right)} \, \D k\,.
\]
Using the table of integrals~\cite[6.623.2, p.~702]{Gradshteyn07} we obtain
\beq
\label{eq:o-u-k-linear}
C(r,\tau)= \frac{\sigma^{2}_{0}\, \Gamma(\frac{d+1}{2})\,}{\pi^{(d+1)/2}}\frac{\left( \beta \tilde{\tau}_{c} +\xi \,\lvert \tau\rvert \, \right) \,\E^{- a\,\lvert \tau\rvert/\tilde{\tau}_{c}}}{\tilde{\tau}_{c} \,\left[ r^{2} + \left( \beta + \frac{\xi\lvert \tau\rvert}{\tilde{\tau}_{c}}\right)^{2} \right]^{(d+1)/2}} \,.
\eeq
The kernel~\eqref{eq:o-u-k-linear} involves four free hyperparameters: $\sigma_{0}$, $\tilde{\tau}_{c}/a$, $\tilde{\tau}_{c}/\xi$ and $\beta$. The spatial and temporal marginal kernels are given respectively by
\begin{subequations}
\label{eq:O-U-kernel-linear-marginals}
\begin{align}
\Cms(r) = & \frac{\sigma^{2}_{0}\, \Gamma(\frac{d+1}{2})\,}{\pi^{(d+1)/2}}\frac{ \beta }{\left( r^{2} +  \beta^2 \right)^{(d+1)/2}} \,\,,
    \\
\Cmt(\tau) = & \frac{\sigma^{2}_{0}\, \Gamma(\frac{d+1}{2})\,}{\pi^{(d+1)/2}}\frac{\,\E^{- a\,\lvert \tau\rvert/\tilde{\tau}_{c}}}{\left( \beta + \frac{\xi\lvert \tau\rvert}{\tilde{\tau}_{c}}\right)^{d}} \,.  
\end{align}
\end{subequations}

The Ornstein-Uhlenbeck covariance kernels with square and linear $k$ dependence of the dispersion functions are illustrated in Fig.~\ref{fig:kernel_OU}. 
\begin{figure}
\centering
\includegraphics[width=.49\linewidth]{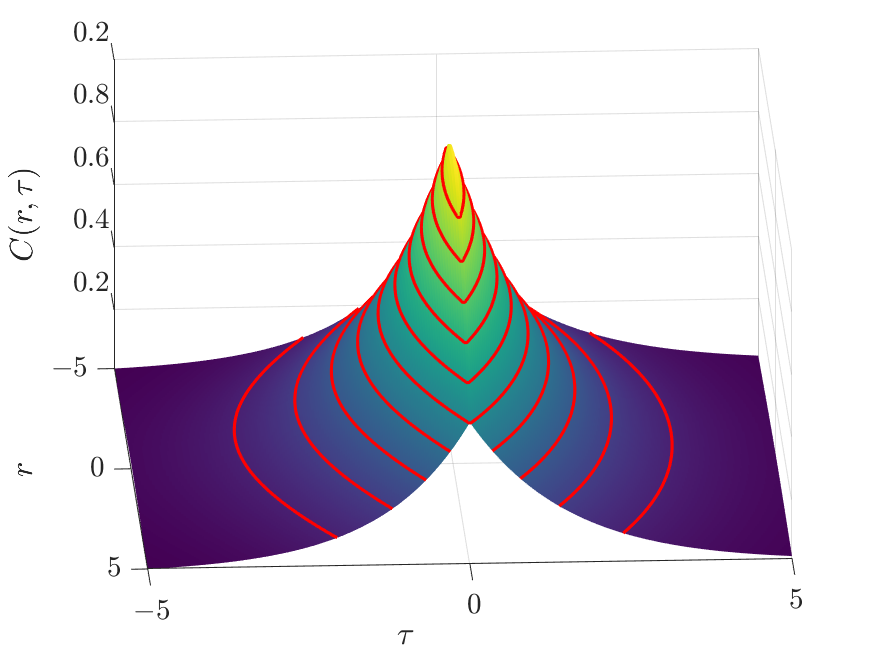}
\includegraphics[width=.49\linewidth]{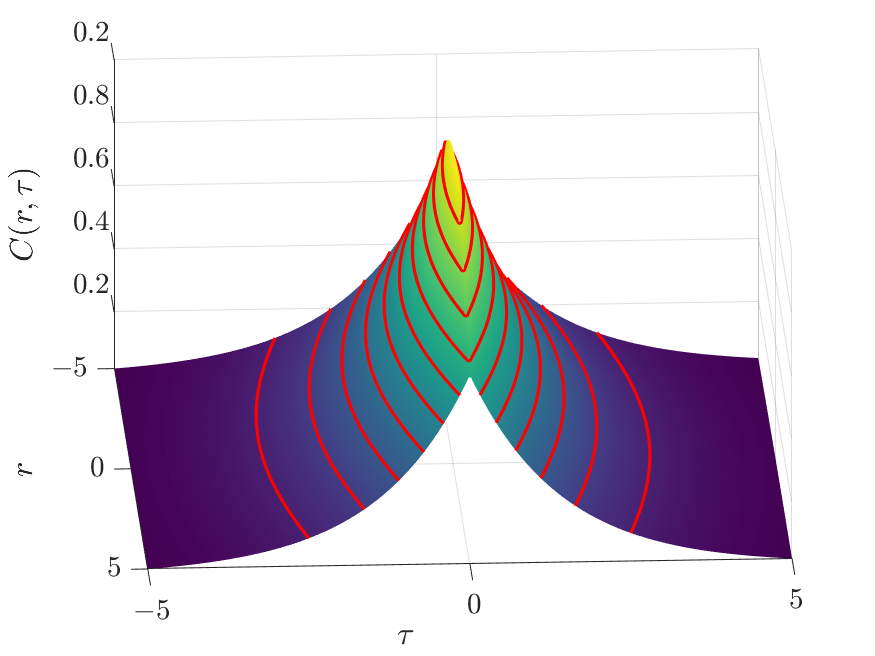}
\caption{Normalized $C(r,\tau)$ and isolevel contour lines (red online) of the Ornstein-Uhlenbeck kernels in $d=2$ spatial dimensions. Left: obtained from the dispersion function~\eqref{eq:o-u-k-square} with $k^2$ dependence using the  hyperparameters 
$\tilde{\tau}_{c}=0.8$, $b=0.4$, $a=0.5$,  $\beta=8$.  Right: obtained from the dispersion functions~\eqref{eq:o-u-k-linear} with $k$ dependence using the  hyperparameters  
$\tilde{\tau}_{c}=0.8$, $\xi=0.4$, $a=0.5$,  $\beta=8$.}
\label{fig:kernel_OU}
\end{figure}

 \end{enumerate}

 \end{document}